\RequirePackage[svgnames]{xcolor}

\documentclass[11pt,letterpaper]{mystyle}

\usepackage[svgnames]{xcolor}
\usepackage[comma,authoryear,compress]{natbib}
\usepackage{hyperref}[citecolor=lightblue]

\hypersetup{
    colorlinks = true,
    citecolor = {YaleBlue},
}

\usepackage{algorithm}
\usepackage{algorithmicx}
\usepackage{algpseudocode}
\usepackage{microtype}
\usepackage{graphicx}
\expandafter\def\csname ver@subfig.sty\endcsname{}
\usepackage{booktabs} %
\usepackage{float}
\usepackage{bigstrut}

\usepackage{amsmath}
\usepackage{amssymb}
\usepackage{mathtools}
\usepackage{amsthm}
\usepackage{mathrsfs}
\usepackage{nicefrac}
\usepackage{dsfont}
\usepackage{enumitem}
\usepackage{subcaption}
\usepackage{cleveref}
\usepackage{bxcoloremoji}

\usepackage[utf8]{inputenc} %
\usepackage[T1]{fontenc}    %
\usepackage{url}            %
\usepackage{amsfonts}       %
\usepackage{wrapfig}
\usepackage{lipsum}
\usepackage{stackengine}
\usepackage[font=small,labelfont=bf]{caption}
\usepackage{color}
\usepackage{adjustbox}

\usepackage{rotating}
\usepackage{makecell}

\usepackage{array}
\usepackage{tabularx}
\usepackage{multirow}
\usepackage{colortbl}
\usepackage{arydshln}
\usepackage{epstopdf}
\usepackage{pifont} 
\usepackage{bbding}

\newcolumntype{L}[1]{>{\raggedright\arraybackslash}m{#1}}
\newcolumntype{C}[1]{>{\centering\arraybackslash}m{#1}}
\newcolumntype{R}[1]{>{\raggedleft\arraybackslash}m{#1}}

\newcommand{\x}{\mathbf{x}}

\definecolor{blanchedalmond}{rgb}{1.0, 0.92, 0.8}
\definecolor{carmine}{rgb}{0.59, 0.0, 0.09}
\definecolor{lightblue}{rgb}{0.22,0.45,0.70}%

\renewcommand{\mathbf}{\boldsymbol}

\makeatletter
\def\Ddots{\mathinner{\mkern1mu\raise\p@
\vbox{\kern7\p@\hbox{.}}\mkern2mu
\raise4\p@\hbox{.}\mkern2mu\raise7\p@\hbox{.}\mkern1mu}}
\makeatother

\definecolor{amaranth}{rgb}{0.9, 0.17, 0.31}
\definecolor{antiquebrass}{rgb}{0.8, 0.58, 0.46}
\definecolor{antiquefuchsia}{rgb}{0.57, 0.36, 0.51}
\definecolor{chromeyellow}{rgb}{0.31, 0.47, 0.26}

\newcommand{\github}{\raisebox{-1.5pt}{\includegraphics[height=1.05em]{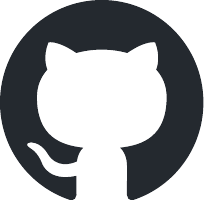}}}

\tcbuselibrary{breakable, skins}
\tcbset{aibox/.style={
    enhanced,
    colback=white,
    colframe=Gray,
    boxrule=0.5pt,
    arc=2pt,
    fonttitle=\bfseries,
    coltitle=black,
    attach title to upper,
    after title={\par\medskip},
    breakable
}}
\newtcolorbox{AIbox}[2][]{aibox,title=#2,#1}
\definecolor{lightblue}{rgb}{0.22,0.45,0.70}%
\definecolor{Gray}{gray}{0.95}
\definecolor{Cornsilk}{rgb}{1.0, 0.97, 0.86}

\newcommand{\cmark}{\ding{51}}%
\newcommand{\xmark}{\ding{55}}%
\newcommand{\our}{ResAdapt}

\definecolor{metablue}{HTML}{0064E0}
\definecolor{metafg}{HTML}{1C2B33}
\definecolor{metabg}{HTML}{F1F4F7}
\definecolor{casiaorange}{HTML}{FF7F0E}
\definecolor{ucaspink}{HTML}{E377C2}
\definecolor{purple}{RGB}{75,46,131}

\usepackage[all]{hypcap}

\crefname{section}{§}{§§}
\Crefname{section}{§}{§§}

\title{\our{}: Adaptive Resolution for Efficient Multimodal Reasoning}

\runningtitle{\our{}: Adaptive Resolution for Efficient Multimodal Reasoning}

\author[ ]{
    Huanxuan Liao\textsuperscript{\textcolor{casiaorange}{$\tau$}, \textcolor{ucaspink}{$\mu$}}, 
    Zhongtao Jiang, 
    Yupu Hao\textsuperscript{\textcolor{casiaorange}{$\tau$}, \textcolor{ucaspink}{$\mu$}}, 
    Yuqiao Tan\textsuperscript{\textcolor{casiaorange}{$\tau$}, \textcolor{ucaspink}{$\mu$}}, 
    Shizhu He\textsuperscript{\textcolor{casiaorange}{$\tau$}, \textcolor{ucaspink}{$\mu$}}, \\
    Ben Wang, 
    Jun Zhao\textsuperscript{\textcolor{casiaorange}{$\tau$}, \textcolor{ucaspink}{$\mu$}}, 
    Kun Xu\textsuperscript{\textcolor{purple}{$\dagger$}},
    Kang Liu\textsuperscript{\textcolor{casiaorange}{$\tau$}, \textcolor{ucaspink}{$\mu$}, \textcolor{purple}{$\ast$}}
}

\affil[ ]{
    \textsuperscript{\textcolor{casiaorange}{$\tau$}}Institute of Automation, Chinese Academy of Sciences \\
    \textsuperscript{\textcolor{ucaspink}{$\mu$}}University of Chinese Academy of Sciences \\
    \textsuperscript{\textcolor{purple}{$\dagger$}}Project Leader \quad
    \textsuperscript{\textcolor{purple}{$\ast$}}Corresponding author: \href{mailto:kliu@nlpr.ia.ac.cn}{kliu@nlpr.ia.ac.cn}
}

\begin{document}

\begin{abstract}
Scaling both spatial resolution and temporal coverage in video reasoning demands visual-token budgets that grow prohibitively for Multimodal Large Language Models (MLLMs). Existing efficiency strategies intervene too late: model-side token pruning discards fine-grained evidence after the encoder has already paid the full computational cost, while output-side iterative retrieval introduces multi-turn latency. We propose \textbf{\our{}}, a framework that reallocates visual budget \emph{before} encoding. A lightweight, query-aware Allocator predicts a per-frame resolution scale, adjusting the pixels the backbone receives while preserving its native token interface and compatibility with optimized inference engines. To train this non-differentiable pipeline, we introduce \textbf{Cost-Aware Policy Optimization (CAPO)}, which combines a dynamic cost pivot with asymmetric reward shaping to jointly maximize reasoning accuracy under strict visual budgets---preventing the policy collapse that plagues direct cost penalties. The resulting Allocator concentrates pixels on information-dense frames, exhibiting content-adaptive active perception learned entirely from task reward. Across video QA and temporal grounding benchmarks, \our{} matches or exceeds uncompressed baselines while eliminating over 90\% of visual tokens. Crucially, the saved spatial budget is reinvested into temporal coverage: under equivalent compute, \our{} processes $16\times$ more frames, yielding $>15\%$ relative gains on complex long-video reasoning tasks.

\vspace{5mm}

\coloremojicode{1F310} \textbf{Project Page}: \href{https://xnhyacinth.github.io/projects/ResAdapt}{https://xnhyacinth.github.io/projects/ResAdapt}

\github{} \textbf{Code Repository}: \href{https://github.com/Xnhyacinth/ResAdapt}{https://github.com/Xnhyacinth/ResAdapt}

\coloremojicode{1F4E7} \textbf{Contact}: \href{mailto:liaohuanxuan2023@ia.ac.cn}{liaohuanxuan2023@ia.ac.cn}

\end{abstract}

\maketitle
\vspace{3mm}
\section{Introduction}

Multimodal Large Language Models (MLLMs) achieve stronger visual understanding by scaling input fidelity, yet the resulting visual-token growth makes jointly sustaining high spatial resolution and long temporal context prohibitive~\citep{guo2025deepseek,bai2025qwen3,liu2025comprehensive,shu2025videoxl,shao2025tokens}. In practice, this trade-off is central to video reasoning: reducing resolution risks losing the small visual cues that determine the answer, whereas shortening the clip removes the temporal context needed for long-horizon inference. Even architecturally efficient encoders~\citep{zhang2026penguin,liu2025nvila} do not remove this tension; they merely shift where it becomes painful.

\begin{figure}[t]
\centering
\includegraphics[width=0.985\linewidth]{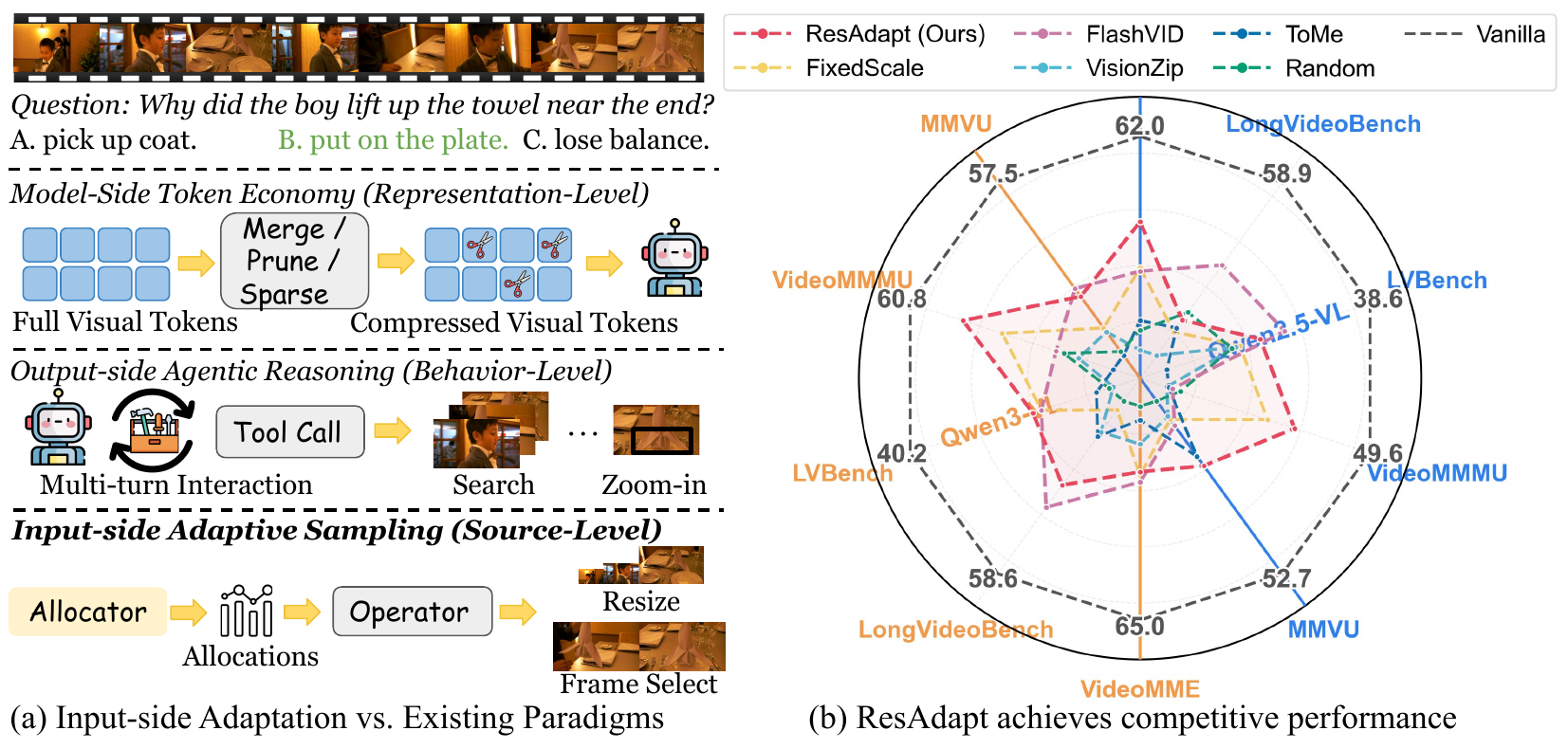}
\caption{\textbf{Input-side Adaptation improves the visual-token efficiency frontier.} \textbf{(a)} Three efficiency paradigms for video reasoning. Model-side methods compress tokens after encoding; output-side methods iteratively retrieve or zoom; \our{} reallocates per-frame visual budget before encoding, preserving the backbone's native token interface and compatibility with optimized inference engines. \textbf{(b)} Qwen2.5-VL-7B results with 32 frames at ${\sim}$10\% visual retention, where \our{} lies on or near the Pareto frontier and shows its largest gain on the reasoning-heavy benchmark.}
\label{fig:intro}
\end{figure}

Mainstream efficiency methods largely fall into two paradigms (Figure~\ref{fig:intro}a), both of which intervene too late and share a common root: \emph{they accept the encoder's full-resolution input as a fixed cost and attempt to recover efficiency downstream}. \textit{Model-side} approaches prune or merge tokens after visual encoding~\citep{khaki2025sparsevila,xu2025streamingvlm,bolya2022token,tao2025dycoke}. Once fine-grained evidence is discarded, it cannot be recovered, and the irregular token layouts that result from pruning or merging disrupt optimized attention kernels and inference engines~\citep{dao2023flashattention,kwon2023efficient,zheng2024sglang}. Conversely, \textit{output-side} agentic reasoning introduces iterative retrieval or zoom steps~\citep{zhang2025rewatch,yang2025longvt,shen2025zoom,zheng2025deepeyes}. While this strategy recovers coverage, it multiplies inference cost: each retrieval step demands a separate backbone call, and the initial coarse view that triggers refinement frequently undersamples the very cues it seeks to recover.

We argue that the intervention point itself is the problem. Rather than compressing representations \emph{after} encoding or retrieving them \emph{after} reasoning, an efficient system should optimize the pixel volume the encoder receives in the first place. Our framework, \textbf{\our{}}, instantiates this \emph{input-side adaptation} principle: a lightweight Allocator predicts a per-frame visual allocation from coarse features and the query, then realizes that allocation through a visual budget operator, such as resolution resizing or frame selection. The backbone therefore processes a standard---albeit shorter---visual-token sequence in a single call, preserving full compatibility with FlashAttention, vLLM~\citep{kwon2023efficient}, and SGLang~\citep{zheng2024sglang} without bespoke kernel engineering. Compared with prior slow--fast pipelines~\citep{yang2025kwai,zhang2026penguin}, which route frames using query-agnostic heuristics or fixed resolution tiers, \our{} learns a query-aware allocation policy directly from task reward.

Optimizing this pre-encoding allocation presents severe reinforcement learning challenges: the action space is continuous, the visual operator is non-differentiable, and naive accuracy--cost penalties catastrophically collapse the policy toward minimum budgets. We overcome these optimization hurdles with \textbf{Cost-Aware Policy Optimization (CAPO)}, which converts sparse rollout feedback into a stable asymmetric learning signal, and a temporal-similarity regularizer that suppresses redundant high-budget allocations on adjacent similar frames. Together, these components transform Input-side adaptation into a trainable, content-aware policy rather than a handcrafted compression rule.

Extensive empirical evaluations across video QA and temporal grounding benchmarks demonstrate that \our{} decisively advances the efficiency--accuracy Pareto frontier. \our{} matches or surpasses state-of-the-art token economy methods while discarding over 90\% of visual tokens (Figure~\ref{fig:intro}b). Crucially, this spatial compression unlocks massive temporal expansion: under equivalent computational budgets, \our{} processes $16\times$ more frames, yielding $>15\%$ relative performance gains. Furthermore, the learned policy exhibits \textit{active perception}---autonomously concentrating visual budget on decisive frames in a single forward pass, without requiring explicit saliency supervision.

Our main contributions are:
\begin{enumerate}[leftmargin=1.5em,itemsep=2pt]
    \item We introduce \textbf{\our{}}, an \textit{input-side adaptation} framework that formulates dynamic per-frame visual budgeting as a contextual bandit problem, fully preserving the native architecture and hardware optimizations of MLLMs.
    \item We propose \textbf{CAPO} with a temporal similarity regularizer, providing a stable, asymmetric learning signal that jointly optimizes accuracy and cost without hand-crafted heuristics.
    \item Extensive experiments and ablations demonstrate that \our{} achieves a superior efficiency--accuracy Pareto frontier across video QA and temporal grounding tasks, with the learned policy exhibiting content-adaptive active perception.
\end{enumerate}

\section{Background and Problem Formulation}
\label{main:preliminaries}

\subsection{Preliminaries}
Given a text query \(\boldsymbol{q}\) and a video \(\mathcal{V}=\{f_t\}_{t=1}^{T}\), let \(\boldsymbol{x}=(\boldsymbol{q},\mathcal{V})\) denote the full input. A backbone policy \(\pi_\phi\) encodes every frame at fixed fidelity and autoregressively generates a rollout \(\boldsymbol{y}=(y_1,\dots,y_L)\):
\begin{equation}
\pi_\phi(\boldsymbol{y}\mid\boldsymbol{x})=\prod_{j=1}^{L}\pi_\phi(y_j\mid y_{<j},\boldsymbol{x}).
\end{equation}
When useful, we write \(\boldsymbol{y}=(\boldsymbol{r},\boldsymbol{o})\) for a reasoning trace \(\boldsymbol{r}\) and a final answer \(\boldsymbol{o}\). The computational inefficiency of this paradigm is stark: visual encoding cost scales quadratically with pixel volume, yet the evidence required to answer complex queries remains remarkably sparse in time.

To control pre-encoding cost, we introduce an Allocator policy \(\pi_\theta\) that emits a per-frame allocation vector
\begin{equation}
\boldsymbol{s}=(s_1,\dots,s_T)\sim\pi_\theta(\cdot\mid\boldsymbol{x}), \qquad s_t\in[s_{\min},s_{\max}],
\end{equation}
and applies a \emph{visual budget operator} \(\mathcal{O}\) to each frame: \(\tilde{f}_t=\mathcal{O}(f_t,s_t)\). The backbone then generates from the transformed input \(\tilde{\boldsymbol{x}}=(\boldsymbol{q},\{\tilde{f}_t\}_{t=1}^{T})\):
\begin{equation}
\pi_\phi(\boldsymbol{y}\mid\tilde{\boldsymbol{x}})=\prod_{j=1}^{L}\pi_\phi(y_j\mid y_{<j},\tilde{\boldsymbol{x}}).
\end{equation}
We keep \(\mathcal{O}\) abstract only to state the decision problem cleanly. The framework is operator-agnostic: \(\mathcal{O}\) may implement resizing, frame selection, or other pre-encoding budget controls. 

\subsection{Problem Formulation}
\label{sec:problem_formulation}
Because the Allocator acts once before decoding, the outer problem is a \emph{Contextual Bandit} (equivalently, a one-step contextual MDP). The context is the raw input \(\boldsymbol{x}\in\mathcal{X}\), and the action is the continuous allocation vector \(\boldsymbol{s}\in[s_{\min},s_{\max}]^{T}\). For joint training, it is convenient to write the induced two-stage policy as
\begin{equation}
p_{\theta,\phi}(\boldsymbol{s}, \boldsymbol{y} \mid \boldsymbol{x}) = \pi_\theta(\boldsymbol{s} \mid \boldsymbol{x}) \, \pi_\phi(\boldsymbol{y} \mid \tilde{\boldsymbol{x}}),
\end{equation}
where \(\tilde{\boldsymbol{x}} = (\boldsymbol{q}, \{\mathcal{O}(f_t, s_t)\}_{t=1}^{T})\) is the deterministically transformed input. The immediate reward is response quality \(r(\boldsymbol{x}, \boldsymbol{s}, \boldsymbol{y}) = Q(\boldsymbol{x}, \boldsymbol{y})\).

Let \(C(\boldsymbol{s})\) denote the visual cost induced by allocation \(\boldsymbol{s}\). The ideal budgeted objective is
\begin{equation}
\begin{aligned}
\max_{\theta,\phi}\quad
&\mathbb{E}_{\boldsymbol{x}\sim\mathcal{D},\,\boldsymbol{s}\sim\pi_\theta(\cdot\mid\boldsymbol{x}),\,\boldsymbol{y}\sim\pi_\phi(\cdot\mid\tilde{\boldsymbol{x}})}
\!\left[Q(\boldsymbol{x},\boldsymbol{y})\right] \\
\text{s.t.}\quad
&\mathbb{E}_{\boldsymbol{x}\sim\mathcal{D},\,\boldsymbol{s}\sim\pi_\theta(\cdot\mid\boldsymbol{x})}
\!\left[C(\boldsymbol{s})\right]\le \tau,
\end{aligned}
\label{eq:problem_constrained}
\end{equation}
where \(\tau\) is the target budget. Lagrangian relaxation yields the unconstrained utility
\begin{equation}
\begin{aligned}
\max_{\theta,\phi}\quad
&\mathbb{E}_{\boldsymbol{x},\boldsymbol{s},\boldsymbol{y}}
\!\left[\mathcal{U}(\boldsymbol{x},\boldsymbol{s},\boldsymbol{y})\right], \\
\mathcal{U}(\boldsymbol{x},\boldsymbol{s},\boldsymbol{y})
&=
Q(\boldsymbol{x},\boldsymbol{y})-\lambda\,C(\boldsymbol{s}),
\end{aligned}
\label{eq:problem_lagrangian}
\end{equation}
for trade-off coefficient \(\lambda\ge 0\).

Equations~\eqref{eq:problem_constrained}--\eqref{eq:problem_lagrangian} define the target trade-off but not yet a stable optimizer. Section~\ref{main:method} instantiates this objective with an Input-side adaptation policy, CAPO, temporal regularization, and PPO-style surrogate losses; the experiments use resize as the concrete operator. Detailed derivations are deferred to Appendix~\ref{app:derivation}.

\section{Method}
\label{main:method}

\begin{figure*}[t]
\centering
\includegraphics[width=0.985\linewidth]{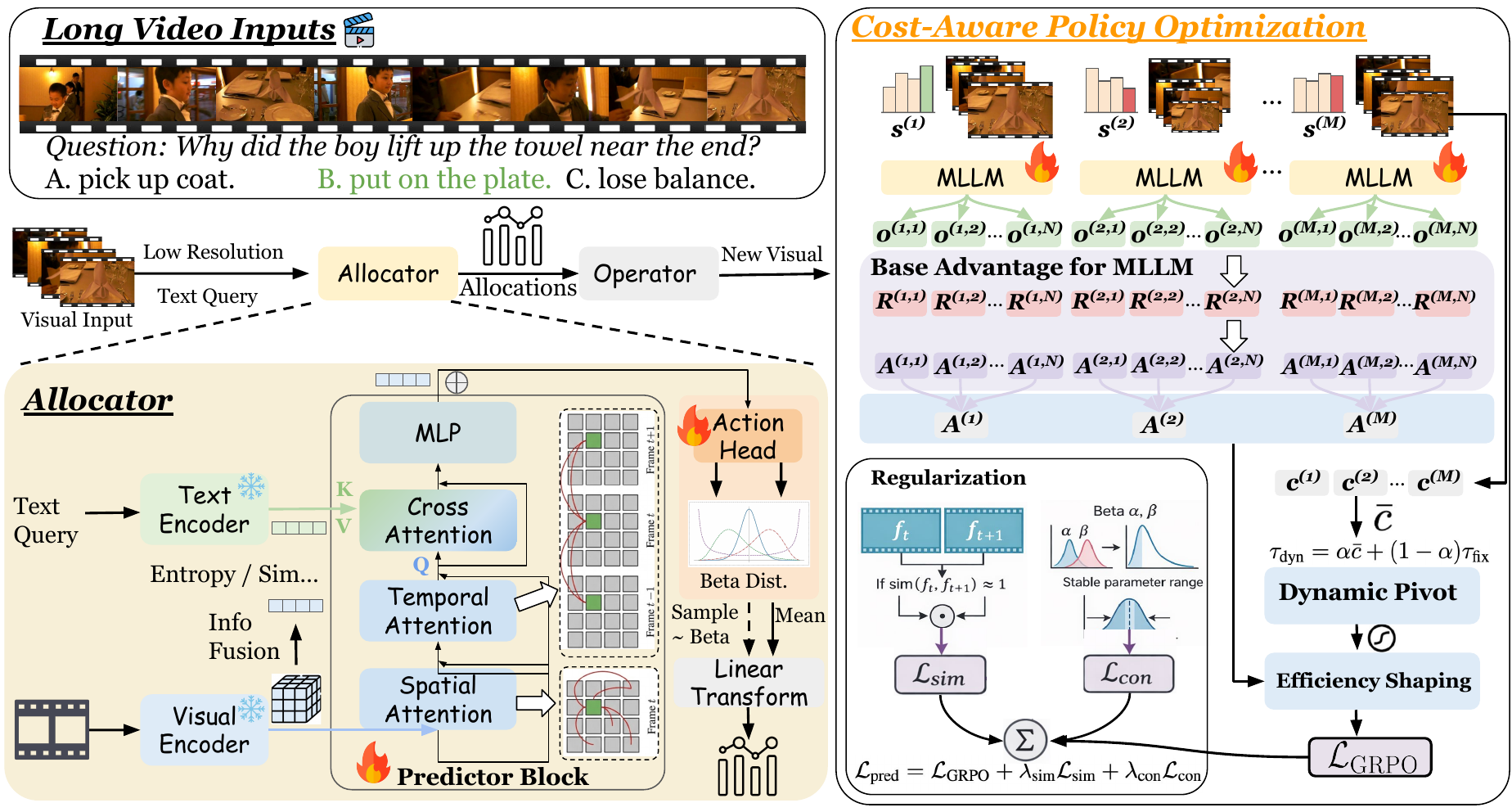}
\caption{\textbf{\our{} framework.} \textbf{(a)} At inference, a lightweight Allocator \(\pi_\theta\) maps coarse visual features and the query to latent actions \(a_t\sim\mathrm{Beta}(\alpha_t,\beta_t)\), which parameterize per-frame input allocations. In the resize instantiation used in our experiments, these allocations are realized as scales \(s_t\in[s_{\min},s_{\max}]\), and the resized frames are processed by the MLLM in a single call. \textbf{(b)} During training, CAPO reshapes group-relative advantages with a dynamic cost pivot \(\tau_{\text{dyn}}\), while temporal-similarity regularization suppresses redundant high-budget allocation on adjacent similar frames.}
\label{fig:method}
\end{figure*}

Figure~\ref{fig:method} illustrates \our{}. At inference, the Allocator predicts a scalar allocation per frame and applies a pre-encoding operator before the video reaches the backbone. In our primary instantiation, the operator \(\mathcal{O}\) performs bilinear resizing: the allocation determines a per-frame resize factor \(s_t\), yielding \(\tilde{f}_t=\mathcal{R}(f_t,s_t)\). At training time, rollout feedback from the backbone updates the Allocator and, optionally, the backbone itself.

\subsection{Joint RL Optimization Framework}
\label{sec:joint_obj}
As formulated in Section~\ref{sec:problem_formulation}, we cast pre-encoding allocation as a contextual bandit. Starting from the marginal probability of generating the correct answer under the transformed input (see Appendix~\ref{app:derivation}), we derive a one-step expected-reward objective. Abstracting the answer-quality term as a rollout utility \(Q(\boldsymbol{x},\boldsymbol{y})\)---treated as parameter-free once \(\boldsymbol{y}\) has been sampled---the joint policy factorizes as
\begin{equation}
p_{\theta,\phi}(\boldsymbol{s},\boldsymbol{y}\mid\boldsymbol{x})
=
\pi_\theta(\boldsymbol{s}\mid\boldsymbol{x})\,
\pi_\phi(\boldsymbol{y}\mid\tilde{\boldsymbol{x}}).
\end{equation}
Here \(\pi_\theta(\boldsymbol{s}\mid\boldsymbol{x})\) is the density induced by the latent Beta policy \(q_\theta(\boldsymbol{a}\mid\boldsymbol{x})\) through the affine map in Eq.~\eqref{eq:scale_map}. Because this map has a \(\theta\)-independent Jacobian, \(\nabla_\theta \log \pi_\theta(\boldsymbol{s}\mid\boldsymbol{x})\) coincides with \(\nabla_\theta \log q_\theta(\boldsymbol{a}\mid\boldsymbol{x})\), so all PPO ratios can be evaluated directly on the latent actions \(a_t\) (Eq.~\ref{eq:logprob}).
The corresponding ideal rollout reward combines task quality and visual cost:
\begin{equation}
R^{\text{ideal}}_{\boldsymbol{s},\boldsymbol{y}}
=
Q(\boldsymbol{x},\boldsymbol{y})-\lambda\,C(\boldsymbol{s}),
\label{eq:ideal_rollout_reward}
\end{equation}
and the optimization target becomes
\begin{equation}
\max_{\theta,\phi}\ \mathcal{J}(\theta,\phi)
=
\mathbb{E}_{\pi_\theta(\boldsymbol{s}\mid\boldsymbol{x})}
\!\left[
\mathbb{E}_{\pi_\phi(\boldsymbol{y}\mid\tilde{\boldsymbol{x}})}
\!\left[
R^{\text{ideal}}_{\boldsymbol{s},\boldsymbol{y}}
\right]
\right].
\label{eq:joint_obj}
\end{equation}
Equation~\eqref{eq:joint_obj} defines the expected return for a single context \(\boldsymbol{x}\); training marginalizes over \(\boldsymbol{x}\sim\mathcal{D}\). While the policy gradients follow the standard score-function estimator (Appendix~\ref{app:derivation}), directly optimizing this objective is brittle in practice due to three challenges:
\begin{enumerate}
\item \textbf{Policy parameterization.} \(\pi_\theta\) must emit a \(T\)-dimensional continuous action with negligible overhead relative to the backbone.
\item \textbf{Credit assignment.} The raw Lagrangian reward \(Q(\boldsymbol{x},\boldsymbol{y})-\lambda C(\boldsymbol{s})\) exhibits extreme variance and frequently collapses the policy to the minimum allowable budget, since every reduction in \(C\) is unconditionally rewarded regardless of answer quality.
\item \textbf{Temporal structure.} Rollout-level rewards carry no frame-level granularity, permitting redundant high-budget allocations on visually near-duplicate neighbors.
\end{enumerate}
The remainder of this section resolves each bottleneck in turn.

\subsection{Allocator Architecture}
\label{sec:allocator}
Each frame $f_t \in \mathbb{R}^{3 \times H_t \times W_t}$ is encoded by a frozen lightweight visual encoder; the query is encoded separately. Both representations are projected to a shared dimension \(D\). A shallow Transformer decoder alternates temporal self-attention over \(\{\boldsymbol{f}_t\}_{t=1}^{T}\) with gated cross-attention to the query, producing per-frame hidden states \(\{h_t\}_{t=1}^{T}\). This architecture exposes both temporal redundancy and query dependence at low cost.

We parameterize each latent action with a Beta distribution, whose bounded support maps naturally to \([s_{\min},s_{\max}]\):
\begin{equation}
a_t \sim \mathrm{Beta}(\alpha_t,\beta_t), \qquad
s_t = s_{\min} + a_t\,(s_{\max}-s_{\min}).
\label{eq:scale_map}
\end{equation}
Since \(a_t\in(0,1)\), the allocation satisfies \(s_t\in(s_{\min},s_{\max})\) almost surely; setting \(0<s_{\min}<1<s_{\max}\) permits both downscaling and selective upscaling. Let \(q_\theta(\boldsymbol{a}\mid\boldsymbol{x})\) denote the joint latent policy over \(\boldsymbol{a}=(a_1,\dots,a_T)\). Conditioned on \(\{h_t\}\), the log-density factorizes across frames:
\begin{equation}
\log q_\theta(\boldsymbol{a}\mid \boldsymbol{x})=\sum_{t=1}^{T}\log \mathrm{Beta}(a_t;\alpha_t,\beta_t).
\label{eq:logprob}
\end{equation}
The affine map \(\boldsymbol{a}\mapsto\boldsymbol{s}\) induces the allocation policy \(\pi_\theta(\boldsymbol{s}\mid\boldsymbol{x})\); change-of-variables details are deferred to Appendix~\ref{app:derivation}.

\subsection{Cost-Aware Policy Optimization (CAPO)}
\label{sec:capo}
A flat penalty on \(C(\boldsymbol{s})\) drives the policy toward uniformly minimal budgets regardless of question difficulty: any cost reduction is rewarded identically whether it preserves or destroys the answer. CAPO replaces this raw penalty with a shaped signal that couples cost awareness to answer correctness.

\textbf{Compute metric.}
For the resize operator, if frame \(f_t \in \mathbb{R}^{3 \times H_t \times W_t}\) is rescaled by \(s_t\), its visual token count satisfies
\(
n_t(s_t)\propto \lceil s_t H_t / P \rceil \lceil s_t W_t / P \rceil
\)
for patch size \(P\). We measure physical compute by the \emph{token retention ratio}
\begin{equation}
\rho(\boldsymbol{s}) = \frac{\sum_{t=1}^{T} n_t(s_t)}{\sum_{t=1}^{T} n_t(1)}
\approx \frac{\sum_{t=1}^{T} s_t^2 H_tW_t}{\sum_{t=1}^{T} H_tW_t}.
\end{equation}
Because frames are normalized to a common base resolution before allocation, \(\rho(\boldsymbol{s})\) reduces to the mean quadratic scale.

\textbf{Proxy cost.}
The quadratic dependence of \(\rho\) on \(s_t\) amplifies a few large allocations and inflates gradient variance. We therefore optimize against a smoother proxy
\begin{equation}
c(\boldsymbol{s})=\frac{\bar{s}-s_{\min}}{s_{\max}-s_{\min}}, \qquad \bar{s}=\frac{1}{T}\sum_{t=1}^{T}s_t,
\end{equation}
used only inside the optimizer; the quadratic \(\rho(\boldsymbol{s})\) remains the efficiency metric reported in all experiments.

\textbf{Base advantage.}
For each prompt \(\boldsymbol{x}\), let \(R^{\text{task}}_{m,n}\) denote the scalar task reward of rollout \((m,n)\) (defined in Appendix~\ref{appendix:reward_design}), \(A^{\text{base}}_{m,n}\) the corresponding GRPO group-normalized advantage, \(c_m=c(\boldsymbol{s}_m)\) the proxy cost of allocation \(m\), and \(u_{m,n}\in\{0,1\}\) a binary correctness indicator (exact-match for QA; thresholded success for continuous metrics).

\textbf{Dynamic cost pivot.} CAPO's key ingredient is a decision boundary that determines whether a sampled cost $c_m$ should be rewarded for efficiency or penalized for being expensive. A fixed target budget \(\tau_{\text{fix}}\) ignores the policy's current state, causing unstable updates when the model operates far from this target. Conversely, using only the prompt-local mean \(\bar{c}_{\text{group}} = \frac{1}{M}\sum_{m=1}^{M} c_m\) encourages relative efficiency but cannot anchor the policy to the absolute compression goal. CAPO interpolates between both via a dynamic pivot:
\begin{equation}
\tau_{\text{dyn}} = \kappa_{\text{mix}}\,\bar{c}_{\text{group}} + (1-\kappa_{\text{mix}})\,\tau_{\text{fix}},
\label{eq:capo_pivot}
\end{equation}
where \(\kappa_{\text{mix}}\in[0,1]\). The group mean provides a state-aware baseline for local cost comparisons, while \(\tau_{\text{fix}}\) continuously steers the policy toward the global compression target.

\textbf{Asymmetric shaping.} With $\tau_{\text{dyn}}$ as pivot, CAPO applies a correctness-dependent cost signal:
\begin{equation}
S_{m,n} =
\begin{cases}
\lambda_{+}\, \sigma\!\left(\dfrac{\tau_{\text{dyn}} - c_m}{\tau_{\text{s}}}\right) & \text{if } u_{m,n}=1, \\[6pt]
-\lambda_{-}\, \sigma\!\left(\dfrac{c_m - \tau_{\text{dyn}}}{\tau_{\text{s}}}\right) & \text{if } u_{m,n}=0,
\end{cases},
\label{eq:capo_shaping}
\end{equation}
with $\lambda_{-}>\lambda_{+}>0$. A correct rollout at below-pivot cost receives a moderate bonus; an incorrect rollout at above-pivot cost receives a stronger penalty. The sigmoid temperature \(\tau_{\text{s}}\) smooths the transition near the boundary. This asymmetry is the mechanism that prevents collapse: reducing cost on correct answers is encouraged, but reducing cost at the expense of correctness is strictly penalized.

\textbf{Final CAPO advantage.} The shaped signal is combined with the base advantage:
\begin{equation}
\tilde{A}_{m,n}=A^{\text{base}}_{m,n} + \lambda_{\text{capo}}\, S_{m,n} - \gamma\, c_m,
\end{equation}
where \(\lambda_{\text{capo}}>0\) scales the shaping term and \(\gamma\geq 0\) applies a residual global cost pressure. The final advantage applies a floor on correct rollouts:
\begin{equation}
A_{m,n} =
\begin{cases}
\max\!\left(\tilde{A}_{m,n},\varepsilon_{+}\right) & \text{if } u_{m,n}=1, \\[4pt]
\tilde{A}_{m,n} & \text{if } u_{m,n}=0,
\end{cases}
\label{eq:capo_adv}
\end{equation}
ensuring that correct, low-cost rollouts always retain a positive learning signal (\(\varepsilon_{+}>0\)).

\subsection{Regularization and Training Objective}
\label{sec:training}
CAPO stabilizes the global accuracy--cost trade-off but does not break the symmetry among visually redundant neighbors: the optimizer can assign identical scales to adjacent near-duplicate frames without penalty. We introduce two regularizers to resolve this.

\textbf{Temporal similarity loss ($\mathcal{L}_{\text{sim}}$).}
Reusing the coarse features \(\boldsymbol{f}_t\) from Sec.~\ref{sec:allocator}, we penalize redundant joint high-budget allocation on similar adjacent pairs:
\begin{equation}
\mathcal{L}_{\text{sim}} =
\frac{1}{T-1}\sum_{t=1}^{T-1}
w_t \cdot
\max\!\left(0,\log s_t + \log s_{t+1} + \eta_{\text{sim}}\right),
\end{equation}
where the similarity-gated weight
\begin{equation}
w_t=\sigma\!\left(\frac{\cos(\boldsymbol{f}_t,\boldsymbol{f}_{t+1})-\tau_{\text{sim}}}{\gamma_{\text{sim}}}\right)
\end{equation}
activates only when adjacent frames exceed a cosine-similarity threshold \(\tau_{\text{sim}}\in(0,1)\), with temperature \(\gamma_{\text{sim}}\). No penalty is incurred when \(s_t s_{t+1}\le e^{-\eta_{\text{sim}}}\).

\textbf{Concentration loss ($\mathcal{L}_{\text{con}}$).}
To prevent the Beta distributions from collapsing to near-deterministic spikes, we softly cap the total concentration at \(\kappa_{\max}>0\):
\begin{equation}
\mathcal{L}_{\text{con}} =
\frac{1}{T}\sum_{t=1}^{T}
\max\!\left(0,\alpha_t+\beta_t-\kappa_{\max}\right).
\end{equation}
Together, \(\mathcal{L}_{\text{sim}}\) forces differentiated allocation across redundant neighbors, while \(\mathcal{L}_{\text{con}}\) preserves sufficient stochasticity for continued exploration.

\textbf{Training procedure.}
We optimize both policies in a single GRPO-style loop~\citep{zheng2025group,yu2025dapo}. For each prompt \(\boldsymbol{x}\), the Allocator draws \(M\) allocation trajectories \(\boldsymbol{s}_{1:M}\); each transformed input \(\tilde{\boldsymbol{x}}^{(m)}\) produces \(N\) response rollouts from the backbone. CAPO computes per-rollout advantages \(A_{m,n}\), which serve as the shared learning signal for both policies (Appendix~\ref{app:derivation}).

\textbf{Allocator objective.}
Rollout advantages are aggregated per allocation,
\(
A^{\text{CAPO}}_m = \frac{1}{N}\sum_n A_{m,n},
\)
and used in a per-frame PPO surrogate:
\begin{equation}
\mathcal{L}_\theta = -\frac{1}{MT}\sum_{m=1}^{M}\sum_{t=1}^{T} \min\!\Big(r_{\theta,t}^{(m)}\, A^{\text{CAPO}}_m,\operatorname{clip}\!\big(r_{\theta,t}^{(m)},1{-}\varepsilon,1{+}\varepsilon\big)\,A^{\text{CAPO}}_m\Big),
\label{eq:loss_theta}
\end{equation}
where the per-frame importance ratio is
\begin{equation}
r_{\theta,t}^{(m)} =
\frac{q_\theta(a_t^{(m)}\mid\boldsymbol{x})}
{q_{\theta_{\text{old}}}(a_t^{(m)}\mid\boldsymbol{x})}.
\end{equation}
The full Allocator loss combines the policy gradient with both regularizers:
\begin{equation}
\mathcal{L}_{\text{alloc}}=
\mathcal{L}_{\theta}
+ \lambda_{\text{sim}}\mathcal{L}_{\text{sim}}
+ \lambda_{\text{con}}\mathcal{L}_{\text{con}}.
\label{eq:total_loss}
\end{equation}

\textbf{Backbone update.}
Conditioned on the sampled allocations, the backbone is updated with the standard token-level PPO surrogate:
\begin{equation}
\mathcal{L}_\phi = -\frac{1}{MN}\sum_{m=1}^{M}\sum_{n=1}^{N}\frac{1}{L_{m,n}}\sum_{j=1}^{L_{m,n}} \min\!\Big(r_{\phi,j}^{(m,n)}\, A_{m,n},\,\operatorname{clip}\!\big(r_{\phi,j}^{(m,n)},1{-}\varepsilon,1{+}\varepsilon\big)\,A_{m,n}\Big),
\label{eq:loss_phi}
\end{equation}
where $L_{m,n}$ is the rollout length and
\begin{equation}
r_{\phi,j}^{(m,n)}
=
\frac{\pi_\phi(y_j^{(m,n)}\mid y_{<j}^{(m,n)},\tilde{\boldsymbol{x}}^{(m)})}
{\pi_{\phi_{\text{old}}}(y_j^{(m,n)}\mid y_{<j}^{(m,n)},\tilde{\boldsymbol{x}}^{(m)})}.
\end{equation}
The two objectives are fully decoupled: \(\mathcal{L}_{\text{alloc}}\) updates only \(\theta\) while \(\mathcal{L}_\phi\) updates only \(\phi\), so either component can be frozen or activated independently. When the backbone is held fixed, only the Allocator is trained; when both are active, the two losses are optimized alternately within the same training loop. Algorithm~\ref{alg:resadapt} summarizes one iteration.

\begin{algorithm}[t]
\caption{\our{} Training (One Iteration)}
\label{alg:resadapt}
\begin{algorithmic}[1]
\Require Prompt batch $\{\boldsymbol{x}_i\}$, Allocator $\pi_\theta$, Backbone $\pi_\phi$, operator $\mathcal{O}$
\For{each prompt $\boldsymbol{x}$}
    \State Sample $M$ allocations: $\boldsymbol{s}_m \sim \pi_\theta(\cdot\mid\boldsymbol{x})$ via Beta policy (Eq.~\ref{eq:scale_map})
    \For{each allocation $m = 1, \ldots, M$}
        \State Apply operator: $\tilde{\boldsymbol{x}}^{(m)} = (\boldsymbol{q}, \{\mathcal{O}(f_t, s_t^{(m)})\}_{t=1}^{T})$
        \State Sample $N$ rollouts: $\boldsymbol{y}_{m,n} \sim \pi_\phi(\cdot\mid\tilde{\boldsymbol{x}}^{(m)})$
        \State Compute task reward $R^{\text{task}}_{m,n}$
    \EndFor
    \State Compute CAPO advantages $A_{m,n}$ (Eqs.~\ref{eq:capo_shaping}--\ref{eq:capo_adv})
    \State Aggregate per-allocation: $A^{\text{CAPO}}_m = \frac{1}{N}\sum_n A_{m,n}$
\EndFor
\State Update Allocator: minimize $\mathcal{L}_{\text{alloc}}$ (Eq.~\ref{eq:total_loss})
\State Update Backbone: minimize $\mathcal{L}_\phi$ (Eq.~\ref{eq:loss_phi}) \Comment{Omit if frozen}
\end{algorithmic}
\end{algorithm}

\section{Experiments}
\label{main:Experiments}

\begin{table}[thbp]
    \centering
    \footnotesize
    \caption{\textbf{Video QA results} across two backbones (Qwen2.5-VL-7B, Qwen3-VL-8B) and two temporal horizons (32/128 frames). Retention ratio $R$ reflects visual token count; \textit{Reasoning} (\cmark/\xmark) indicates chain-of-thought use; \textbf{bold} marks the best result per group.}
    \resizebox{\linewidth}{!}{
    \begin{tabular}{clcccccccc}
    \toprule
     \multirow{2}{*}{\textbf{Backbone}} & \multirow{2}{*}{\textbf{Method}} & \multirow{2}{*}{\textbf{\makecell{Retention \\ Ratio $R$}}} & \multirow{2}{*}{\textbf{\makecell{Reasoning}}} & \multicolumn{4}{c}{\textbf{Video Perception Benchmark}} & \multicolumn{2}{c}{\textbf{Video Reasoning Benchmark}} \\
    \cmidrule{5-8}\cmidrule{9-10}
                           & & & & VideoMME & LongVideoBench & MMVU & MLVU & VideoMMMU & LVBench  \\
    \midrule
    \multirow{30}{*}{\rotatebox{90}{\textbf{Qwen2.5-VL-7B}}} 
    & \multicolumn{9}{c}{\textit{32 Frames}} \\
    \cmidrule{2-10}
    & Vanilla & 100\% & \xmark & 62.0 & 58.9 & 52.7 & 63.1 & 49.6 & 38.6 \\
    \cdashline{2-10}
    & Random Drop & 25.0\% & \xmark & 58.9 & 57.8 & 49.6 & 58.3 & 45.3 & 36.7 \\
    & ToMe~\citep{bolya2022token} & 25.0\% & \xmark & 58.7 & 58.0 & 51.0 & 58.7 & 41.8 & 37.7 \\
    & VisionZip~\citep{visionzip} & 25.0\% & \xmark & 59.4 & 57.1 & 49.8 & 57.9 & 42.4 & 36.5 \\
    & FlashVid~\citep{flashvid} & 29.3\% & \xmark & 60.2 & \textbf{58.6} & 51.1 & 59.2 & 46.3 & 36.9 \\
    & FixedScale & 25.0\% & \xmark & 60.0 & 56.8 & 51.2 & 59.8 & 46.7 & 37.3 \\
    & \cellcolor{metabg}\textbf{\our{} (Ours)} & \cellcolor{metabg}23.8\% & \cellcolor{metabg}\xmark & \cellcolor{metabg}\textbf{60.3} & \cellcolor{metabg}58.2 & \cellcolor{metabg}\textbf{51.9} & \cellcolor{metabg}\textbf{60.1} & \cellcolor{metabg}\textbf{48.8} & \cellcolor{metabg}\textbf{37.9} \\
    \cdashline{2-10}
    & Random Drop & 10.0\% & \xmark & 56.1 & 55.6 & 47.1 & 56.5 & 39.8 & 35.2 \\
    & ToMe~\citep{bolya2022token} & 10.0\% & \xmark & 56.4 & 55.2 & 48.9 & 58.0 & 39.2 & 33.6 \\
    & VisionZip~\citep{visionzip} & 10.0\% & \xmark & 55.5 & 54.5 & 47.6 & 57.3 & 39.1 & 35.3 \\
    & FlashVid~\citep{flashvid} & 10.4\% & \xmark & 57.9 & \textbf{56.8} & 47.9 & 57.7 & 39.4 & \textbf{36.5} \\
    & FixedScale & 12.3\% & \xmark & 58.0 & 55.1 & 47.7 & 57.5 & 44.3 & 35.4 \\
    & \cellcolor{metabg}\textbf{\our{} (Ours)} & \cellcolor{metabg}11.4\% & \cellcolor{metabg}\xmark & \cellcolor{metabg}\textbf{59.4} & \cellcolor{metabg}55.4 & \cellcolor{metabg}\textbf{49.2} & \cellcolor{metabg}\textbf{58.4} & \cellcolor{metabg}\textbf{45.7} & \cellcolor{metabg}35.9 \\
    \cdashline{2-10}
    & VideoAuto-R1~\citep{liu2026videoauto} & 100\% & \cmark & 63.2 & 58.9 & 55.0 & 60.0 & 53.6 & 41.5 \\
    & \cellcolor{metabg} + \textbf{\our{} (Ours)} & \cellcolor{metabg}23.8\% & \cellcolor{metabg}\cmark & \cellcolor{metabg}60.4 & \cellcolor{metabg}57.1 & \cellcolor{metabg}53.2 & \cellcolor{metabg}61.1 & \cellcolor{metabg}51.2 & \cellcolor{metabg}38.7 \\
    & \cellcolor{metabg} + \textbf{\our{} (Ours)} & \cellcolor{metabg}11.4\% & \cellcolor{metabg}\cmark & \cellcolor{metabg}59.3 & \cellcolor{metabg}56.3 & \cellcolor{metabg}51.8 & \cellcolor{metabg}59.3 & \cellcolor{metabg}49.1 & \cellcolor{metabg}36.7 \\
    \cmidrule{2-10}
    & \multicolumn{9}{c}{\textit{128 Frames}} \\
    \cmidrule{2-10}
    & Vanilla & 100\% & \xmark & 65.3 & 60.3 & 53.1 & 66.5 & 47.9 & 42.0 \\
    \cdashline{2-10}
    & Random Drop & 25.0\% & \xmark & 64.9 & 61.2 & 50.8 & 64.8 & 48.1 & 41.3 \\
    & ToMe~\citep{bolya2022token} & 25.0\% & \xmark & 65.1 & \textbf{61.6} & 51.9 & 63.1 & 46.6 & \textbf{42.1} \\
    & VisionZip~\citep{visionzip} & 25.0\% & \xmark & 64.8 & 61.3 & 51.1 & 64.5 & 47.3 & 41.5 \\
    & \cellcolor{metabg}\textbf{\our{} (Ours)} & \cellcolor{metabg}22.9\% & \cellcolor{metabg}\xmark & \cellcolor{metabg}\textbf{65.6} & \cellcolor{metabg}60.2 & \cellcolor{metabg}\textbf{52.8} & \cellcolor{metabg}\textbf{65.9} & \cellcolor{metabg}\textbf{51.1} & \cellcolor{metabg}\textbf{42.1} \\
    \cdashline{2-10}
    & Random Drop & 10.0\% & \xmark & 63.0 & 59.0 & 45.8 & 63.4 & 46.7 & 38.0 \\
    & ToMe~\citep{bolya2022token} & 10.0\% & \xmark & 60.6 & 56.3 & 44.2 & 63.5 & 41.8 & 39.5 \\
    & VisionZip~\citep{visionzip} & 10.0\% & \xmark & 61.8 & 56.1 & 44.8 & 63.2 & 42.1 & 38.2 \\
    & FixedScale & 12.3\% & \xmark & \textbf{64.1} & \textbf{60.9} & \textbf{49.6} & \textbf{64.5} & 46.9 & \textbf{40.3} \\
    & \cellcolor{metabg}\textbf{\our{} (Ours)} & \cellcolor{metabg}11.1\% & \cellcolor{metabg}\xmark & \cellcolor{metabg}63.8 & \cellcolor{metabg}58.6 & \cellcolor{metabg}49.0 & \cellcolor{metabg}64.3 & \cellcolor{metabg}\textbf{49.2} & \cellcolor{metabg}39.9 \\
    \cdashline{2-10}
    & VideoAuto-R1~\citep{liu2026videoauto} & 100\% & \cmark & 64.7 & 59.1 & 56.7 & 65.1 & 52.2 & 41.2 \\
    & \cellcolor{metabg} + \textbf{\our{} (Ours)} & \cellcolor{metabg}23.8\% & \cellcolor{metabg}\cmark & \cellcolor{metabg}66.2 & \cellcolor{metabg}60.2 & \cellcolor{metabg}53.5 & \cellcolor{metabg}66.0 & \cellcolor{metabg}52.6 & \cellcolor{metabg}41.8 \\
    & \cellcolor{metabg} + \textbf{\our{} (Ours)} & \cellcolor{metabg}11.4\% & \cellcolor{metabg}\cmark & \cellcolor{metabg}64.7 & \cellcolor{metabg}57.8 & \cellcolor{metabg}52.4 & \cellcolor{metabg}64.6 & \cellcolor{metabg}51.3 & \cellcolor{metabg}39.5 \\
    
    \midrule
    \multirow{24}{*}{\rotatebox{90}{\textbf{Qwen3-VL-8B}}} 
    & \multicolumn{9}{c}{\textit{32 Frames}} \\
    \cmidrule{2-10}
    & Vanilla & 100\% & \xmark & 65.0 & 58.6 & 57.5 & 64.0 & 60.8 & 40.2 \\
    \cdashline{2-10}
    & Random Drop & 25.0\% & \xmark & 61.3 & 58.4 & \textbf{57.1} & 60.2 & 53.4 & 37.8 \\
    & ToMe~\citep{bolya2022token} & 25.0\% & \xmark & 62.4 & 57.4 & 56.0 & 60.8 & 49.1 & 36.4 \\
    & VisionZip~\citep{visionzip} & 25.0\% & \xmark & 61.8 & 57.2 & 54.4 & 60.6 & 51.5 & 37.3 \\
    & FlashVid~\citep{flashvid} & 30.0\% & \xmark & \textbf{63.9} & \textbf{59.0} & 54.8 & \textbf{61.9} & 55.1 & \textbf{38.5} \\
    & \cellcolor{metabg}\textbf{\our{} (Ours)} & \cellcolor{metabg}23.8\% & \cellcolor{metabg}\xmark & \cellcolor{metabg}62.6 & \cellcolor{metabg}57.5 & \cellcolor{metabg}55.3 & \cellcolor{metabg}61.0 & \cellcolor{metabg}\textbf{58.4} & \cellcolor{metabg}\textbf{38.5} \\
    \cdashline{2-10}
    & Random Drop & 10.0\% & \xmark & 58.8 & 54.7 & 53.2 & 56.6 & 47.1 & 35.5 \\
    & ToMe~\citep{bolya2022token} & 10.0\% & \xmark & 59.2 & 55.5 & 53.1 & 58.5 & 42.7 & 35.8 \\
    & VisionZip~\citep{visionzip} & 10.0\% & \xmark & 59.9 & 55.4 & 53.7 & 58.8 & 45.8 & 35.4 \\
    & FlashVid~\citep{flashvid} & 12.2\% & \xmark & \textbf{61.0} & \textbf{57.1} & \textbf{54.8} & 59.1 & 47.8 & 37.1 \\
    & FixedScale & 12.3\% & \xmark & 60.8 & 54.9 & 53.8 & 58.4 & 52.6 & 37.1 \\
    & \cellcolor{metabg}\textbf{\our{} (Ours)} & \cellcolor{metabg}11.4\% & \cellcolor{metabg}\xmark & \cellcolor{metabg}60.7 & \cellcolor{metabg}56.6 & \cellcolor{metabg}54.6 & \cellcolor{metabg}\textbf{59.6} & \cellcolor{metabg}\textbf{56.1} & \cellcolor{metabg}\textbf{37.3} \\
    \cmidrule{2-10}
    & \multicolumn{9}{c}{\textit{128 Frames}} \\
    \cmidrule{2-10}
    & Vanilla & 100\% & \xmark & 69.4 & 64.3 & 58.5 & 72.7 & 63.0 & 45.7 \\
    \cdashline{2-10}
    & Random Drop & 25.0\% & \xmark & 67.2 & 61.3 & \textbf{56.8} & 67.4 & 55.3 & 42.4 \\
    & ToMe~\citep{bolya2022token} & 25.0\% & \xmark & 67.2 & \textbf{62.0} & 55.9 & 70.4 & 53.5 & 43.1 \\
    & VisionZip~\citep{visionzip} & 25.0\% & \xmark & 67.1 & 61.3 & 55.7 & 69.2 & 56.8 & 41.2 \\
    & \cellcolor{metabg}\textbf{\our{} (Ours)} & \cellcolor{metabg}22.9\% & \cellcolor{metabg}\xmark & \cellcolor{metabg}\textbf{67.4} & \cellcolor{metabg}61.9 & \cellcolor{metabg}56.3 & \cellcolor{metabg}\textbf{70.8} & \cellcolor{metabg}\textbf{59.6} & \cellcolor{metabg}\textbf{43.3} \\
    \cdashline{2-10}
    & Random Drop & 10.0\% & \xmark & 64.1 & 58.3 & \textbf{55.4} & 62.4 & 55.5 & 38.8 \\
    & ToMe~\citep{bolya2022token} & 10.0\% & \xmark & 64.7 & 58.6 & 55.1 & 67.3 & 46.3 & 40.5 \\
    & VisionZip~\citep{visionzip} & 10.0\% & \xmark & 64.2 & 59.1 & 54.2 & 66.8 & 47.6 & 39.4 \\
    & FixedScale & 12.3\% & \xmark & 66.7 & 59.5 & 54.4 & 67.7 & 56.3 & 41.7 \\
    & \cellcolor{metabg}\textbf{\our{} (Ours)} & \cellcolor{metabg}11.1\% & \cellcolor{metabg}\xmark & \cellcolor{metabg}\textbf{66.8} & \cellcolor{metabg}\textbf{60.2} & \cellcolor{metabg}\textbf{55.4} & \cellcolor{metabg}\textbf{69.4} & \cellcolor{metabg}\textbf{58.2} & \cellcolor{metabg}\textbf{42.6} \\
    
    \bottomrule
    \end{tabular}
    }
    \label{table:benchmark_qa}
\end{table}

\subsection{Setup}
\label{sec:exp_details}

\noindent \textbf{Implementation.}
The Allocator \(\pi_\theta\) uses the SmolVLM architecture~\citep{marafioti2025smolvlm} for high-throughput front-end prediction. Throughout, we instantiate input-side allocation with \emph{resize}, so the learned allocations are realized as per-frame resize factors. We train the Allocator on Qwen2.5-VL-7B-Instruct~\citep{bai2025qwen2} and additionally test transfer to Qwen3-VL-8B-Instruct~\citep{bai2025qwen3}. We report two settings: \textbf{\our{}-RL}, obtained by jointly updating the Allocator and the backbone, and \textbf{\our{}}, which directly reuses the trained Allocator with a frozen backbone to evaluate plug-and-play generalization. Resize is used during training because it provides the continuous action space required by our optimizer; thresholded frame selection is treated only as the conceptual zero-budget limit of the same pre-encoding interface. Full hyperparameters, hardware, prompts, and reward definitions are deferred to Appendix~\ref{appendix:implementation_details}.

\noindent \textbf{Baselines.}
We compare against three classes of methods: \textbf{heuristic baselines} (Random Drop, FixedScale), \textbf{model-side compression} (ToMe~\citep{bolya2022token}, FlashVid~\citep{flashvid}, VisionZip~\citep{visionzip}), and \textbf{reasoning-time inference augmentation} (VideoAuto-R1~\citep{liu2026videoauto}). We use visual-token retention ratio \(R\)\footnote{$R$ corresponds to $\rho(\boldsymbol{s})$ in Sec.~\ref{sec:capo}; we use $R$ in tables for compactness.} as the primary budget descriptor and report the exact retained budget for every method. For reasoning-time baselines, \(R\) measures only visual encoder tokens; unless latency is reported separately, these comparisons should therefore be read as visual-budget comparisons rather than total-inference-budget matches. Because several baselines admit only discrete operating points, some comparisons are only approximately budget matched and should be interpreted relative to the explicit trade-offs shown in each table.

\noindent \textbf{Benchmarks.}
For \textit{video QA}, we report results on VideoMME~\citep{fu2025video}, LongVideoBench~\citep{wu2024longvideobench}, MMVU~\citep{zhao2025mmvu}, MLVU~\citep{mlvu}, VideoMMMU~\citep{hu2025video}, and LVBench~\citep{wang2025lvbench}. For \textit{temporal grounding}, we report Recall@$\{0.3,0.5,0.7\}$ and mIoU on Charades-STA~\citep{gao2017tall} and ActivityNet~\citep{caba2015activitynet}, plus grounding QA on NExT-GQA~\citep{xiao2024can}. For \textit{image understanding}, we evaluate on MathVista~\citep{lu2023mathvista}, MMMU~\citep{yue2024mmmu}, OCRBench~\citep{liu2024ocrbench}, ChartQA~\citep{masry2022chartqa}, AI2D~\citep{kembhavi2016diagram}, and TextVQA~\citep{singh2019towards}. Unless stated otherwise, figures and analyses use Qwen2.5-VL-7B with 32 input frames. All evaluations use lmms-eval~\citep{zhang2024lmmsevalrealitycheckevaluation}; the exact token budgets and decoding limits are reported in Appendix~\ref{appendix:implementation_details}.

\begin{figure*}[t]
    \centering
    \includegraphics[width=\linewidth]{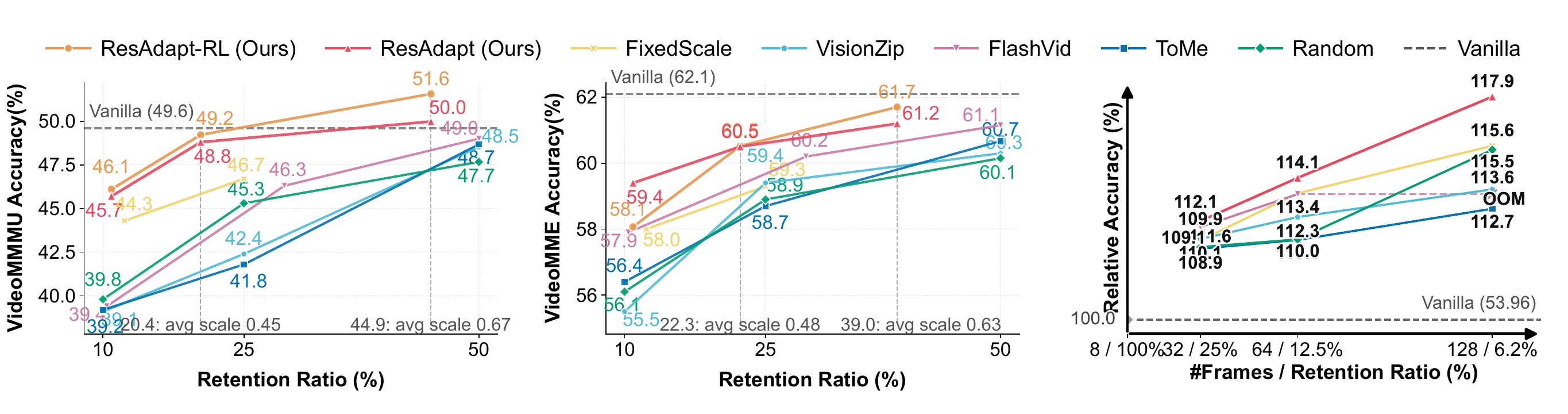}
    \caption{\textbf{Efficiency--accuracy trade-offs and temporal reallocation.} \textbf{(a,b)} VideoMMMU and VideoMME versus visual-token retention ratio $R$. \our{} is on or near the Pareto frontier, with the clearest advantage on reasoning-heavy settings at low retention. \textbf{(c)} Relative gain from trading spatial resolution for temporal coverage under a fixed 8-frame-equivalent budget.}
    \label{fig:pareto_frontier}
\end{figure*}

\begin{table*}[thbp]
    \centering
    \footnotesize
    \caption{\textbf{Temporal grounding results} across two backbones and two temporal horizons. Notation follows Table~\ref{table:benchmark_qa}. See Sec.~\ref{sec:temporal_results} for analysis of compression sensitivity.}
    \resizebox{\linewidth}{!}{
    \begin{tabular}{clcccccccccccccc}
    \toprule
    \multirow{3}{*}{\textbf{Backbone}} & \multirow{3}{*}{\textbf{Method}}& \multirow{3}{*}{\textbf{\makecell{Retention \\ Ratio $R$}}} & \multirow{3}{*}{\textbf{\makecell{Reasoning}}}  & \multicolumn{12}{c}{\textbf{Temporal Grounding Benchmark}} \\
    \cmidrule{5-16}
                                       &  &  &  & \multicolumn{4}{c}{Charades-STA} & & \multicolumn{4}{c}{ActivityNet} & & \multicolumn{2}{c}{NExT-GQA} \\
                                       &  &  &  & 0.3 & 0.5 & 0.7 & mIoU & & 0.3 & 0.5 & 0.7 & mIoU & & Acc & mIoU\\
    \midrule
    \multirow{28}{*}{\rotatebox{90}{\textbf{Qwen2.5-VL-7B}}}
     & \multicolumn{15}{c}{\textit{32 Frames}} \\
    \cmidrule{2-16}
    & Vanilla & 100\% & \xmark & 71.0 & 51.4 & 26.0 & 47.3 & & 30.4 & 18.0 & 8.9 & 22.6 & & 78.9 & 28.0 \\
    \cdashline{2-16}
    & Random Drop & 25.0\% & \xmark & 39.4 & 23.2 & 11.0 & 25.7 & & 15.2 & 8.1 & 3.7 & 11.7 & & 77.5 & 16.6 \\
    & ToMe~\citep{bolya2022token} & 25.0\% & \xmark & 39.5 & 23.9 & 11.4 & 26.0 & & 16.0 & 8.4 & 4.0 & 12.1 & & 77.8 & 16.3 \\
    & FlashVid~\citep{flashvid} & 31.3\% & \xmark & 40.7 & 24.2 & 11.3 & 26.6 & & 15.8 & 8.4 & 3.8 & 12.0 & & \textbf{78.1} & 16.5 \\
    & FixedScale & 25.0\% & \xmark & 36.7 & 24.7 & 12.3 & 24.9 & & 18.6 & 9.4 & 4.3 & 14.1 & & 77.7 & 12.3 \\
    & \cellcolor{metabg}\textbf{\our{} (Ours)} & \cellcolor{metabg}16.2\% & \cellcolor{metabg}\xmark & \cellcolor{metabg}\textbf{53.8} & \cellcolor{metabg}\textbf{34.8} & \cellcolor{metabg}\textbf{17.0} & \cellcolor{metabg}\textbf{35.6} & \cellcolor{metabg} & \cellcolor{metabg}\textbf{19.8} & \cellcolor{metabg}\textbf{10.8} & \cellcolor{metabg}\textbf{5.2} & \cellcolor{metabg}\textbf{15.3} & \cellcolor{metabg} & \cellcolor{metabg}76.6 & \cellcolor{metabg}\textbf{23.2} \\
    \cdashline{2-16}
    & Random Drop & 10.0\% & \xmark & 36.9 & 23.2 & 11.6 & 24.6 & & 14.3 & 7.5 & 3.6 & 11.1 & & 76.3 & 15.4 \\
    & ToMe~\citep{bolya2022token} & 10.0\% & \xmark & 41.3 & 26.9 & 14.1 & 27.4 & & 16.0 & 8.4 & \textbf{4.0} & 12.2 & & 77.3 & 15.7 \\
    & FlashVid~\citep{flashvid} & 12.6\% & \xmark & 38.2 & 22.9 & 11.1 & 25.1 & & 15.4 & 8.1 & 3.7 & 11.8 & & \textbf{77.4} & 16.1 \\
    & FixedScale & 12.3\% & \xmark & \textbf{48.0} & \textbf{31.5} & \textbf{15.4} & \textbf{32.0} & & \textbf{17.5} & \textbf{8.9} & \textbf{4.0} & \textbf{13.3} & & 76.1 & 13.7 \\
    & FixedScale & 6.3\% & \xmark & 39.9 & 26.8 & 13.3 & 26.7 & & 15.2 & 8.1 & 3.9 & 11.9 & & 74.1 & 15.4 \\
    & \cellcolor{metabg}\textbf{\our{} (Ours)} & \cellcolor{metabg}6.8\% & \cellcolor{metabg}\xmark & \cellcolor{metabg}41.0 & \cellcolor{metabg}27.8 & \cellcolor{metabg}14.0 & \cellcolor{metabg}27.2 & \cellcolor{metabg} & \cellcolor{metabg}16.3 & \cellcolor{metabg}8.5 & \cellcolor{metabg}3.9 & \cellcolor{metabg}12.5 & \cellcolor{metabg} & \cellcolor{metabg}74.3 & \cellcolor{metabg}\textbf{20.4} \\
    \cdashline{2-16}
    & VideoAuto-R1~\citep{liu2026videoauto} & 100\% & \cmark & 60.0 & 48.3 & 27.2 & 41.5 & & 50.8 & 34.1 & 17.4 & 34.4 & & 73.6 & 33.8 \\
    & \cellcolor{metabg} + \textbf{\our{} (Ours)} & \cellcolor{metabg}6.8\% & \cellcolor{metabg}\cmark & \cellcolor{metabg}43.5 & \cellcolor{metabg}30.1 & \cellcolor{metabg}15.8 & \cellcolor{metabg}30.0 & \cellcolor{metabg} & \cellcolor{metabg}35.4 & \cellcolor{metabg}21.5 & \cellcolor{metabg}10.0 & \cellcolor{metabg}24.4 & \cellcolor{metabg} & \cellcolor{metabg}74.7 & \cellcolor{metabg}24.7 \\
    \cmidrule{2-16}
    & \multicolumn{15}{c}{\textit{128 Frames}} \\
    \cmidrule{2-16}
    & Vanilla & 100\% & \xmark & 77.5 & 60.3 & 34.1 & 52.8 & & 47.9 & 30.9 & 17.5 & 34.4 & & 79.8 & 29.9 \\
    \cdashline{2-16}
    & Random Drop & 25.0\% & \xmark & 32.3 & 19.6 & 7.9 & 20.7 & & 26.7 & 13.9 & 6.3 & 18.8 & & \textbf{80.3} & 10.7 \\
    & ToMe~\citep{bolya2022token} & 25.0\% & \xmark & 32.4 & 19.8 & 7.9 & 20.7 & & 27.2 & 14.4 & 6.4 & 19.1 & & \textbf{80.3} & 10.9 \\
    & \cellcolor{metabg}\textbf{\our{} (Ours)} & \cellcolor{metabg}16.1\% & \cellcolor{metabg}\xmark & \cellcolor{metabg}\textbf{63.5} & \cellcolor{metabg}\textbf{43.6} & \cellcolor{metabg}\textbf{21.3} & \cellcolor{metabg}\textbf{42.0} & \cellcolor{metabg} & \cellcolor{metabg}\textbf{33.1} & \cellcolor{metabg}\textbf{19.3} & \cellcolor{metabg}\textbf{10.2} & \cellcolor{metabg}\textbf{24.3} & \cellcolor{metabg} & \cellcolor{metabg}78.1 & \cellcolor{metabg}\textbf{27.2} \\
    \cdashline{2-16}
    & Random Drop & 10.0\% & \xmark & 37.8 & 23.8 & 11.2 & 24.7 & & 23.8 & 12.0 & 5.3 & 17.0 & & \textbf{79.4} & 12.8 \\
    & ToMe~\citep{bolya2022token} & 10.0\% & \xmark & 27.9 & 16.2 & 7.3 & 17.9 & & 22.9 & 11.8 & 5.5 & 16.4 & & 79.1 & 11.1 \\
    & FixedScale & 12.3\% & \xmark & 34.7 & 22.3 & 10.5 & 22.7 & & \textbf{25.0} & \textbf{13.8} & 5.9 & \textbf{18.3} & & 77.9 & 11.3 \\
    & FixedScale & 6.3\% & \xmark & 42.6 & 28.4 & 14.3 & 28.3 & & 22.8 & 12.8 & 5.7 & 17.1 & & 75.7 & 12.9 \\
    & \cellcolor{metabg}\textbf{\our{} (Ours)} & \cellcolor{metabg}6.8\% & \cellcolor{metabg}\xmark & \cellcolor{metabg}\textbf{43.5} & \cellcolor{metabg}\textbf{29.8} & \cellcolor{metabg}\textbf{15.0} & \cellcolor{metabg}\textbf{28.9} & \cellcolor{metabg} & \cellcolor{metabg}23.5 & \cellcolor{metabg}12.9 & \cellcolor{metabg}\textbf{6.1} & \cellcolor{metabg}17.2 & \cellcolor{metabg} & \cellcolor{metabg}76.2 & \cellcolor{metabg}\textbf{23.9} \\
    \cdashline{2-16}
    & VideoAuto-R1~\citep{liu2026videoauto} & 100\% & \cmark & 40.3 & 33.7 & 22.1 & 28.9 & & 49.4 & 34.3 & 18.5 & 33.5 & & 68.0 & 31.0 \\
    & \cellcolor{metabg} + \textbf{\our{} (Ours)} & \cellcolor{metabg}16.1\% & \cellcolor{metabg}\cmark & \cellcolor{metabg}72.8 & \cellcolor{metabg}53.0 & \cellcolor{metabg}27.5 & \cellcolor{metabg}49.1 & \cellcolor{metabg} & \cellcolor{metabg}65.8 & \cellcolor{metabg}44.9 & \cellcolor{metabg}23.8 & \cellcolor{metabg}44.7 & \cellcolor{metabg} & \cellcolor{metabg}79.3 & \cellcolor{metabg}35.3 \\
    & \cellcolor{metabg} + \textbf{\our{} (Ours)} & \cellcolor{metabg}6.8\% & \cellcolor{metabg}\cmark & \cellcolor{metabg}50.1 & \cellcolor{metabg}33.2 & \cellcolor{metabg}16.6 & \cellcolor{metabg}34.2 & \cellcolor{metabg} & \cellcolor{metabg}53.4 & \cellcolor{metabg}34.0 & \cellcolor{metabg}16.4 & \cellcolor{metabg}35.7 & \cellcolor{metabg} & \cellcolor{metabg}76.6 & \cellcolor{metabg}29.4 \\
    \cmidrule{2-16}

    \multirow{23}{*}{\rotatebox{90}{\textbf{Qwen3-VL-8B}}}
     & \multicolumn{15}{c}{\textit{32 Frames}} \\
    \cmidrule{2-16}
    & Vanilla & 100\% & \xmark & 73.0 & 49.0 & 21.4 & 46.4 & & 44.6 & 28.3 & 15.5 & 31.8 & & 78.7 & 34.2 \\
    \cdashline{2-16}
    & Random Drop & 25.0\% & \xmark & 16.2 & 8.6 & 3.8 & 12.1 & & 12.4 & 6.7 & 3.2 & 10.0 & & 77.2 & 15.6 \\
    & ToMe~\citep{bolya2022token} & 25.0\% & \xmark & 68.7 & 42.1 & 17.6 & 43.1 & & 45.9 & 28.8 & 15.6 & 32.6 & & 77.1 & 31.7 \\
    & FlashVid~\citep{flashvid} & 31.3\% & \xmark & \textbf{72.9} & \textbf{52.3} & \textbf{25.1} & \textbf{47.7} & & \textbf{51.9} & \textbf{33.4} & \textbf{19.0} & \textbf{36.8} & & \textbf{77.8} & \textbf{33.9} \\
    & \cellcolor{metabg}\textbf{\our{} (Ours)} & \cellcolor{metabg}16.2\% & \cellcolor{metabg}\xmark & \cellcolor{metabg}64.4 & \cellcolor{metabg}37.3 & \cellcolor{metabg}16.3 & \cellcolor{metabg}39.9 & \cellcolor{metabg} & \cellcolor{metabg}40.0 & \cellcolor{metabg}24.4 & \cellcolor{metabg}13.0 & \cellcolor{metabg}28.5 & \cellcolor{metabg} & \cellcolor{metabg}75.1 & \cellcolor{metabg}30.2 \\
    \cdashline{2-16}
    & Random Drop & 10.0\% & \xmark & 4.1 & 1.8 & 0.7 & 4.4 & & 4.7 & 2.4 & 1.0 & 5.0 & & 74.3 & 11.3 \\
    & ToMe~\citep{bolya2022token} & 10.0\% & \xmark & 67.6 & 39.3 & 16.6 & 41.8 & & 46.3 & 31.0 & \textbf{19.2} & 34.1 & & \textbf{79.2} & \textbf{34.0} \\
    & FlashVid~\citep{flashvid} & 12.6\% & \xmark & \textbf{68.8} & \textbf{46.9} & \textbf{22.9} & \textbf{44.6} & & \textbf{49.9} & \textbf{31.5} & 17.4 & \textbf{35.2} & & 75.6 & 31.8 \\
    & FixedScale & 12.3\% & \xmark & 61.3 & 34.3 & 14.6 & 37.9 & & 39.6 & 24.2 & 13.1 & 28.4 & & 74.2 & 29.9 \\
    & FixedScale & 6.3\% & \xmark & 52.7 & 28.2 & 11.3 & 33.2 & & 37.0 & 22.3 & 12.0 & 27.0 & & 71.5 & 28.0 \\
    & \cellcolor{metabg}\textbf{\our{} (Ours)} & \cellcolor{metabg}6.8\% & \cellcolor{metabg}\xmark & \cellcolor{metabg}53.6 & \cellcolor{metabg}29.0 & \cellcolor{metabg}11.8 & \cellcolor{metabg}33.6 & \cellcolor{metabg} & \cellcolor{metabg}37.5 & \cellcolor{metabg}22.5 & \cellcolor{metabg}12.3 & \cellcolor{metabg}27.2 & \cellcolor{metabg} & \cellcolor{metabg}71.8 & \cellcolor{metabg}28.2 \\
    \cmidrule{2-16}
    & \multicolumn{15}{c}{\textit{128 Frames}} \\
    \cmidrule{2-16}
    & Vanilla & 100\% & \xmark & 72.8 & 46.0 & 20.1 & 45.6 & & 45.8 & 31.1 & 19.2 & 33.9 & & 81.1 & 36.6 \\
    \cdashline{2-16}
    & Random Drop & 25.0\% & \xmark & 41.6 & 25.2 & 10.6 & 27.4 & & 36.1 & 21.1 & 12.7 & 26.3 & & 79.3 & 22.4 \\
    & \cellcolor{metabg}\textbf{\our{} (Ours)} & \cellcolor{metabg}16.1\% & \cellcolor{metabg}\xmark & \cellcolor{metabg}\textbf{64.4} & \cellcolor{metabg}\textbf{37.0} & \cellcolor{metabg}\textbf{15.9} & \cellcolor{metabg}\textbf{39.8} & \cellcolor{metabg} & \cellcolor{metabg}\textbf{40.6} & \cellcolor{metabg}\textbf{26.7} & \cellcolor{metabg}\textbf{15.7} & \cellcolor{metabg}\textbf{30.0} & \cellcolor{metabg} & \cellcolor{metabg}76.8 & \cellcolor{metabg}\textbf{33.3} \\
    \cdashline{2-16}
    & Random Drop & 10.0\% & \xmark & 32.6 & 19.0 & 7.8 & 21.9 & & 33.5 & 18.6 & 11.5 & 24.8 & & 76.9 & 19.9 \\
    & ToMe~\citep{bolya2022token} & 10.0\% & \xmark & 61.6 & 33.8 & 13.3 & \textbf{38.1} & & \textbf{42.4} & \textbf{27.6} & \textbf{16.6} & \textbf{31.4} & & \textbf{77.4} & 31.5 \\
    & FixedScale & 12.3\% & \xmark & \textbf{61.7} & \textbf{34.9} & \textbf{14.7} & \textbf{38.1} & & 39.9 & 26.2 & 15.3 & 29.5 & & 75.4 & 32.6 \\
    & FixedScale & 6.3\% & \xmark & 53.7 & 28.2 & 11.8 & 33.6 & & 37.9 & 24.3 & 14.3 & 28.1 & & 73.0 & 39.1 \\
    & \cellcolor{metabg}\textbf{\our{} (Ours)} & \cellcolor{metabg}6.8\% & \cellcolor{metabg}\xmark & \cellcolor{metabg}54.3 & \cellcolor{metabg}28.0 & \cellcolor{metabg}11.7 & \cellcolor{metabg}33.7 & \cellcolor{metabg} & \cellcolor{metabg}38.3 & \cellcolor{metabg}24.5 & \cellcolor{metabg}14.4 & \cellcolor{metabg}28.4 & \cellcolor{metabg} & \cellcolor{metabg}73.2 & \cellcolor{metabg}\textbf{43.9} \\
    \bottomrule
    \end{tabular}
    }
    \label{table:benchmark_grounding}
\end{table*}

\subsection{Main Results}
\label{sec:pareto}

\subsubsection{Video QA}
\label{sec:video_qa_results}
We first test whether input-side allocation via continuous resizing improves low-budget operating points, especially on reasoning-heavy benchmarks (Table~\ref{table:benchmark_qa}).

\textbf{Disproportionate gains on multi-step reasoning.}
Under aggressive compression (${\sim}$10\% retention), content-agnostic methods inevitably discard sparse but decisive evidence. On Qwen2.5-VL with 32 frames, \our{} achieves \textbf{45.7} on VideoMMMU at 11.4\% retention, substantially outperforming ToMe (\textbf{39.2}), VisionZip (\textbf{39.1}), FlashVid (\textbf{39.4}), and FixedScale (\textbf{44.3}), while maintaining competitiveness on perception-focused benchmarks. The gap is largest on VideoMMMU, the most reasoning-intensive benchmark in the suite, confirming that input-side allocation selectively preserves the sparse visual evidence that multi-step reasoning demands. The transferred Allocator remains robust on Qwen3-VL, securing \textbf{56.1} on VideoMMMU at the same 11.4\% retention, confirming cross-architecture generalizability.

\textbf{Spatial savings reinvested as temporal coverage.}
Extending the context from 32 to 128 frames drastically amplifies this advantage. At 22.9\% retention on Qwen2.5-VL, \our{} reaches \textbf{51.1} on VideoMMMU, exceeding the \textbf{47.9} achieved by the 128-frame uncompressed model while recovering near-optimal perception performance at a fraction of the visual cost. Even at 11.1\% retention, \our{} attains \textbf{49.2}, again surpassing the uncompressed 128-frame score. This validates the central claim of input-side adaptation: spatial budget savings translate directly into temporal headroom, enabling the model to process \textbf{$4{\times}$} more frames without the native-resolution compute penalty (Figure~\ref{fig:pareto_frontier}).

\subsubsection{Temporal Grounding}
\label{sec:temporal_results}

Temporal grounding is far more sensitive to compression than standard QA, since localization depends on fine-grained temporal cues rather than holistic scene understanding. Table~\ref{table:benchmark_grounding} compares methods across comparable operating points.

\textbf{Pre-encoding allocation dominates frame dropping.}
On Qwen2.5-VL (32F), Random Drop, ToMe, FlashVid, and FixedScale severely degrade Charades-STA mIoU from \textbf{47.3} to \textbf{25.7}, \textbf{26.0}, \textbf{26.6}, and \textbf{24.9}, respectively, at $\approx$25--31\% retention. In contrast, operating at a strictly lower \textbf{16.2\%} budget, \our{} preserves an mIoU of \textbf{35.6}. Allocating pixels \emph{before} encoding---rather than dropping frames or pruning tokens post-hoc---confers robustness that these baselines cannot match, even at tighter budgets.

\textbf{Reasoning without temporal anchors regresses.}
On VideoAuto-R1 (Qwen2.5-VL), naively extending from 32 to 128 frames \emph{degrades} Charades-STA mIoU from \textbf{41.5} to \textbf{28.9}: longer reasoning chains cannot compensate for the diluted temporal signal that accompanies quadratically growing token sequences. Incorporating \our{} at 16.1\% retention raises the 128-frame score to \textbf{49.1}, demonstrating that input-side allocation rescues long-context reasoning by concentrating visual budget on the temporally decisive frames.

\textbf{Emergent denoising.}
On NExT-GQA (Qwen3-VL, 128F), \our{} improves mIoU from \textbf{36.6} to \textbf{43.9} at 6.8\% retention. Aggressively suppressing question-irrelevant frames sharpens localization: removing noise is itself a form of signal enhancement.

\subsubsection{Image-Task Transfer}
\label{sec:image_results}
We evaluate \our{} on static image benchmarks to characterize domain boundaries. The clearest positive result is ChartQA on Qwen2.5-VL, where the Allocator upscales chart-bearing images to \textbf{105\%} of the native budget. However, text-heavy tasks degrade once resolution drops below their evidence threshold. Full results appear in Appendix~\ref{app:image_transfer}; we treat image transfer as a boundary condition rather than a primary contribution.

\begin{table}[t]
    \centering
    \footnotesize
    \setlength{\tabcolsep}{0.35em}
    \caption{\textbf{Latency breakdown (ms, $\downarrow$) on Qwen2.5-VL-7B with single-GPU Allocator and 4-GPU vLLM engine.} Averaged over 200 runs after 5 warm-up; E2E latency $=$ Scale Time $+$ Gen. Time.}
    \label{table:latency}
    \resizebox{\linewidth}{!}{
    \begin{tabular}{lccccccccccccc}
    \toprule
    \multirow{2}{*}{\textbf{Method}} & \multirow{2}{*}{\textbf{\#Frames}} & \multirow{2}{*}{\textbf{\makecell{Retention\\Ratio $R$}}} & \multicolumn{6}{c}{\textbf{Scale}} & \multicolumn{3}{c}{\textbf{Inference}} & \multicolumn{2}{c}{\textbf{Total}} \\
    \cmidrule(lr){4-9} \cmidrule(lr){10-12} \cmidrule(lr){13-14}
     &  &  & \textbf{TFLOPs} & \textbf{\makecell{Text\\Enc.}} & \textbf{\makecell{Visual\\Enc.}} & \textbf{\makecell{Scale\\Pred.}} & \textbf{\makecell{Scale\\Apply}} & \textbf{\makecell{Scale\\Time}} & \textbf{TFLOPs} & \textbf{TTFT} & \textbf{\makecell{Gen.\\Time}} & \textbf{TFLOPs} & \textbf{\makecell{E2E\\Time}} \\
    \midrule
    Vanilla & 16 & 100\% & -- & -- & -- & -- & -- & -- & 111.4 & 378.9 & 527.9 & 111.4 & 527.9 \\
    \cellcolor{metabg}\our{} & \cellcolor{metabg}16 & \cellcolor{metabg}76.3\%  & \cellcolor{metabg}1.5  & \cellcolor{metabg}19.8 & \cellcolor{metabg}94.1  & \cellcolor{metabg}85.6 & \cellcolor{metabg}6.3  & \cellcolor{metabg}205.8 & \cellcolor{metabg}77.2~\textcolor{green!60!black}{\tiny($\downarrow$30.7\%)} & \cellcolor{metabg}272.5~\textcolor{green!60!black}{\tiny($\downarrow$28.1\%)} & \cellcolor{metabg}370.7~\textcolor{green!60!black}{\tiny($\downarrow$29.8\%)} & \cellcolor{metabg}80.1~\textcolor{green!60!black}{\tiny($\downarrow$28.1\%)} & \cellcolor{metabg}576.5~\textcolor{red!60!black}{\tiny($\uparrow$9.2\%)} \\
    \cellcolor{metabg}\our{} & \cellcolor{metabg}16 & \cellcolor{metabg}52.8\%  & \cellcolor{metabg}1.5  & \cellcolor{metabg}19.9 & \cellcolor{metabg}102.9 & \cellcolor{metabg}94.5 & \cellcolor{metabg}8.4  & \cellcolor{metabg}225.7 & \cellcolor{metabg}51.5~\textcolor{green!60!black}{\tiny($\downarrow$53.8\%)} & \cellcolor{metabg}261.5~\textcolor{green!60!black}{\tiny($\downarrow$31.0\%)} & \cellcolor{metabg}313.1~\textcolor{green!60!black}{\tiny($\downarrow$40.7\%)} & \cellcolor{metabg}54.4~\textcolor{green!60!black}{\tiny($\downarrow$51.2\%)} & \cellcolor{metabg}538.8~\textcolor{red!60!black}{\tiny($\uparrow$2.1\%)} \\
    \cellcolor{metabg}\our{} & \cellcolor{metabg}16 & \cellcolor{metabg}28.9\%  & \cellcolor{metabg}1.5  & \cellcolor{metabg}20.4 & \cellcolor{metabg}103.4 & \cellcolor{metabg}92.2 & \cellcolor{metabg}9.0  & \cellcolor{metabg}225.0 & \cellcolor{metabg}31.0~\textcolor{green!60!black}{\tiny($\downarrow$72.2\%)} & \cellcolor{metabg}227.2~\textcolor{green!60!black}{\tiny($\downarrow$40.0\%)} & \cellcolor{metabg}237.9~\textcolor{green!60!black}{\tiny($\downarrow$54.9\%)} & \cellcolor{metabg}33.9~\textcolor{green!60!black}{\tiny($\downarrow$69.6\%)} & \cellcolor{metabg}462.9~\textcolor{green!60!black}{\tiny($\downarrow$12.3\%)} \\
    Vanilla & 32 & 100\% & -- & -- & -- & -- & -- & -- & 222.5 & 723.3 & 881.9 & 222.5 & 881.9 \\
    \cellcolor{metabg}\our{} & \cellcolor{metabg}32 & \cellcolor{metabg}74.4\%  & \cellcolor{metabg}2.9  & \cellcolor{metabg}19.9 & \cellcolor{metabg}204.1 & \cellcolor{metabg}97.4 & \cellcolor{metabg}14.4 & \cellcolor{metabg}335.9 & \cellcolor{metabg}153.9~\textcolor{green!60!black}{\tiny($\downarrow$30.8\%)} & \cellcolor{metabg}589.4~\textcolor{green!60!black}{\tiny($\downarrow$18.5\%)} & \cellcolor{metabg}627.6~\textcolor{green!60!black}{\tiny($\downarrow$28.8\%)} & \cellcolor{metabg}159.7~\textcolor{green!60!black}{\tiny($\downarrow$28.2\%)} & \cellcolor{metabg}963.5~\textcolor{red!60!black}{\tiny($\uparrow$9.2\%)} \\
    \cellcolor{metabg}\our{} & \cellcolor{metabg}32 & \cellcolor{metabg}51.5\%  & \cellcolor{metabg}2.9  & \cellcolor{metabg}20.0 & \cellcolor{metabg}193.2 & \cellcolor{metabg}92.0 & \cellcolor{metabg}16.2 & \cellcolor{metabg}321.4 & \cellcolor{metabg}102.4~\textcolor{green!60!black}{\tiny($\downarrow$54.0\%)} & \cellcolor{metabg}505.0~\textcolor{green!60!black}{\tiny($\downarrow$30.2\%)} & \cellcolor{metabg}467.1~\textcolor{green!60!black}{\tiny($\downarrow$47.0\%)} & \cellcolor{metabg}108.2~\textcolor{green!60!black}{\tiny($\downarrow$51.4\%)} & \cellcolor{metabg}788.5~\textcolor{green!60!black}{\tiny($\downarrow$10.6\%)} \\
    \cellcolor{metabg}\our{} & \cellcolor{metabg}32 & \cellcolor{metabg}28.2\%  & \cellcolor{metabg}2.9  & \cellcolor{metabg}20.3 & \cellcolor{metabg}190.4 & \cellcolor{metabg}90.3 & \cellcolor{metabg}17.3 & \cellcolor{metabg}318.3 & \cellcolor{metabg}61.4~\textcolor{green!60!black}{\tiny($\downarrow$72.4\%)} & \cellcolor{metabg}451.8~\textcolor{green!60!black}{\tiny($\downarrow$37.5\%)} & \cellcolor{metabg}332.6~\textcolor{green!60!black}{\tiny($\downarrow$62.3\%)} & \cellcolor{metabg}67.2~\textcolor{green!60!black}{\tiny($\downarrow$69.8\%)} & \cellcolor{metabg}650.9~\textcolor{green!60!black}{\tiny($\downarrow$26.2\%)} \\
    Vanilla & 64 & 100\% & -- & -- & -- & -- & -- & -- & 444.6 & 1457.5 & 2059.6 & 444.6 & 2059.6 \\
    \cellcolor{metabg}\our{} & \cellcolor{metabg}64 & \cellcolor{metabg}73.2\%  & \cellcolor{metabg}5.8 & \cellcolor{metabg}19.8 & \cellcolor{metabg}389.5 & \cellcolor{metabg}95.8 & \cellcolor{metabg}26.4 & \cellcolor{metabg}531.5 & \cellcolor{metabg}307.3~\textcolor{green!60!black}{\tiny($\downarrow$30.9\%)} & \cellcolor{metabg}1093.1~\textcolor{green!60!black}{\tiny($\downarrow$25.0\%)} & \cellcolor{metabg}1327.0~\textcolor{green!60!black}{\tiny($\downarrow$35.6\%)} & \cellcolor{metabg}318.9~\textcolor{green!60!black}{\tiny($\downarrow$28.3\%)} & \cellcolor{metabg}1858.5~\textcolor{green!60!black}{\tiny($\downarrow$9.8\%)} \\
    \cellcolor{metabg}\our{} & \cellcolor{metabg}64 & \cellcolor{metabg}50.7\%  & \cellcolor{metabg}5.8 & \cellcolor{metabg}20.1 & \cellcolor{metabg}382.1 & \cellcolor{metabg}94.9 & \cellcolor{metabg}29.9 & \cellcolor{metabg}527.0 & \cellcolor{metabg}204.3~\textcolor{green!60!black}{\tiny($\downarrow$54.0\%)} & \cellcolor{metabg}991.8~\textcolor{green!60!black}{\tiny($\downarrow$31.9\%)} & \cellcolor{metabg}740.5~\textcolor{green!60!black}{\tiny($\downarrow$64.0\%)} & \cellcolor{metabg}215.9~\textcolor{green!60!black}{\tiny($\downarrow$51.4\%)} & \cellcolor{metabg}1267.5~\textcolor{green!60!black}{\tiny($\downarrow$38.5\%)} \\
    \cellcolor{metabg}\our{} & \cellcolor{metabg}64 & \cellcolor{metabg}27.8\%  & \cellcolor{metabg}5.8 & \cellcolor{metabg}20.0 & \cellcolor{metabg}371.6 & \cellcolor{metabg}90.2 & \cellcolor{metabg}34.8 & \cellcolor{metabg}516.6 & \cellcolor{metabg}122.2~\textcolor{green!60!black}{\tiny($\downarrow$72.5\%)} & \cellcolor{metabg}899.2~\textcolor{green!60!black}{\tiny($\downarrow$38.3\%)} & \cellcolor{metabg}511.4~\textcolor{green!60!black}{\tiny($\downarrow$75.2\%)} & \cellcolor{metabg}133.8~\textcolor{green!60!black}{\tiny($\downarrow$69.9\%)} & \cellcolor{metabg}1028.0~\textcolor{green!60!black}{\tiny($\downarrow$50.1\%)} \\
    Vanilla & 128 & 100\% & -- & -- & -- & -- & -- & -- & 888.9 & 2936.3 & 4877.0 & 888.9 & 4877.0 \\
    \cellcolor{metabg}\our{} & \cellcolor{metabg}128 & \cellcolor{metabg}74.2\%  & \cellcolor{metabg}11.6 & \cellcolor{metabg}20.1 & \cellcolor{metabg}766.3 & \cellcolor{metabg}95.0 & \cellcolor{metabg}53.1 & \cellcolor{metabg}934.5 & \cellcolor{metabg}614.1~\textcolor{green!60!black}{\tiny($\downarrow$30.9\%)} & \cellcolor{metabg}2286.6~\textcolor{green!60!black}{\tiny($\downarrow$22.1\%)} & \cellcolor{metabg}2323.6~\textcolor{green!60!black}{\tiny($\downarrow$52.4\%)} & \cellcolor{metabg}637.3~\textcolor{green!60!black}{\tiny($\downarrow$28.3\%)} & \cellcolor{metabg}3258.1~\textcolor{green!60!black}{\tiny($\downarrow$33.2\%)} \\
    \cellcolor{metabg}\our{} & \cellcolor{metabg}128 & \cellcolor{metabg}51.4\%  & \cellcolor{metabg}11.6 & \cellcolor{metabg}20.2 & \cellcolor{metabg}755.3 & \cellcolor{metabg}93.8 & \cellcolor{metabg}59.4 & \cellcolor{metabg}928.7 & \cellcolor{metabg}408.0~\textcolor{green!60!black}{\tiny($\downarrow$54.1\%)} & \cellcolor{metabg}2071.0~\textcolor{green!60!black}{\tiny($\downarrow$29.5\%)} & \cellcolor{metabg}1496.0~\textcolor{green!60!black}{\tiny($\downarrow$69.3\%)} & \cellcolor{metabg}431.2~\textcolor{green!60!black}{\tiny($\downarrow$51.5\%)} & \cellcolor{metabg}2424.7~\textcolor{green!60!black}{\tiny($\downarrow$50.3\%)} \\
    \cellcolor{metabg}\our{} & \cellcolor{metabg}128 & \cellcolor{metabg}28.2\%  & \cellcolor{metabg}11.6 & \cellcolor{metabg}20.4 & \cellcolor{metabg}734.5 & \cellcolor{metabg}92.0 & \cellcolor{metabg}68.6 & \cellcolor{metabg}915.5 & \cellcolor{metabg}243.9~\textcolor{green!60!black}{\tiny($\downarrow$72.6\%)} & \cellcolor{metabg}1766.7~\textcolor{green!60!black}{\tiny($\downarrow$39.8\%)} & \cellcolor{metabg}1061.8~\textcolor{green!60!black}{\tiny($\downarrow$78.2\%)} & \cellcolor{metabg}267.1~\textcolor{green!60!black}{\tiny($\downarrow$70.0\%)} & \cellcolor{metabg}1977.3~\textcolor{green!60!black}{\tiny($\downarrow$59.5\%)} \\
    \bottomrule
    \end{tabular}}
\end{table}

\subsection{Runtime Overhead}
\label{sec:latency}

Table~\ref{table:latency} measures pipeline latency using a dedicated single-GPU Allocator and a separate 4-GPU vLLM engine. The key trade-off is when downstream token savings amortize the front-end allocation cost.

At $R{\approx}\mathbf{74\%}$, generation time drops \textbf{29--52\%} but end-to-end (E2E) savings appear only at $\geq$64 frames ($\mathbf{-9.8\%}$), growing to $\mathbf{-33.2\%}$ at 128 frames. At $R{\approx}\mathbf{51\%}$, break-even shifts to 32 frames ($\mathbf{-10.6\%}$ E2E); at $R{\approx}\mathbf{28\%}$, wall-clock savings emerge even at 16 frames ($\mathbf{-12.3\%}$), reaching $\mathbf{-59.5\%}$ at 128 frames with \textbf{78\%} generation-time reduction. Backbone savings compound faster than the fixed Allocator overhead as sequences grow---a direct consequence of the quadratic attention cost---making \our{} most impactful in the long-context regime.

\begin{figure}[t]
    \centering
    \includegraphics[width=\linewidth]{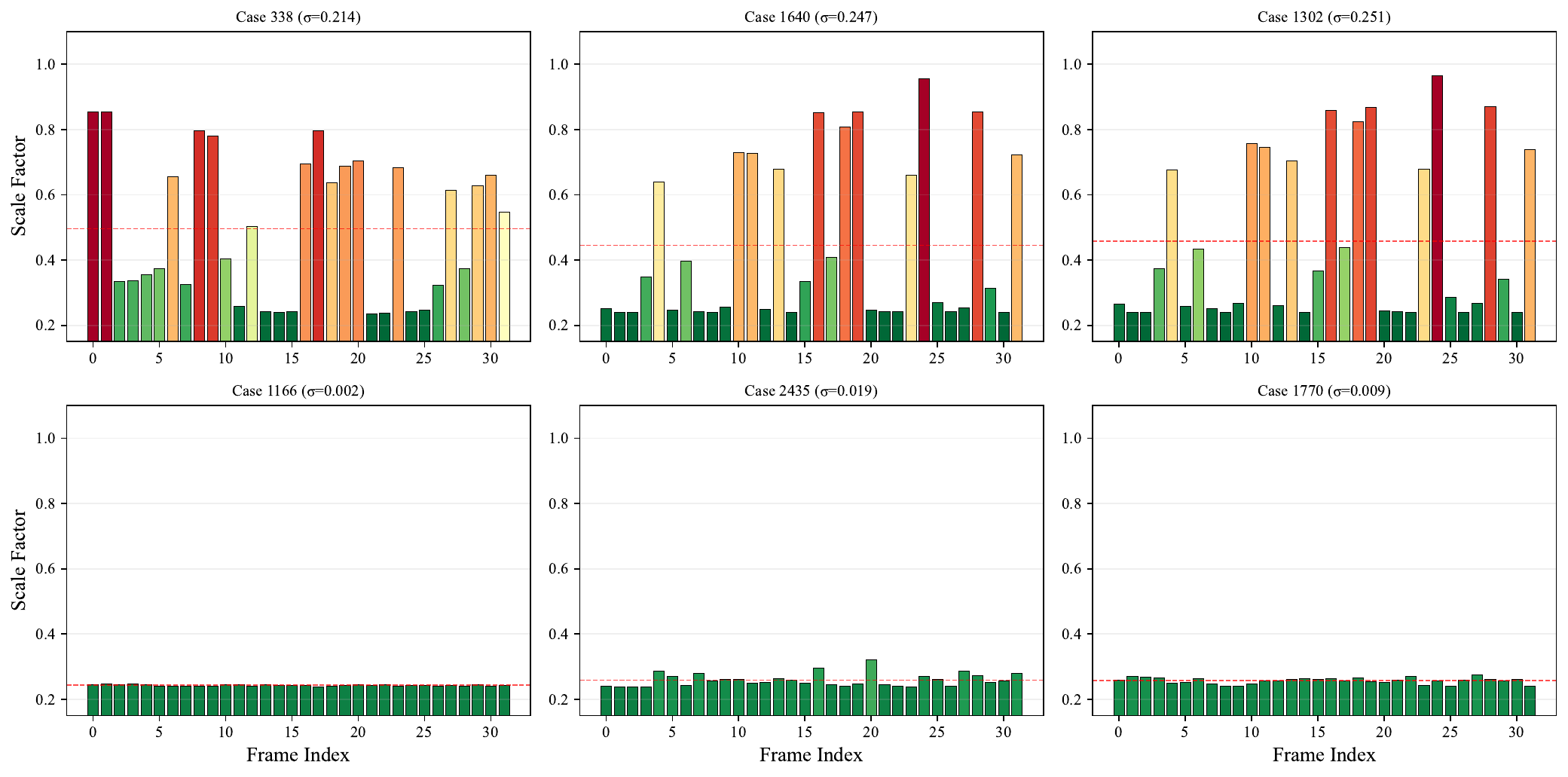}
    \caption{\textbf{Emergent active perception.} Per-frame scale $s_t$ over frame index for six VideoMME videos, grouped by intra-video scale diversity $\sigma$. High-diversity videos show localized scale spikes on scene changes, text overlays, and rapid motion; low-diversity videos remain near-uniform.}
    \label{fig:scale_profiles}
\end{figure}

\begin{figure}[t]
    \centering
    \includegraphics[width=\linewidth]{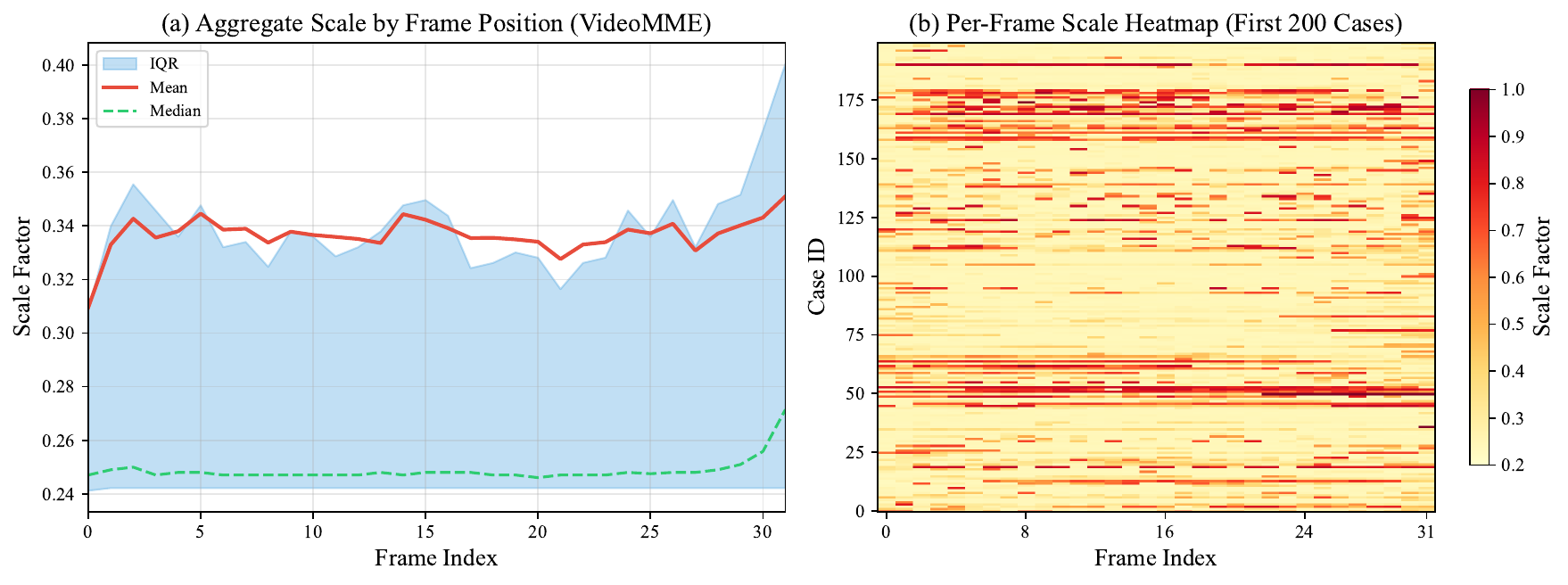}
    \caption{\textbf{Global allocation statistics on VideoMME.} \textbf{(a)} Aggregate predicted scale by frame position. \textbf{(b)} Case$\times$frame heatmap for the first 200 videos. High-scale allocation appears as localized bursts rather than a fixed positional pattern.}
    \label{fig:scale_aggregate}
\end{figure}

\subsection{Analysis}
\label{sec:analysis}

\subsubsection{Emergent Active Perception}
\label{sec:active_perception}

Figures~\ref{fig:scale_profiles} and~\ref{fig:scale_aggregate} reveal the mechanism behind the main results. The Allocator does not learn a uniform compression policy; rather, it learns a strongly sparse temporal allocation, keeping most of the video at near-minimum resolution and concentrating budget on short bursts around text overlays, scene transitions, or other brief informative events.

This behavior is not a trivial positional prior. The median scale stays close to the low end of the range, while the mean is lifted by localized peaks---indicating that high-resolution allocation is the exception, not the default. The per-video heatmap further confirms that these peaks are content-driven segments, not a fixed bias toward the beginning or end of the clip. In short, the policy spends pixels where the answer is likely to be decided.

\subsubsection{Ablation Studies}
\label{sec:ablation_reward}

\begin{table*}[t]
    \centering
    \footnotesize
    \renewcommand{\arraystretch}{1.05}
    \caption{\textbf{Distribution family ablation for CAPO.} The two variants follow the same training protocol.}
    \resizebox{\linewidth}{!}{
    \begin{tabular}{lccccccccc}
    \toprule
    \textbf{Variant} & \textbf{$\bar{s}$} & \textbf{VideoMME} & \textbf{LongVideoBench} & \textbf{MMVU} & \multicolumn{3}{c}{\textbf{VideoMMMU}} & \textbf{LVBench} \\
    \cmidrule(lr){6-8}
     &  &  &  &  & \textbf{Per.} & \textbf{Comp.} & \textbf{Adap.} &  \\
    \midrule
    $\beta$-CAPO & 0.54 & 60.3 & \textbf{58.2} & 51.2 & 65.0 & \textbf{54.3} & 28.7 & \textbf{37.6} \\
    $\mathcal{N}$-CAPO & 0.60 & \textbf{61.0} & 57.4 & \textbf{51.8} & \textbf{66.0} & 50.0 & \textbf{30.3} & 37.2 \\
    \bottomrule
    \end{tabular}
    }
    \label{table:capo_dist_ablation}
\end{table*}

\textbf{CAPO reward design.}
Two questions arise: \emph{how} should cost enter the optimization, and \emph{what} prevents the policy from collapsing to a uniform scaler?

\begin{figure}[t]
    \centering
    \includegraphics[width=\linewidth]{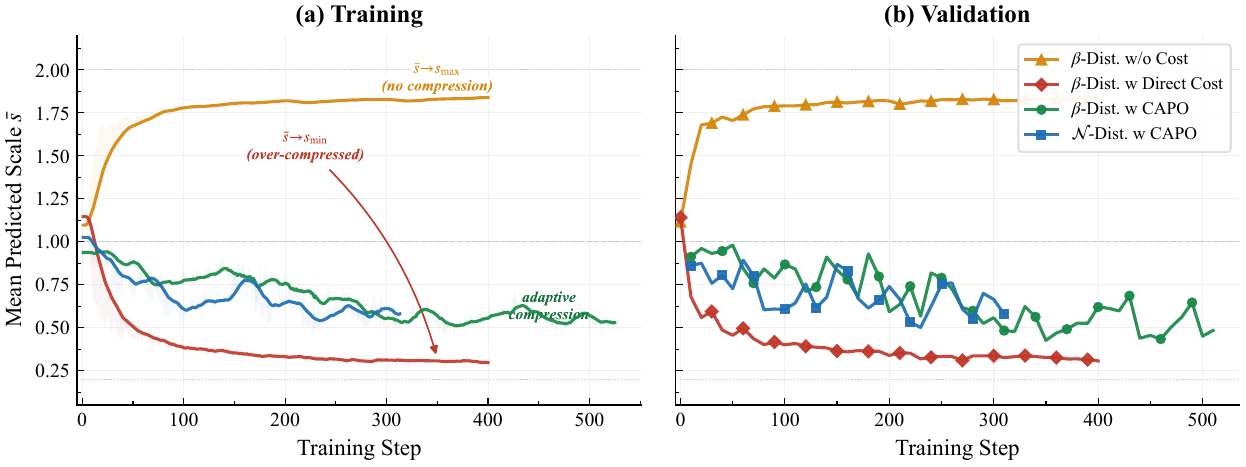}
    \caption{\textbf{Reward-design ablation.} Mean predicted scale $\bar{s}$ during training and validation. Direct cost penalties collapse to the minimum scale, whereas CAPO variants converge to stable intermediate operating points.}
    \label{fig:reward_ablation}
\end{figure}

Table~\ref{table:capo_dist_ablation} shows that the exact policy family is secondary: $\beta$-CAPO and $\mathcal{N}$-CAPO trade marginal advantages across benchmarks with neither variant consistently dominating. The shared ingredient that matters is CAPO's asymmetric cost shaping, not the parametric form. Figure~\ref{fig:reward_ablation} makes this more explicit from a training-dynamics perspective. Direct cost penalties drive the policy rapidly toward the minimum-scale boundary, while removing cost altogether pushes toward the upper bound. CAPO stabilizes an intermediate operating point where the model is rewarded for being selective---not merely cheap and not merely accurate. Further analysis of per-sample adaptivity and convergence appears in Appendix~\ref{app:ablation_studies}.

\begin{wraptable}[14]{r}{0.58\textwidth}
\vspace{-4mm}
\centering
\small
\renewcommand{\arraystretch}{0.95}
\caption{\textbf{Operator generalization.} Zero-shot transfer of \our{} scores to frame selection. Combining top-$K$ selection with adaptive resizing from 128 candidate frames outperforms uniform sampling baselines at lower token budgets.}
\label{tab:frame_select}
\resizebox{\linewidth}{!}{
\begin{tabular}{lcccc}
\toprule
\textbf{Method} & \textbf{VideoMME} & \textbf{LongVideoBench} & \textbf{LVBench} & \textbf{MMVU} \\
\midrule
\multicolumn{5}{l}{\textit{Budget: 8 frames}} \\
Vanilla & 54.0 & 53.9 & 33.3 & 48.9 \\
Top-8 Select & 52.2 & 51.1 & 32.0 & 49.2 \\
\midrule
\multicolumn{5}{l}{\textit{Budget: 16 frames}} \\
Vanilla & 58.9 & 56.0 & 36.1 & 50.9 \\
Threshold Select & 58.0 & 57.4 & 36.4 & 51.0 \\
\quad \textit{\scriptsize Avg. Budget (Retention Ratio)} & \scriptsize 12.2f (9.5\%) & \scriptsize 23.2f (18.1\%) & \scriptsize 16.7f (13.0\%) & \scriptsize 17.2f (13.4\%) \\
\cdashline{1-5}
\cellcolor{metabg}Top-32 Select + Resize & \cellcolor{metabg} 60.6 & \cellcolor{metabg} 57.2 & \cellcolor{metabg} 38.9 & \cellcolor{metabg} 50.2 \\
\cellcolor{metabg} \quad \textit{\scriptsize Avg. Budget (Retention Ratio)} & \cellcolor{metabg} \scriptsize 11.7f (9.1\%) & \cellcolor{metabg} \scriptsize 16.9f (13.2\%) & \cellcolor{metabg} \scriptsize 13.7f (10.7\%) & \cellcolor{metabg} \scriptsize 14.1f (11.0\%) \\
\midrule
\multicolumn{5}{l}{\textit{Budget: 32 frames}} \\
Vanilla & 62.3 & 58.7 & 39.5 & 52.0 \\
Top-32 Select & 59.7 & 55.7 & 37.0 & 51.2 \\
\cdashline{1-5}
\cellcolor{metabg}Top-64 Select + Resize & \cellcolor{metabg} 62.5 & \cellcolor{metabg} 58.4 & \cellcolor{metabg} 40.0 & \cellcolor{metabg} 52.3 \\
\cellcolor{metabg} \quad \textit{\scriptsize Avg. Budget (Retention Ratio)} & \cellcolor{metabg} \scriptsize 23.8f (18.6\%) & \cellcolor{metabg} \scriptsize 36.2f (28.3\%) & \cellcolor{metabg} \scriptsize 24.1f (18.8\%) & \cellcolor{metabg} \scriptsize 32.5f (25.4\%) \\
\bottomrule
\end{tabular}
}
\vspace{-4mm}
\end{wraptable}

\textbf{Operator generalization.}
Although \our{} is trained exclusively for adaptive resizing, its learned policy transfers zero-shot to frame selection. We repurpose the Allocator's predicted scales as importance scores to rank and filter 128 candidate frames. Table~\ref{tab:frame_select} shows that selecting and resizing the top-32 or top-64 frames consistently outperforms the vanilla 16-frame and 32-frame baselines, respectively, despite consuming fewer tokens. The policy thus captures an operator-agnostic notion of visual importance that generalizes beyond the training-time operator.

\begin{figure}[t]
    \centering
    \includegraphics[width=\linewidth]{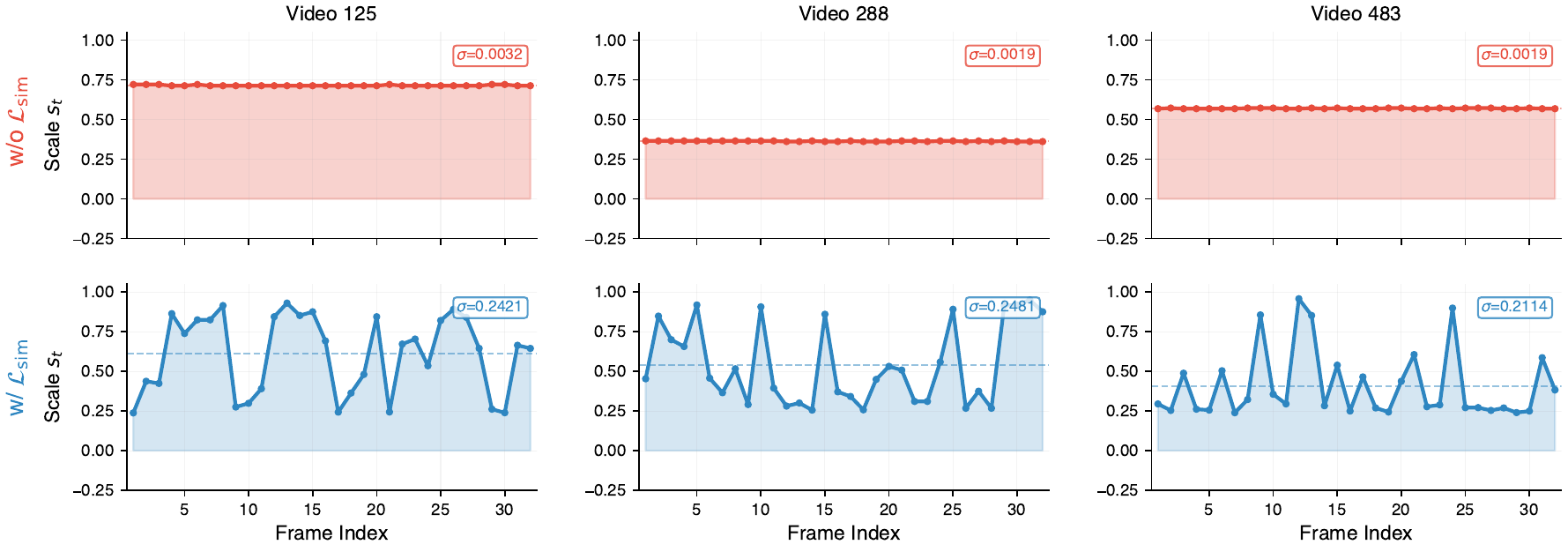}
    \caption{\textbf{$\mathcal{L}_{\text{sim}}$ ablation: per-frame scale profiles.} Without temporal-similarity regularization, the Allocator approaches near-uniform scaling; with it, the policy concentrates resolution on selected frames and suppresses redundant neighbors.}
    \label{fig:sim_profiles}
\end{figure}

\textbf{Temporal regularization complements CAPO.}
\label{sec:ablation_sim}
CAPO determines how cost enters the learning signal, but it does not by itself force the Allocator to distinguish among visually redundant neighbors. Figure~\ref{fig:sim_profiles} shows that removing $\mathcal{L}_{\text{sim}}$ collapses the scale trace toward a near-constant profile resembling FixedScale. Reintroducing $\mathcal{L}_{\text{sim}}$ restores sharp frame-level differentiation. The two mechanisms are complementary: CAPO stabilizes the accuracy--cost operating point, while $\mathcal{L}_{\text{sim}}$ breaks the symmetry that otherwise favors uniform allocation.

\subsubsection{Robustness and Limitations}
\label{sec:robustness}
Adaptive allocation is not a lossless compression layer. \our{} preserves a large majority of originally correct predictions, but can still miss decisive evidence when the relevant cue appears briefly against a simple background. Because the policy is open-loop---budget decisions are committed before any backbone processing begins---it cannot revise allocations once reasoning starts. We therefore interpret the performance gains as selective redistribution of visual budget rather than guaranteed preservation of all useful information. Detailed failure-case analysis is provided in Appendix~\ref{app:case_studies}.

\section{Related Work}

\paragraph{Input-side adaptation before visual encoding.}
A growing body of work reduces visual cost \emph{before} or \emph{during} input construction. Early approaches primarily perform temporal downsampling through keyframe selection or clip condensation~\citep{liang2024keyvideollm,zhu2025focus,sun2025frames,tang2025adaptive}. More recent methods incorporate query awareness and iterative search, tailoring frame selection to question types or intermediate evidence~\citep{zou2025air,li2025divide,guo2025logic,he2025framethinker}. Beyond selecting \emph{which} frames to process, several works allocate perceptual budgets via multi-resolution encoding. Slow--fast pipelines~\citep{yang2025kwai,zhang2026penguin} use inter-frame similarity to route frames to high- or low-resolution paths, but their binary, query-agnostic routing cannot adapt to the downstream question; we omit direct experimental comparisons because these systems target different backbone families and evaluation protocols. Query-aware multi-resolution strategies~\citep{zhang2025qframe} and early truncation of less informative visual tokens~\citep{chen2026soft} go further by conditioning on the query, yet still rely on handcrafted rules or fixed resolution bins. We note that QFrame~\citep{zhang2025qframe} operates with predefined resolution tiers and rule-based selection, whereas \our{} learns continuous allocations end-to-end from task reward; a controlled comparison would require re-implementing QFrame's proprietary routing logic on our backbone, which we leave for future work. In contrast, \our{} is an Input-side adaptation framework: it learns input-side allocations from task reward via RL and can realize them through different pre-encoding operators, including resizing and frame selection; the experiments in this paper study the continuous resize instantiation.

\paragraph{Model-side token economy after encoding.}
Post-encoding methods prune, merge, or redistribute visual tokens in embedding space. For images, representative approaches include token merging~\citep{bolya2022token}, attention- or saliency-guided pruning~\citep{fastv,visionzip,llavaprumerge,vispruner}, progressive dropping~\citep{pyramiddrop,sparsevlm}, context compression~\citep{liao2025beyond}, KV cache sparsity~\citep{liao2025spark}, and diversity-based budget allocation~\citep{divprune,topv,vscan}. Video-specific extensions exploit spatiotemporal redundancy via static/dynamic token separation~\citep{prunevid,fastvid}, hierarchical merging~\citep{hyun2025multi}, and segment-level fusion or budget allocation~\citep{tao2025dycoke,framefusion,holitom}. These methods are complementary to \our{}: they operate \emph{after} visual encoding and cannot recover high-frequency details lost to undersampling before encoding. Our focus is earlier in the pipeline, deciding how many pixels to encode in the first place.

\paragraph{Output-side agentic reasoning.}
Another strategy leaves the input fixed and recovers efficiency through iterative reasoning: retrieve candidate frames, zoom into regions, then re-query the model. Approaches range from static toolsets with predefined cropping or clipping operators~\citep{zheng2025deepeyes,wang2025pixel,song2026adareasoner} to dynamic tooling via code-generation primitives~\citep{zhang2025thyme,zhao2025pyvision,hong2025deepeyesv2}, often exposed through executable interfaces~\citep{wang2024executable}. While these methods can target hard evidence precisely, they are multi-pass by construction and rely on an initial coarse view to trigger subsequent refinement. \our{} instead studies whether a \emph{single-pass} pre-encoding allocation policy can recover much of this benefit without the latency and control overhead of iterative interaction.

\paragraph{RL for multimodal reasoning and perception control.}
Recent work has extended RL post-training from language models~\citep{shao2024deepseekmath,guo2025deepseek,tan2025bottom} to multimodal reasoning and video understanding. Algorithmic refinements include improved advantage estimation and PPO-style stabilization~\citep{liu2025understanding,yu2025dapo,zheng2025group}, while video-domain extensions strengthen reasoning through iterative frame selection and evidence refinement~\citep{feng2025videor1,li2025videochat,liu2026videoauto,yang2025longvt,chen2025scaling,wang2025time,fu2025love}. Our use of RL is orthogonal: we apply it to \emph{input-side perception control}---learning frame-level visual allocations under an explicit accuracy--cost trade-off---rather than output-side reasoning policies. CAPO is designed for this setting, where naive cost penalties drive the policy to a degenerate low-budget solution.

\section{Conclusion}

Mainstream video MLLM efficiency methods compress tokens \emph{after} encoding, forcing models to absorb the full computational cost of dense pixel processing before realizing any savings. \our{} dismantles this bottleneck. By introducing a query-aware Allocator, our framework assigns per-frame visual budgets \emph{prior} to feature extraction. The vision encoder therefore processes a strictly task-relevant, sparse input, leaving native kernel optimizations entirely untouched. To train this non-differentiable pipeline, we propose CAPO. This asymmetric reward formulation heavily penalizes wasteful allocations, while a temporal-similarity regularizer breaks uniform-resolution equilibria, forcing the policy to differentiate among visually redundant neighbors.

This input-side adaptation fundamentally alters the efficiency-accuracy tradeoff. Operating at merely ${\sim}$10\% visual-token retention, \our{} matches or exceeds the performance of dense baselines on complex reasoning tasks. Crucially, it translates these spatial savings directly into expanded temporal coverage. Under equivalent compute, the model processes up to $16{\times}$ more frames, yielding $>$15\% relative gains on long-video benchmarks. Beyond macroscopic metrics, the learned policy demonstrates emergent active perception. Driven entirely by task-level rewards, it concentrates high resolution strictly on the sparse frames required to resolve the query.

Two boundaries remain. The current allocation mechanism operates open-loop: once the model commits a resolution budget, it cannot dynamically revise this decision during reasoning. Moreover, the specific instantiation of spatial resizing transfers unevenly to non-video modalities. Closing this loop defines the next critical frontier. By allowing intermediate backbone states to trigger selective re-encoding, future work can transform static budget prediction into a fully adaptive, reasoning-aware allocation paradigm.

\bibliography{main}

\section*{Limitations and Future Work}

\our{} substantially advances the efficiency--accuracy Pareto front for long-video MLLMs. Nonetheless, four specific design choices constrain our current findings.

\textit{(i) Front-end overhead is amortized only in the long-context regime.}
The Allocator imposes a fixed pre-encoding cost---comprising coarse visual encoding, cross-frame fusion, and distribution prediction---before realizing any backbone savings. For short sequences ($T\!\leq\!32$), this constant overhead offsets a significant fraction of the downstream attention reduction. Consequently, definitive wall-clock speedups emerge primarily in the long-context regime (Sec.~\ref{sec:latency}). Future work must reduce this fixed cost via cached video features, lightweight front-ends, or distilled allocation rules.

\textit{(ii) Allocation relies on coarse visual evidence.}
The Allocator operates on frozen coarse features $\boldsymbol{f}_t\!\in\!\mathbb{R}^{D}$ rather than full high-resolution frames. This representation suffices to detect broad redundancies and scene structures. However, it struggles with small text, subtle objects, and transient answer-critical cues embedded within otherwise simple frames (Figure~\ref{fig:case_fork}). Incorporating multi-scale conditioning, motion-aware features, or lightweight local refinement could bridge this gap without sacrificing the speed of the current front-end.

\textit{(iii) Single video-centric instantiation limits broad validation.}
While we formulate \our{} as a general input-side adaptation framework, our experiments instantiate the operator strictly as resizing and train the policy primarily on video tasks. Transfer beyond this regime remains uneven. The learned policy occasionally identifies static images requiring higher fidelity, yet fails to deliver uniformly efficiency-preserving gains on image-centric benchmarks (Table~\ref{table:benchmark_image}). Extending the training mixture to image--video data and exploring alternative operators, such as hard frame selection, remain open problems.

\textit{(iv) Open-loop allocation ignores reasoning state.}
The framework commits all budget decisions before the backbone processes any visual tokens. The policy cannot revise a mistaken low-resolution choice once partial reasoning or uncertainty signals emerge. Closing this loop represents a natural extension. Early backbone states could trigger re-encoding, budget revision, or a secondary visual pass only when strictly necessary.

\section*{Software and Data}

The code for this paper is available at: \url{https://github.com/Xnhyacinth/ResAdapt}

\appendix

\section{Derivations and Theoretical Analysis}
\label{app:theoretical_analysis}

This section consolidates the mathematical foundations underpinning \our{}. We first derive the joint RL formulation and then formalize the computational bounds.

\subsection{Derivation of Joint RL Formulation}
\label{app:derivation}

This section details derivations omitted from Sec.~\ref{main:method} and clarifies how the one-step contextual MDP (Contextual Bandit) introduced in Sec.~\ref{sec:problem_formulation} motivates our practical surrogate objectives. We state all derivations for a single context (video and query). The full objective requires taking the expectation over the dataset $\mathcal{D}$.

\textit{Notation.} Let $\boldsymbol{x} = (\boldsymbol{q}, \mathcal{V})$ denote the prompt context. The Allocator samples latent actions $\boldsymbol{a}$ from a Beta policy $q_\theta(\boldsymbol{a}\mid\boldsymbol{x})$ (Sec.~\ref{sec:allocator}). The continuous allocation $\boldsymbol{s}$ is the deterministic mapping of $\boldsymbol{a}$ via Eq.~\eqref{eq:scale_map}. Let $\pi_\theta(\boldsymbol{s}\mid\boldsymbol{x})$ denote the induced density (pushforward). A deterministic transformation yields the operator-transformed input $\tilde{\boldsymbol{x}} = (\boldsymbol{q}, \{\mathcal{O}(f_t, s_t)\}_{t=1}^T)$. In our experiments, \(\mathcal{O}\) implements bilinear resizing. The MLLM backbone policy $\pi_\phi(\boldsymbol{y} \mid \tilde{\boldsymbol{x}})$ then samples a complete response rollout $\boldsymbol{y}=(\boldsymbol{r},\boldsymbol{o})$, comprising the reasoning trace $\boldsymbol{r}$ and the final answer $\boldsymbol{o}$.

\textbf{One-Step Contextual MDP and the Joint Objective.}
Sec.~\ref{sec:problem_formulation} defines the system as a one-step contextual MDP. Here, sequential state transitions across time steps $t$ do not exist. The episode terminates immediately after the system samples the allocation \(\boldsymbol{s}\) and generates the corresponding rollout \(\boldsymbol{y}\). Consequently, value functions collapse to immediate rewards. The Policy Gradient Theorem simplifies drastically, eliminating the need for temporal discount factors or complex credit assignment across Markov states.

The joint distribution of the allocation and the rollout factorizes conditionally:
\begin{equation}
p_{\theta,\phi}(\boldsymbol{s},\boldsymbol{y} \mid \boldsymbol{x}) = \pi_\theta(\boldsymbol{s} \mid \boldsymbol{x}) \, \pi_\phi(\boldsymbol{y} \mid \tilde{\boldsymbol{x}}).
\end{equation}

For a single context with ground-truth answer $\boldsymbol{o}^{\star}$, the marginal answer probability under the transformed input is:
\begin{equation}
p_{\theta,\phi}(\boldsymbol{o}^{\star} \mid \boldsymbol{x})
=
\mathbb{E}_{\pi_\theta(\boldsymbol{s} \mid \boldsymbol{x})}
\!\left[
\mathbb{E}_{\pi_\phi(\boldsymbol{r} \mid \tilde{\boldsymbol{x}})}
\!\left[
\pi_\phi(\boldsymbol{o}^{\star} \mid \tilde{\boldsymbol{x}}, \boldsymbol{r})
\right]
\right].
\label{eq:ans_marginal_app}
\end{equation}
Equation~\eqref{eq:ans_marginal_app} expresses the law of total expectation under an autoregressive factorization $\pi_\phi(\boldsymbol{y}\mid\tilde{\boldsymbol{x}})=\pi_\phi(\boldsymbol{r}\mid\tilde{\boldsymbol{x}})\,\pi_\phi(\boldsymbol{o}\mid\tilde{\boldsymbol{x}},\boldsymbol{r})$. The inner term represents the conditional probability of the ground-truth answer $\boldsymbol{o}^{\star}$ given the reasoning prefix $\boldsymbol{r}$. Integrating over $\boldsymbol{r}$ yields the marginal $\mathbb{P}(\boldsymbol{o}^{\star}\mid\boldsymbol{x})$ strictly under this generative ordering. The subsequent RL objective does not require a closed-form evaluation of Eq.~\eqref{eq:ans_marginal_app}.

Since $\log(\cdot)$ is monotonically increasing, maximizing $\log p_{\theta,\phi}(\boldsymbol{o}^{\star}\mid\boldsymbol{x})$ serves as an equivalent objective. However, the RL derivation below avoids introducing the logarithm directly. It merely requires evaluating a scalar utility after sampling \((\boldsymbol{s},\boldsymbol{y})\). We abstract the answer-quality term as a rollout utility \(Q(\boldsymbol{x},\boldsymbol{y})\), where \(\boldsymbol{y}=(\boldsymbol{r},\boldsymbol{o})\), and treat it as parameter-independent post-sampling. This represents a modeling abstraction rather than an exact reformulation. When we define \(Q\) as an answer-aligned task score, the resulting RL problem acts as a surrogate for likelihood maximization. This formulation allows us to define the ideal rollout reward:
\begin{equation}
R^{\text{ideal}}_{\boldsymbol{s},\boldsymbol{y}} = Q(\boldsymbol{x},\boldsymbol{y}) - \lambda\,C(\boldsymbol{s}),
\end{equation}
and optimize the one-step expected return:
\begin{equation}
\max_{\theta,\phi}\ \mathcal{J}(\theta, \phi)
=
\mathbb{E}_{\boldsymbol{x} \sim \mathcal{D}}
\mathbb{E}_{\pi_\theta(\boldsymbol{s} \mid \boldsymbol{x})}
\!\left[
\mathbb{E}_{\pi_\phi(\boldsymbol{y} \mid \tilde{\boldsymbol{x}})}
\!\left[
R^{\text{ideal}}_{\boldsymbol{s},\boldsymbol{y}}
\right]
\right].
\label{eq:joint_obj_app}
\end{equation}

\textbf{Policy Gradient and Alternating Optimization.}
Because the objective involves two distinct parameterized policies, its gradients follow the score-function estimator (the REINFORCE identity). This establishes the underlying policy-gradient structure. GRPO/PPO retains this structure but replaces the raw reward with normalized advantages and clipped surrogates to stabilize optimization. Taking the gradient of $\mathcal{J}(\theta, \phi)$ with respect to the backbone parameters $\phi$:
\begin{align}
\nabla_\phi \mathcal{J}(\theta,\phi)
&= \mathbb{E}_{\boldsymbol{x}} \mathbb{E}_{\pi_\theta(\boldsymbol{s} \mid \boldsymbol{x})} \!\left[ \nabla_\phi \int \pi_\phi(\boldsymbol{y} \mid \tilde{\boldsymbol{x}}) R^{\text{ideal}}_{\boldsymbol{s},\boldsymbol{y}} d\boldsymbol{y} \right] \nonumber \\
&= \mathbb{E}_{\boldsymbol{x}} \mathbb{E}_{\pi_\theta(\boldsymbol{s} \mid \boldsymbol{x})} \mathbb{E}_{\pi_\phi(\boldsymbol{y} \mid \tilde{\boldsymbol{x}})} \!\left[ R^{\text{ideal}}_{\boldsymbol{s},\boldsymbol{y}}\, \nabla_\phi \log \pi_\phi(\boldsymbol{y} \mid \tilde{\boldsymbol{x}}) \right]. \label{eq:grad_phi_app}
\end{align}
Similarly, the gradient with respect to the Allocator parameters $\theta$ relies on the marginalized reward $R^{\text{ideal}}_{\boldsymbol{s}} = \mathbb{E}_{\pi_\phi(\boldsymbol{y} \mid \tilde{\boldsymbol{x}})}[R^{\text{ideal}}_{\boldsymbol{s},\boldsymbol{y}}]$:
\begin{align}
\nabla_\theta \mathcal{J}(\theta,\phi)
&= \mathbb{E}_{\boldsymbol{x}} \mathbb{E}_{\pi_\theta(\boldsymbol{s} \mid \boldsymbol{x})} \!\left[ R^{\text{ideal}}_{\boldsymbol{s}}\, \nabla_\theta \log \pi_\theta(\boldsymbol{s} \mid \boldsymbol{x}) \right]. \label{eq:grad_theta_app}
\end{align}

To optimize this objective via GRPO/PPO, we introduce importance sampling from behavior policies $\pi_{\theta_{\text{old}}}$ and $\pi_{\phi_{\text{old}}}$. A naive joint importance weight $\frac{\pi_\theta \pi_\phi}{\pi_{\theta_{\text{old}}} \pi_{\phi_{\text{old}}}}$ suffers from compounded variance. We mitigate this using an \textbf{alternating block-coordinate ascent} approximation.
When updating the MLLM ($\phi$), we fix the Allocator to its behavior policy ($\pi_\theta = \pi_{\theta_{\text{old}}}$), yielding an importance ratio of exactly $1$. The off-policy surrogate gradient for $\phi$ becomes:
\begin{align}
\nabla_\phi \mathcal{J}_{\text{surr}}(\phi)
&= \mathbb{E}_{\pi_{\theta_{\text{old}}}} \mathbb{E}_{\pi_{\phi_{\text{old}}}} \!\left[ \frac{\pi_\phi(\boldsymbol{y} \mid \tilde{\boldsymbol{x}})}{\pi_{\phi_{\text{old}}}(\boldsymbol{y} \mid \tilde{\boldsymbol{x}})} R^{\text{ideal}}_{\boldsymbol{s},\boldsymbol{y}}\, \nabla_\phi \log \pi_\phi(\boldsymbol{y} \mid \tilde{\boldsymbol{x}}) \right]. \label{eq:grad_phi_is}
\end{align}
Applying the log-derivative identity $\nabla_\phi r_\phi = r_\phi \nabla_\phi \log \pi_\phi$ (where $r_\phi = \pi_\phi / \pi_{\phi_{\text{old}}}$), we derive the surrogate objective:
\begin{equation}
\mathcal{L}_\phi^{\text{ideal}} = \mathbb{E}_{\pi_{\theta_{\text{old}}}} \mathbb{E}_{\pi_{\phi_{\text{old}}}} \!\left[ r_\phi(\boldsymbol{y} \mid \tilde{\boldsymbol{x}}) R^{\text{ideal}}_{\boldsymbol{s},\boldsymbol{y}} \right].
\end{equation}
Policy-gradient ascent on $\phi$ maximizes $\mathcal{L}_\phi^{\text{ideal}}$. Sec.~\ref{sec:training} implements the clipped PPO surrogate, substituting $R^{\text{ideal}}$ with advantages.

Conversely, when updating the Allocator ($\theta$), we fix the backbone to its behavior policy ($\pi_\phi=\pi_{\phi_{\text{old}}}$). The corresponding ideal allocator surrogate is:
\begin{equation}
\mathcal{L}_\theta^{\text{ideal}}
=
\mathbb{E}_{\pi_{\theta_{\text{old}}}}
\!\left[
r_\theta(\boldsymbol{s}\mid\boldsymbol{x})\,R^{\text{ideal}}_{\boldsymbol{s}}
\right], \qquad
r_\theta(\boldsymbol{s}\mid\boldsymbol{x})
=
\frac{\pi_\theta(\boldsymbol{s}\mid\boldsymbol{x})}{\pi_{\theta_{\text{old}}}(\boldsymbol{s}\mid\boldsymbol{x})},
\end{equation}
where \(R^{\text{ideal}}_{\boldsymbol{s}}=\mathbb{E}_{\pi_{\phi_{\text{old}}}(\boldsymbol{y}\mid\tilde{\boldsymbol{x}})}[R^{\text{ideal}}_{\boldsymbol{s},\boldsymbol{y}}]\). In practice, we approximate this expectation using Monte Carlo rollouts under the frozen backbone.

\textbf{Sequential allocator--backbone updates within one iteration.}
The alternating derivation above fixes one policy while updating the other, rendering the inactive policy's importance ratio unity. In implementations that first update the Allocator from $\theta_{\text{old}}$ to $\theta'$ and subsequently update the MLLM on the \emph{same} rollout batch, trajectories originate from the behavior pair $(\theta_{\text{old}},\phi_{\text{old}})$. The MLLM gradient evaluation occurs under $\phi$ at fixed $(\boldsymbol{x},\boldsymbol{a},\boldsymbol{y})$. The importance weight $\omega_{\theta}=q_{\theta'}(\boldsymbol{a}\mid\boldsymbol{x})/q_{\theta_{\text{old}}}(\boldsymbol{a}\mid\boldsymbol{x})=\pi_{\theta'}(\boldsymbol{s}\mid\boldsymbol{x})/\pi_{\theta_{\text{old}}}(\boldsymbol{s}\mid\boldsymbol{x})$ corrects the shift in the marginal allocation distribution. Multiplying rollout-level advantages by $\omega_{\theta}$ prior to the token-level PPO surrogate for $\phi$ implements the standard importance-sampling correction. This matches the practical ``\texttt{ispred}'' path in our codebase.

\textbf{Advantage Shaping and Monte Carlo Surrogates.}
The ideal linear penalty $-\lambda C(\boldsymbol{s})$ inside $R^{\text{ideal}}$ frequently triggers catastrophic collapse to minimum budgets. CAPO mitigates this by replacing the raw reward with a cost-shaped, group-normalized advantage $A_{\boldsymbol{s},\boldsymbol{y}}$ (denoted $A_{m,n}$ in the main text). This substitution is \emph{not} an unbiased baseline transformation of $R^{\text{ideal}}_{\boldsymbol{s},\boldsymbol{y}}$. Instead, it constitutes a deliberately biased surrogate objective that sacrifices exact fidelity to the Lagrangian reward in exchange for reduced variance and robust budget control.

Applying PPO clipping to the exact joint ratios would entangle all frame- and token-level factors, yielding prohibitive noise. We therefore adopt decoupled objectives. For a batch of $M$ allocations and $N$ rollouts per allocation, the MLLM sequence-level surrogate is:
\begin{equation}
\mathcal{L}_{\phi}^{\text{seq}} = -\frac{1}{MN}\sum_{m=1}^M \sum_{n=1}^N \min\!\Big(r_\phi^{(m,n)}\, A_{m,n},\; \operatorname{clip}(r_\phi^{(m,n)}, 1{-}\varepsilon, 1{+}\varepsilon)\, A_{m,n}\Big).
\end{equation}
This sequence-level loss remains approximate by substituting the CAPO-shaped advantage for the ideal reward. To enable finer credit assignment for the autoregressive MLLM, we factorize $\pi_\phi(\boldsymbol{y}\mid\tilde{\boldsymbol{x}})$ into token-level probabilities, distribute the rollout-level advantage $A_{m,n}$ across all tokens, and average over the sequence length $L_{m,n}$. Equation~\eqref{eq:loss_phi} represents the standard token-level PPO approximation to this sequence-level surrogate.

When updating the Allocator ($\theta$), we fix the MLLM ($\pi_\phi = \pi_{\phi_{\text{old}}}$) and apply the aggregated advantage $A^{\text{CAPO}}_m = \frac{1}{N} \sum_n A_{m,n}$. Since the Allocator's output distribution factorizes conditionally across frames (Eq.~\ref{eq:logprob}), its score function decomposes additively:
\begin{equation}
\nabla_\theta \log \pi_\theta(\boldsymbol{s}^{(m)} \mid \boldsymbol{x}) = \sum_{t=1}^{T} \nabla_\theta \log \mathrm{Beta}(a_t^{(m)};\, \alpha_t, \beta_t).
\end{equation}
This additive log-probability structure facilitates low-variance frame-level credit assignment. Equation~\eqref{eq:loss_theta} provides a practical approximation to a trajectory-level clipped objective. We deploy this per-frame surrogate to guarantee stability in large-scale training.

\subsection{Complexity Analysis}
\label{sec:complexity}

We derive formal computational bounds for \our{} to establish when Allocator overhead becomes negligible compared to backbone savings. For clarity, we assume a standard Transformer backbone with quadratic self-attention and a uniform native resolution $H \times W$ over $T$ frames. Replacing $HW$ with per-frame products $H_tW_t$ extends this immediately to heterogeneous resolutions.

\textbf{Baseline cost.}
Let $P$ denote the ViT patch size. A vanilla MLLM encoding $T$ frames at full resolution generates a visual token count of:
\begin{equation}
N_0 \;=\; T \cdot \left\lceil \frac{H}{P} \right\rceil \left\lceil \frac{W}{P} \right\rceil \;\approx\; \frac{THW}{P^2}.
\end{equation}

\textbf{Adaptive cost and token retention ratio.}
In our resize instantiation, the operator rescales frame $f_t$ by factor $s_t \in [s_{\min}, s_{\max}]$, producing $n_t(s_t) = \lceil s_t H/P \rceil \lceil s_t W/P \rceil \approx s_t^2 \cdot HW/P^2$ tokens. Summing across the sequence and normalizing by $N_0$ yields the \emph{token retention ratio}:
\begin{equation}
N^{\mathrm{adapt}} = \sum_{t=1}^T n_t(s_t) \;\approx\; \frac{HW}{P^2}\sum_{t=1}^T s_t^2, \qquad
\rho \;\triangleq\; \frac{N^{\mathrm{adapt}}}{N_0} \;=\; \frac{1}{T}\sum_{t=1}^T s_t^2.
\end{equation}
Because the learned Beta policy concentrates redundant frames near $s_{\min}$ (Figure~\ref{fig:scale_aggregate}), $\rho$ remains substantially smaller than $1$. Across our evaluation suite, $\rho \in [0.06,\,0.16]$.

\textbf{Quadratic FLOPs reduction.}
For an $L_{\mathrm{mllm}}$-layer MLLM with hidden dimension $D_{\mathrm{mllm}}$, self-attention cost scales quadratically with visual sequence length: $\Phi(N) = O(L_{\mathrm{mllm}} N^2 D_{\mathrm{mllm}})$. Substituting $N^{\mathrm{adapt}} = \rho \cdot N_0$ gives:
\begin{equation}
\Phi_{\mathrm{mllm}}^{\mathrm{adapt}} \;=\; O\!\left(L_{\mathrm{mllm}} \cdot \rho^2 N_0^2 \cdot D_{\mathrm{mllm}}\right).
\end{equation}
This reflects a reduction by a factor of $\rho^2$ relative to full-resolution processing. At a representative operating point of $\rho = 0.11$, we achieve $\rho^2 \approx 0.012$, eliminating roughly $83\times$ of the backbone attention FLOPs.

\textbf{Allocator overhead.}
The Allocator processes $N_c = T \cdot \lceil H/P_c \rceil \lceil W/P_c \rceil$ coarsely pooled tokens across $L_{\mathrm{pred}}$ layers with dimension $D_{\mathrm{pred}}$, utilizing a coarse spatial stride $P_c \gg P$. Its computational cost and relative overhead are:
\begin{equation}
\Phi_{\mathrm{pred}} = O\!\left(L_{\mathrm{pred}} \cdot N_c^2 \cdot D_{\mathrm{pred}}\right), \qquad
\frac{\Phi_{\mathrm{pred}}}{\Phi_{\mathrm{mllm}}^{\mathrm{base}}} \;=\; O\!\left(\frac{L_{\mathrm{pred}}\, D_{\mathrm{pred}}}{L_{\mathrm{mllm}}\, D_{\mathrm{mllm}}} \cdot \left(\frac{P}{P_c}\right)^{\!4}\right) \ll 1.
\end{equation}
Inserting our implementation parameters ($P_c {=} 14$, $L_{\mathrm{pred}} {=} 4$, $D_{\mathrm{pred}} {=} 1{,}024$ versus $L_{\mathrm{mllm}} {=} 28$, $D_{\mathrm{mllm}} {=} 3{,}584$), the Allocator consumes less than $3\%$ of total inference FLOPs. The decision stage overhead remains trivial compared to the backbone computation it bypasses.

\textbf{Net speedup.}
Combining these bounds under the first-order approximation $\Phi_{\mathrm{mllm}}^{\mathrm{base}} \gg \Phi_{\mathrm{pred}}$:
\begin{equation}
\mathrm{Speedup} \;\approx\; \frac{\Phi_{\mathrm{mllm}}^{\mathrm{base}}}{\Phi_{\mathrm{mllm}}^{\mathrm{adapt}} + \Phi_{\mathrm{pred}}} \;\approx\; \frac{N_0^2}{(N^{\mathrm{adapt}})^2} \;=\; \frac{1}{\rho^2}.
\end{equation}
At $\rho = 0.11$, this dictates a theoretical $83\times$ acceleration in backbone attention.

\textbf{Temporal context scaling.}
These savings translate directly into expanded \emph{temporal coverage}. Given a strict token budget $B$, a vanilla MLLM accommodates only $T_0 = BP^2/(HW)$ full-resolution frames. \our{} processes $T_0/\rho$ adaptively resized frames within the identical budget. This yields a $1/\rho \approx 6$--$16\times$ expansion in temporal horizon, unlocking the long-context performance detailed in Sec.~\ref{sec:pareto}.

\textbf{Acceleration transparency.}
Input-side adaptation guarantees the backbone receives a standard, albeit shorter, visual-token sequence. Consequently, \our{} integrates seamlessly with optimized attention kernels like FlashAttention, vLLM~\citep{kwon2023efficient}, and SGLang~\citep{zheng2024sglang} without necessitating low-level modifications. Conversely, model-side pruning and merging strategies introduce irregular token layouts that disrupt these kernels, demanding bespoke engineering fallbacks.

\section{Implementation Details}
\label{appendix:implementation_details}

\subsection{Training Data}
\label{appendix:data}

\textbf{Data Composition.}
We construct the training corpus from the difficulty-filtered VideoAuto-R1~\citep{liu2026videoauto} dataset, strictly retaining image and video samples while discarding pure-text examples. To guarantee robust coverage of visually demanding subdomains, we inject 16{,}500 high-complexity video instances from Video-R1~\citep{feng2025videor1}, prioritizing OCR, free-form QA, and regression-style tasks. The finalized pool comprises approximately 93.4K training samples. We rigorously purge all evaluation examples from this corpus to preclude data leakage.

\subsection{Training Configuration}
\label{appendix:training_config}

We train the models for one epoch using AdamW with a global batch size of 128. We apply a learning rate of \(2\times 10^{-5}\) to the Allocator and \(1\times 10^{-6}\) to the backbone. We enforce a weight decay of \(0.01\) and cap gradient clipping at \(1.0\). We constrain the maximum video token budget to \(8{,}192\), sample \(T{=}128\) frames during training, and bound the scale factors within \([s_{\min},s_{\max}] = [0.2,1.8]\). This range explicitly enables both aggressive downscaling and selective upscaling. CAPO samples \(M{=}16\) allocation trajectories per prompt and executes \(N{=}1\) rollout per trajectory. We orchestrate training across 32 H100 GPUs utilizing VeRL~\citep{sheng2025hybridflow}, DeepSpeed~\citep{rasley2020deepspeed}, and vLLM~\citep{kwon2023efficient}. We execute evaluations via lmms-eval~\citep{zhang2024lmmsevalrealitycheckevaluation}. We truncate standard response lengths at 256 tokens and extend the limit to 4{,}096 tokens for reasoning models.

\subsection{Reward Design}
\label{appendix:reward_design}

We detail the reward structures supplementing Sec.~\ref{sec:capo}. The base scalar reward \(R^{\text{task}}_{m,n}\) isolates task-specific performance. Efficiency constraints manifest through CAPO advantage shaping rather than primitive additive reward terms.

\textbf{Base Task Reward ($R^{\text{task}}_{m,n}$).}
We define objective metrics for four task types:
\begin{itemize}
\item \textit{Question Answering.} For math problems, we extract the numeric answer and tolerate a $10^{-2}$ deviation from the ground truth. For multiple-choice questions, we parse the exact option letter. For standard QA, we execute exact string matching post-normalization (case-folding and whitespace stripping). This yields a binary reward $R_{\text{QA}}(\hat{o},o) \in \{0,1\}$.
\item \textit{Free-form Generation.} We quantify open-ended generation quality via the ROUGE-L score between the prediction \(\hat{o}\) and the reference \(o\): $R_{\text{Gen}}(\hat{o},o)=\text{ROUGE-L}(\hat{o},o)\in[0,1]$.
\item \textit{Temporal Grounding.} Let $\mathcal{G}=\{[s_j,e_j]\}_j$ denote ground-truth segments and $\widehat{\mathcal{G}}=\{[\hat{s}_k,\hat{e}_k]\}_k$ denote predictions. We isolate the highest temporal IoU across all pairs: $R_{\text{TG}}(\widehat{\mathcal{G}},\mathcal{G}) = \max_{[\hat{s},\hat{e}]\in \widehat{\mathcal{G}},\;[s,e]\in \mathcal{G}} \mathrm{tIoU}\!\left([\hat{s},\hat{e}],[s,e]\right)\in[0,1]$. Invalid parsings default to $0$.
\item \textit{Grounding QA.} We parse both the textual response and temporal segments, summing their independent scores: $R_{\text{GQA}}(\hat{o},\widehat{\mathcal{G}};o,\mathcal{G}) = R_{\text{QA}}(\hat{o},o)+R_{\text{TG}}(\widehat{\mathcal{G}},\mathcal{G}) \in[0,2]$.
\end{itemize}

These metrics establish the scalar base reward \(R^{\text{task}}_{m,n}\). CAPO subsequently defines a binary success indicator \(u_{m,n}\in\{0,1\}\). Exact-match QA utilizes the binary outcome directly. Continuous metrics (ROUGE-L, temporal IoU, Grounding QA) apply a strict \(0.35\) threshold. Format validation, when active, injects a weighted penalty prior to GRPO normalization, while \(u_{m,n}\) strictly isolates the task metric.

\textbf{Format Reward.}
We enforce a binary format reward \(R_{\text{fmt}}(\hat{o})\in\{0,1\}\) via rigid regex validation. The generation must output exactly one \texttt{<think>...</think>} block and one \texttt{<answer>...</answer>} block. The final answer must reside within \verb|\\boxed{...}| inside the \texttt{<answer>} tags. Malformed outputs trigger a penalty, integrating into the scalar reward with a \(0.2\) weight.

\subsection{Prompt Template}
\label{appendix:prompt_template}

We deploy the standard GRPO training prompt (Table~\ref{table:prompt_template}). The model must encapsulate its reasoning trace within \verb|<think> </think>| tags. While optional for \our{} (as the MLLM $\pi_\phi$ internalizes reasoning), we mandate this structure to ensure parity with reasoning-based baselines. The final answer must emerge encased in \verb|\\boxed{}|.

\begin{table*}[h!]
    \centering
    \caption{\textbf{Prompt template used for CAPO training.} The template presents video frames and the task question, requires intermediate reasoning inside \texttt{<think>} tags, and places the final answer in \texttt{\textbackslash boxed\{\}} within \texttt{<answer>} tags. This structure enables automatic reward extraction from MLLM outputs.}
    \label{table:prompt_template}
    \begin{tcolorbox}[
        enhanced,
        colback=cyan!2,
        colframe=black,
        colbacktitle=blue!5!gray!10,
        coltitle=black,
        fonttitle=\bfseries\Large\centering,
        title=Prompt Template for Training with Thinking,
        arc=3mm,
        boxrule=0.8pt,
        halign title=center,
        toptitle=2mm,
        bottomtitle=2mm,
        left=4mm, right=4mm, top=4mm, bottom=4mm
    ]
    \textbf{System Prompt:}\\
    You are a helpful assistant. \\
    You FIRST think about the reasoning process as an internal monologue and then provide the final answer. \\
    The reasoning process MUST BE enclosed within \textcolor{red}{\texttt{<think> </think>}} tags and the answer MUST BE enclosed within \textcolor{red}{\texttt{<answer> </answer>}} tags. \\
    The final answer MUST BE put in \textcolor{blue}{\texttt{\textbackslash boxed\{\}}} and the \textcolor{blue}{\texttt{\textbackslash boxed\{\}}} expression MUST BE contained entirely within the \textcolor{red}{\texttt{<answer> </answer>}} tags. \\
    Do not include any reasoning or explanations outside these tags.
    \end{tcolorbox}
\end{table*}

\section{Extended Analysis}
\label{app:scale_analysis}

This section extends the analysis of Sec.~\ref{sec:analysis}. We first examine the learned allocation policy at per-benchmark, per-duration, and per-category granularity (Sec.~\ref{app:behavioral_analysis}), then present extended ablations on the temporal regularizer and reward design (Sec.~\ref{app:ablation_studies}), qualitative case studies linking allocation to reasoning outcomes (Sec.~\ref{app:case_studies}), and a boundary-case transfer test beyond video (Sec.~\ref{app:image_transfer}). Unless otherwise noted, all plots use Qwen2.5-VL-7B processing 32 uniformly sampled frames.

\subsection{Extended Scale Policy Analysis}
\label{app:behavioral_analysis}

\begin{figure}[t]
    \centering
    \includegraphics[width=\linewidth]{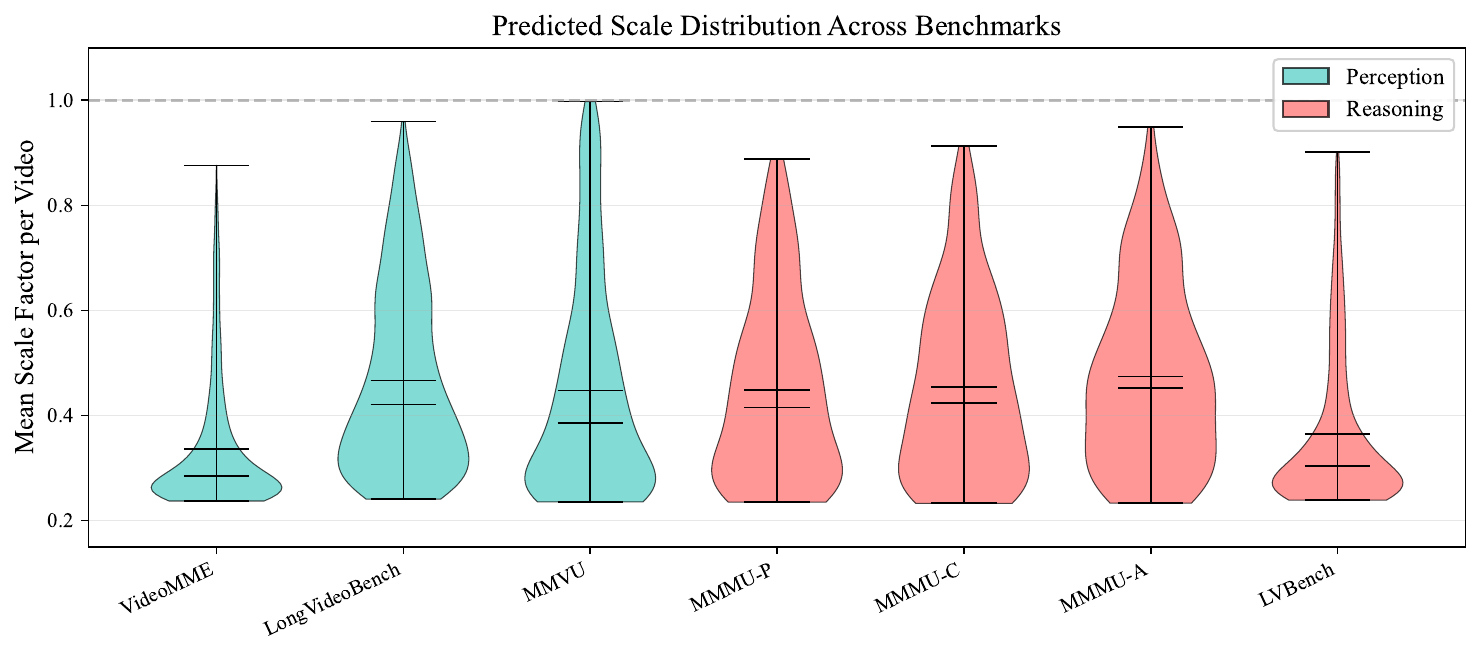}
    \caption{\textbf{Per-video mean scale across benchmarks.} Kernel density estimates of the per-video mean scale $\bar{s}$. Reasoning-heavy benchmarks shift toward larger $\bar{s}$ than perception-heavy ones, indicating that the learned policy spends more fidelity where fine-grained evidence is more likely to matter.}
    \label{fig:scale_dist_bench}
\end{figure}

\textbf{Benchmark-level budget allocation.}
Figure~\ref{fig:scale_dist_bench} reveals a clear benchmark-level ordering, despite the policy operating without any knowledge of benchmark identity during training. Reasoning-intensive tasks consistently command higher mean scales than perception-oriented tasks (0.435 vs.~0.417). MMMU-Adaptation anchors the high-fidelity extreme, while VideoMME anchors the low-fidelity extreme. The policy does not apply a fixed compression ratio; it calibrates its operating point to the anticipated visual complexity of each task.

\begin{figure*}[t]
    \centering
    \includegraphics[width=\linewidth]{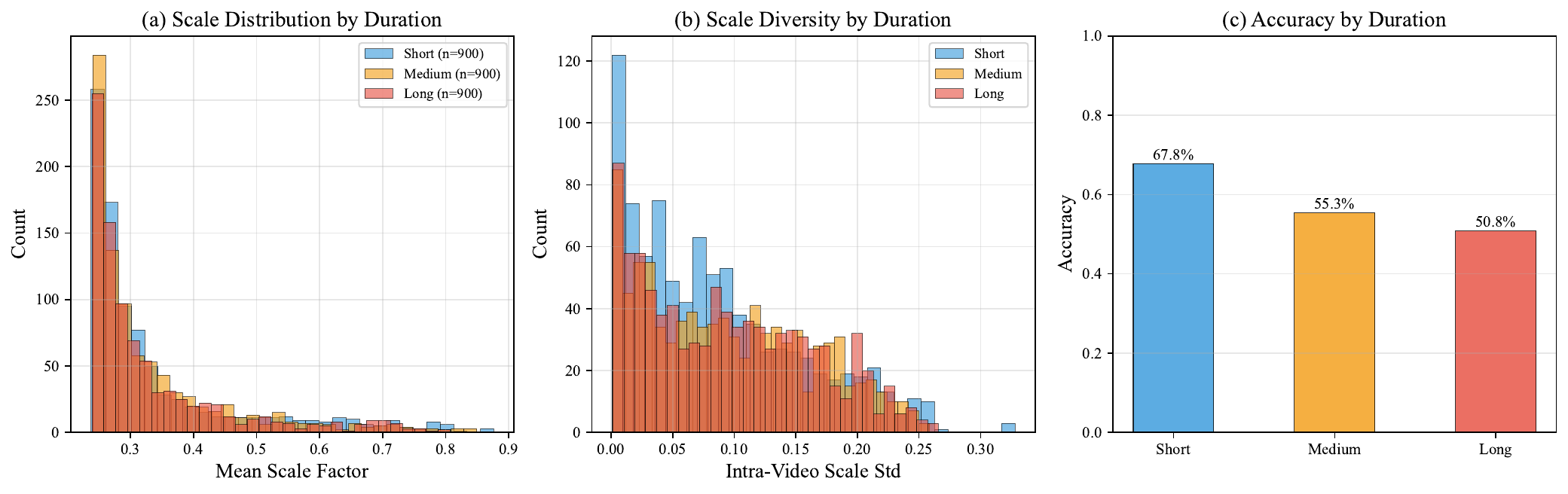}
    \caption{\textbf{VideoMME broken down by video duration.} As clip duration grows, the policy lowers the average scale, increases within-video scale diversity, and faces lower task accuracy. Longer clips are therefore processed more aggressively and more selectively.}
    \label{fig:duration_analysis}
\end{figure*}

\textbf{Long-context behavior.}
Figure~\ref{fig:duration_analysis} clarifies how the policy handles increasing clip length. As duration grows, the mean scale decreases (0.342$\!\to\!$0.336$\!\to\!$0.332) while within-video diversity simultaneously increases (0.085$\!\to\!\sim$0.095). The policy compresses longer videos more aggressively overall, but with far greater selectivity---precisely the regime where uniform resizing fails.

\begin{figure*}[t]
    \centering
    \includegraphics[width=0.76\linewidth]{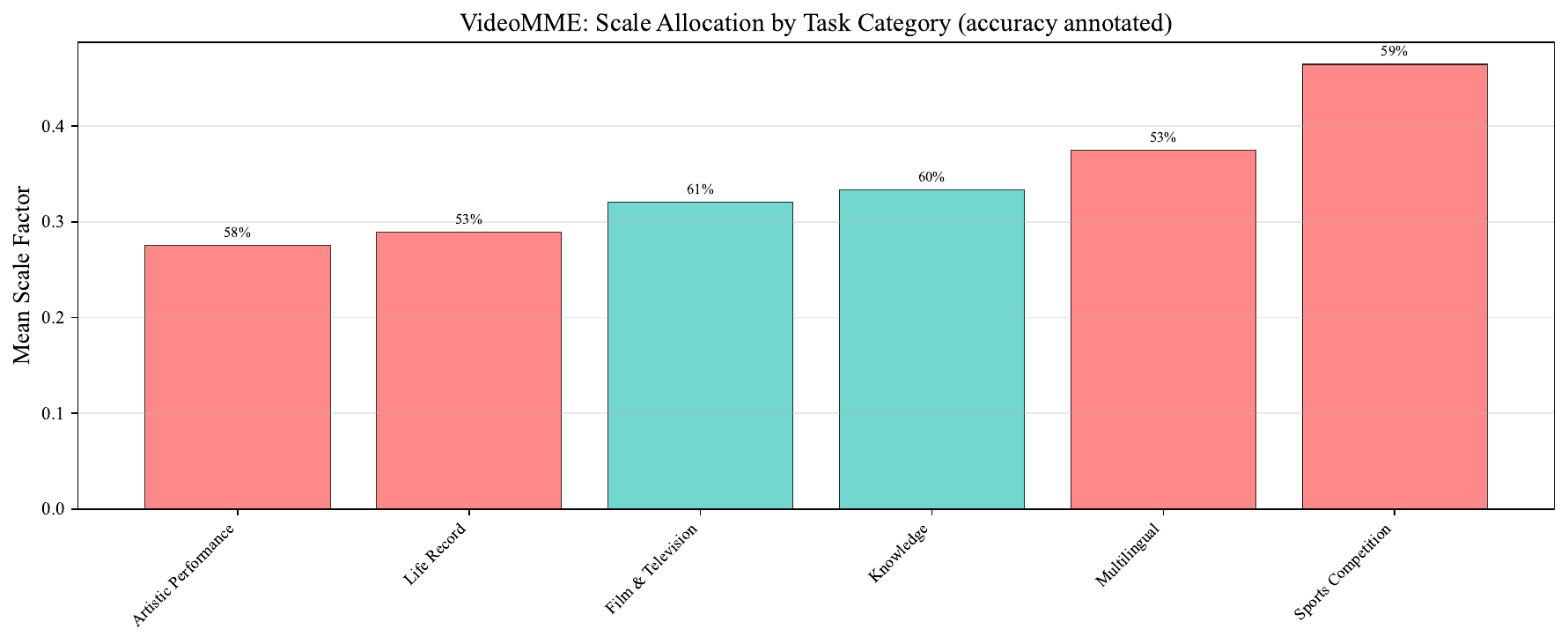}
    \caption{\textbf{Scale allocation by VideoMME task category.} Mean $\bar{s}$ varies substantially across categories, with larger budgets assigned to categories that contain crowded motion or finer local evidence. Accuracy annotations show that allocation is not a trivial proxy for which category is easiest.}
    \label{fig:category_scale}
\end{figure*}

Figure~\ref{fig:category_scale} provides a category-level view within VideoMME. The policy assigns maximum budget to \textit{Sports Competition} (dense, high-motion) and minimum budget to \textit{Artistic Performance} (visually sparse). Allocation tracks spatial complexity rather than task difficulty.

\begin{figure*}[t]
    \centering
    \includegraphics[width=\linewidth]{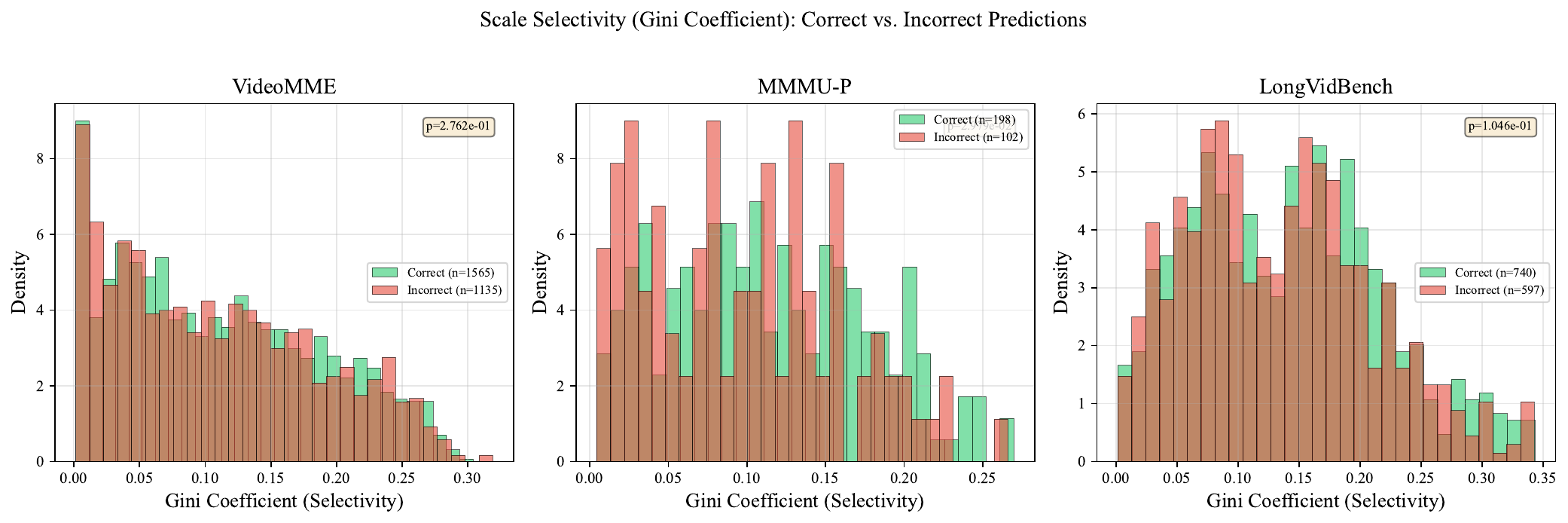}
    \caption{\textbf{Selectivity versus prediction correctness on three representative benchmarks.} Per-video Gini coefficients of the frame-level scales. Correct predictions tend to have higher Gini than incorrect ones, linking success to sharper concentration of resolution rather than merely larger average budgets.}
    \label{fig:gini}
\end{figure*}

\textbf{Selectivity and success.}
We measure frame-level selectivity via the Gini coefficient of predicted scales: a high Gini indicates concentrated budget on a sparse subset of frames. Figure~\ref{fig:gini} shows that correct predictions consistently map to higher selectivity, peaking on MMMU-P. Success correlates not with larger average budgets, but with sharper concentration of resolution onto the decisive frames.

\begin{figure*}[t]
    \centering
    \begin{subfigure}[t]{0.6\textwidth}
        \centering
        \includegraphics[width=\linewidth,height=0.30\textheight,keepaspectratio]{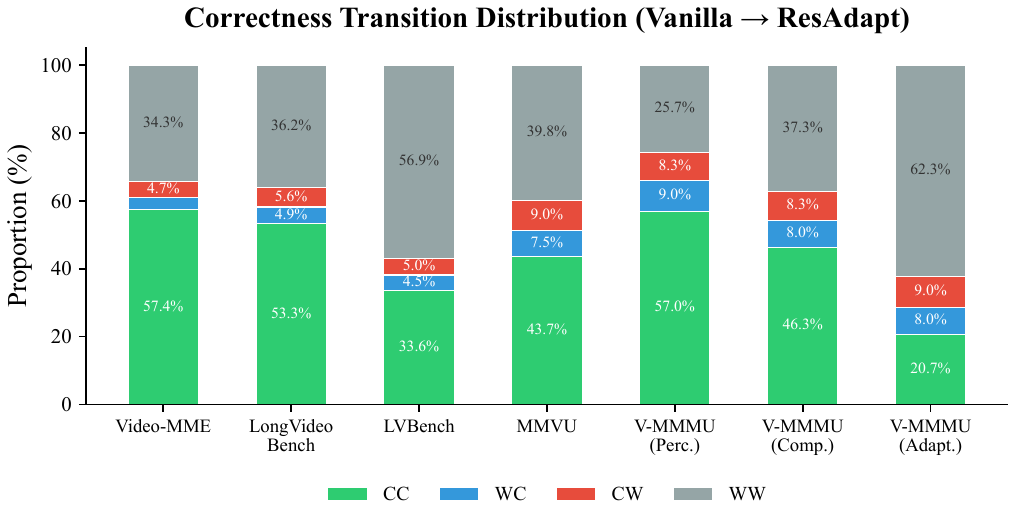}
    \end{subfigure}
    \hfill
    \begin{subfigure}[t]{0.38\textwidth}
        \centering
        \includegraphics[width=\linewidth,height=0.30\textheight,keepaspectratio]{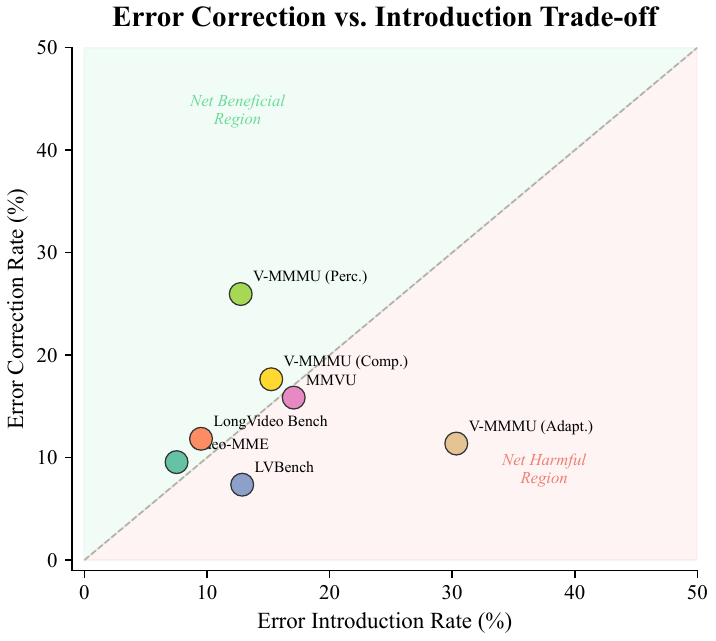}
    \end{subfigure}
    \caption{\textbf{Sample-level robustness at 25\% retention.} Most originally correct predictions remain correct, but corrected and newly introduced errors are of comparable magnitude. Adaptive allocation is therefore selective rather than lossless.}
    \label{fig:error_dynamics}
\end{figure*}

\textbf{Robustness and failure modes.}
Figure~\ref{fig:error_dynamics} examines whether adaptive compression preserves correct reasoning paths or merely reshuffles errors. Prediction stability is robust: $89\%$ of originally correct samples survive compression. However, error correction and error induction rates remain comparable. The policy executes \emph{selective redistribution}---rescuing certain failures by magnifying critical details, while occasionally destroying fine-grained evidence when the decisive cue is transient or visually inconspicuous.

\subsection{Extended Ablation Studies}
\label{app:ablation_studies}

\begin{figure*}[t]
    \centering
    \includegraphics[width=0.95\linewidth]{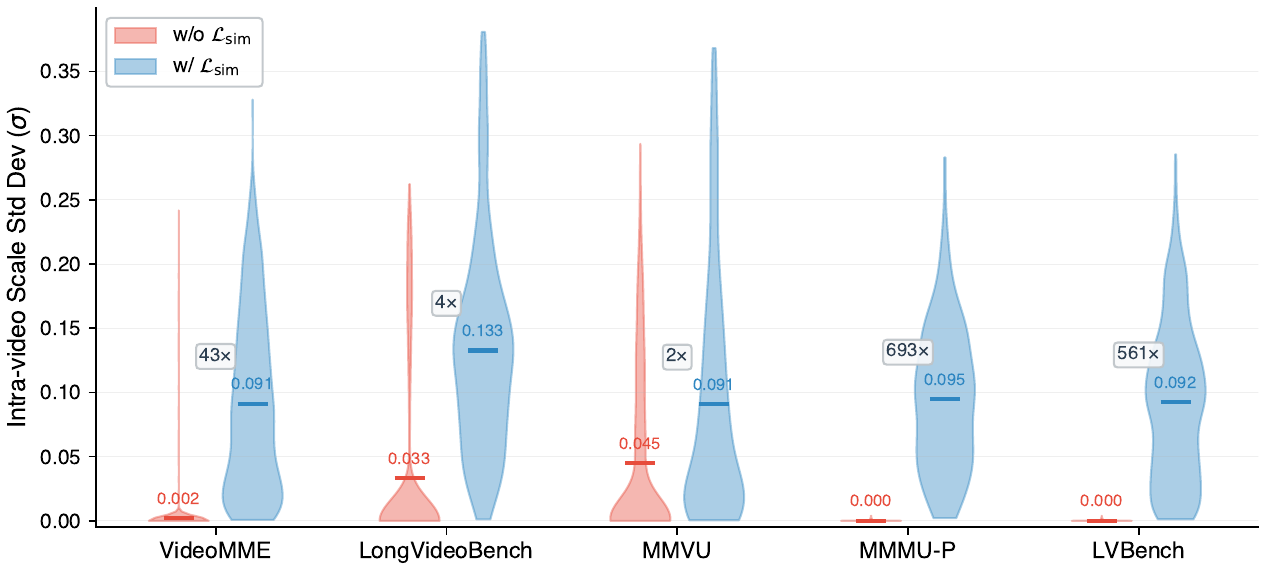}
    \caption{\textbf{Cross-benchmark scale diversity with and without $\mathcal{L}_{\text{sim}}$.} Per-video scale standard deviation $\sigma$ across five benchmarks. Without the regularizer, diversity collapses toward zero; adding $\mathcal{L}_{\text{sim}}$ restores broad within-video variation on every benchmark.}
    \label{fig:sim_diversity_bench}
\end{figure*}

\textbf{Temporal similarity: cross-benchmark view.}
Figure~\ref{fig:sim_diversity_bench} isolates the impact of $\mathcal{L}_{\text{sim}}$. Without it, scale diversity collapses to near-zero across all benchmarks ($\sigma < 0.003$). Activating $\mathcal{L}_{\text{sim}}$ restores within-video variation by $4\times$--$693\times$. CAPO controls the global budget level; $\mathcal{L}_{\text{sim}}$ breaks the uniform-scale equilibrium.

\begin{figure*}[t]
    \centering
    \includegraphics[width=\linewidth]{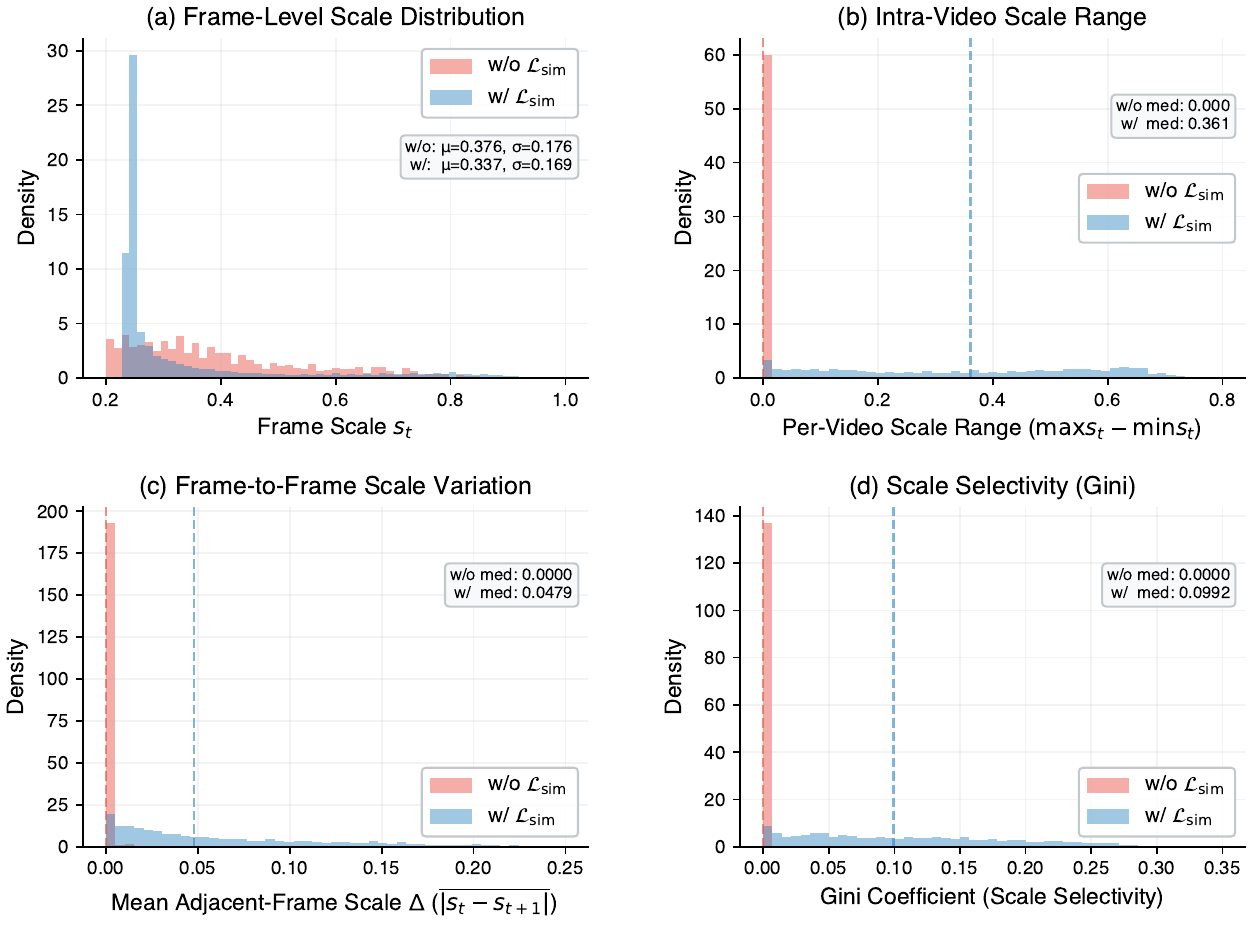}
    \caption{\textbf{Four diagnostics of the $\mathcal{L}_{\text{sim}}$ ablation on VideoMME.} With the regularizer, the frame-scale histogram becomes bimodal, the per-video range expands, adjacent-frame variation increases, and the Gini coefficient rises. The policy moves from near-uniform allocation to a genuinely selective regime.}
    \label{fig:sim_ablation_panel}
\end{figure*}

\textbf{Temporal similarity: structural diagnostics.}
Figure~\ref{fig:sim_ablation_panel} provides four complementary views. With the regularizer active, the frame-scale histogram becomes bimodal, the per-video range expands, adjacent-frame variation increases, and the Gini coefficient rises. The policy transitions from a degenerate uniform allocator to a genuinely selective one.

\begin{figure*}[t]
    \centering
    \includegraphics[width=\textwidth]{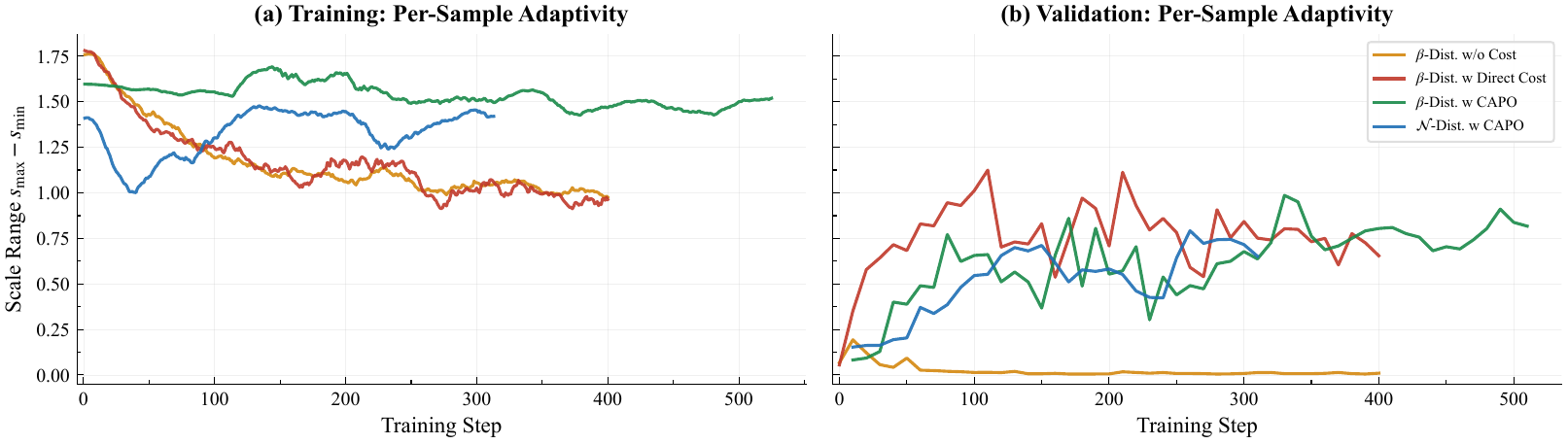}
    \caption{\textbf{Per-sample scale adaptivity under different reward designs.} Scale range $s_{\max}-s_{\min}$ over training on \textbf{(a)} training and \textbf{(b)} validation splits. CAPO keeps a non-trivial adaptive range, whereas direct cost collapses and cost-free training saturates.}
    \label{fig:reward_adaptivity}
\end{figure*}

\textbf{Reward design: adaptivity.}
Figure~\ref{fig:reward_adaptivity} tracks the per-sample scale range $s_{\max}-s_{\min}$ across training. CAPO maintains robust adaptivity on the validation split. Direct cost penalties collapse the range to zero, while cost-free training saturates at a uniform high-scale plateau.

\begin{figure*}[t]
    \centering
    \includegraphics[width=\linewidth]{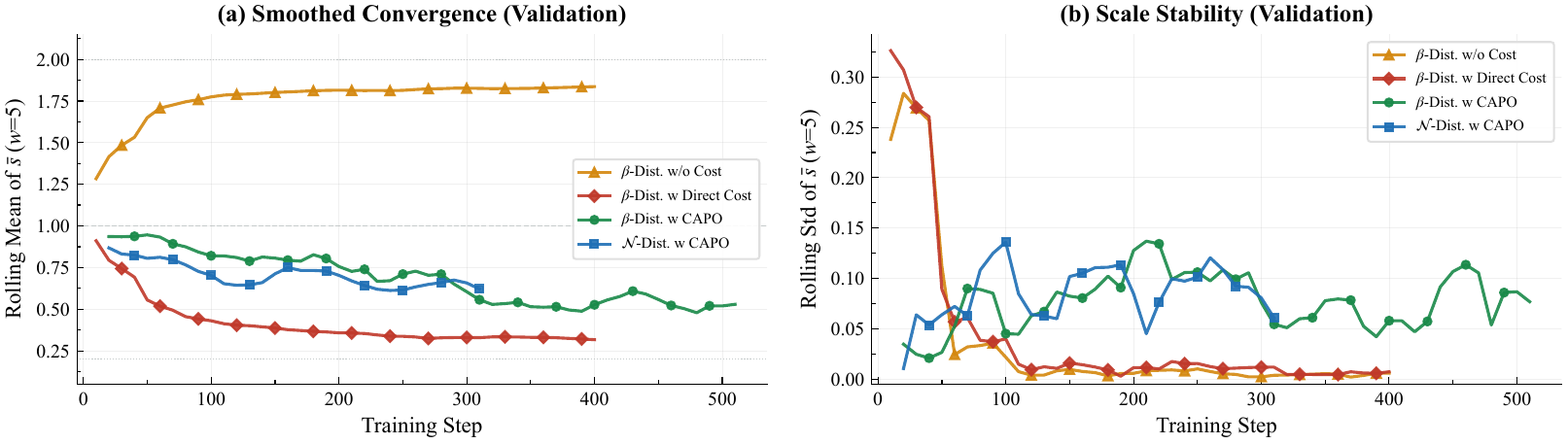}
    \caption{\textbf{Validation-time convergence under different reward designs.} CAPO variants converge to stable intermediate operating points, while cost-free training saturates at the upper boundary and direct cost collapses to the lower boundary. Stability alone is not sufficient; the key is where the policy stabilizes.}
    \label{fig:reward_convergence}
\end{figure*}

\textbf{Reward design: convergence.}
Figure~\ref{fig:reward_convergence} identifies the failure modes. Accuracy-only training saturates near $s_{\max}$, abandoning compression. Direct cost optimization collapses to $s_{\min}$, abandoning task quality. CAPO converges to a stable intermediate operating point that preserves content-adaptive allocation. The critical difference is not stability alone---both degenerate baselines are stable---but \emph{where} the policy stabilizes.

\subsection{Qualitative Case Studies}
\label{app:case_studies}

\begin{figure*}[t]
    \centering
    \begin{tcolorbox}[
        colback=gray!3, colframe=gray!50, arc=1.5mm, boxrule=0.4pt,
        left=3mm, right=3mm, top=1.5mm, bottom=1.5mm, fontupper=\small
    ]
    \textbf{Q:} Evaluate five statements about Urban Geography City Models (concentric zone, Hoyt sector, multiple nuclei, galactic, Latin American); identify which are correct. \textit{Please ignore the Quiz question in last frame of the video.}
    \end{tcolorbox}
    \vspace{-2mm}
    \includegraphics[width=0.76\linewidth]{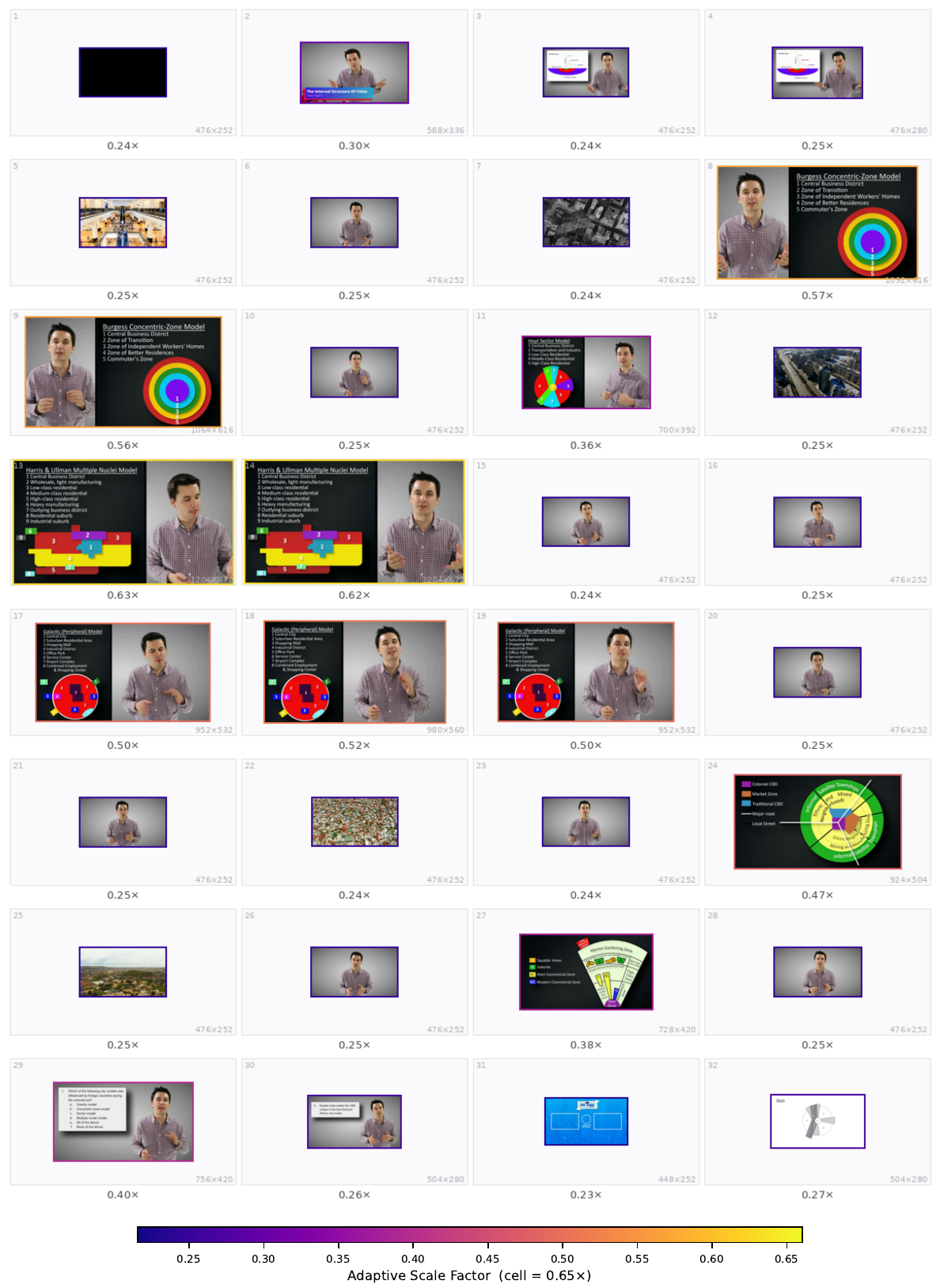}
    \caption{\textbf{Case~1: Video-MMMU Comprehension~\citep{hu2025video} (Vanilla \textcolor{red}{$\times$} $\to$ \our{} \textcolor{green!60!black}{$\checkmark$}).} The policy concentrates resolution on diagram-bearing slide frames, compresses lecturer-only frames, and suppresses the final quiz frame that the prompt explicitly marks as irrelevant.}
    \label{fig:case_comp}
\end{figure*}

\begin{figure*}[t]
    \centering
    \begin{tcolorbox}[
        enhanced, colback=gray!3, colframe=gray!50, arc=1.5mm, boxrule=0.4pt,
        left=3mm, right=3mm, top=1.5mm, bottom=1.5mm, fontupper=\small
    ]
    \textbf{Q:} Watch and learn the video content. Then apply what you learned to answer: Table~11.47 provides a survey of the youngest online entrepreneurs (ages 17--30) whose net worth $\geq$ \$1M. We want to know whether ages and net worth are independent. $\chi^2$ test statistic $=$ \_\_\_\_\_\_
    \end{tcolorbox}
    \vspace{-2mm}
    \includegraphics[width=0.76\linewidth]{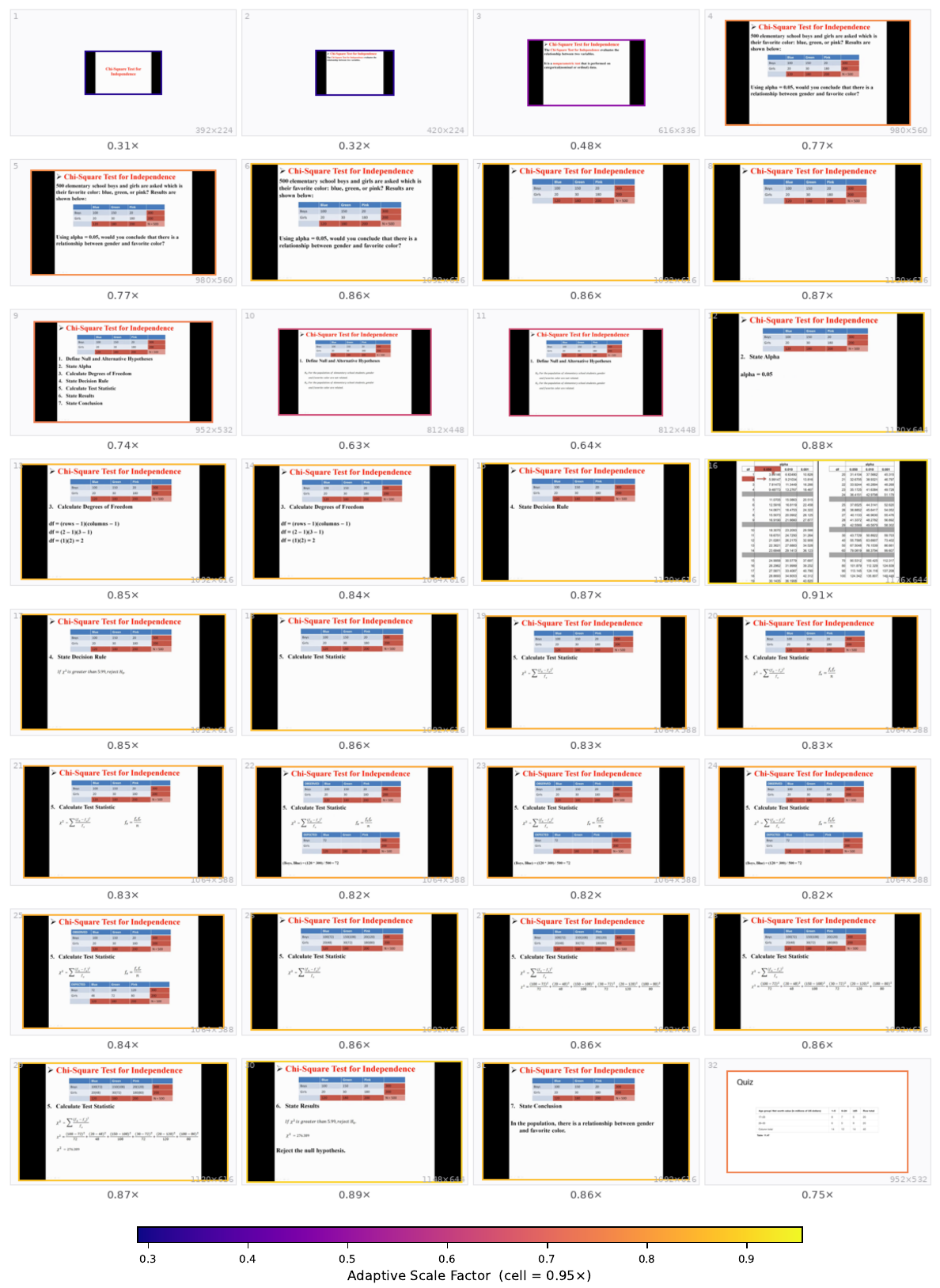}
    \caption{\textbf{Case~2: Video-MMMU Adaptation~\citep{hu2025video} (Vanilla \textcolor{red}{$\times$} $\to$ \our{} \textcolor{green!60!black}{$\checkmark$}).} When the answer depends on reading a numeric table and performing a $\chi^2$ computation, the policy keeps a much higher global budget and strongly upscales the table-bearing frames.}
    \label{fig:case_adapt}
\end{figure*}

\begin{figure*}[t]
    \centering
    \begin{tcolorbox}[
        enhanced, colback=gray!3, colframe=gray!50, arc=1.5mm, boxrule=0.4pt,
        left=3mm, right=3mm, top=1.5mm, bottom=1.5mm, fontupper=\small
    ]
    \textbf{Q:} When is the zodiacal light visible from the video? (A)~Mar.~19, (B)~Mar.~24, (C)~Mar.~25, (D)~Mar.~29.
    \end{tcolorbox}
    \vspace{-2mm}
    \includegraphics[width=0.76\linewidth]{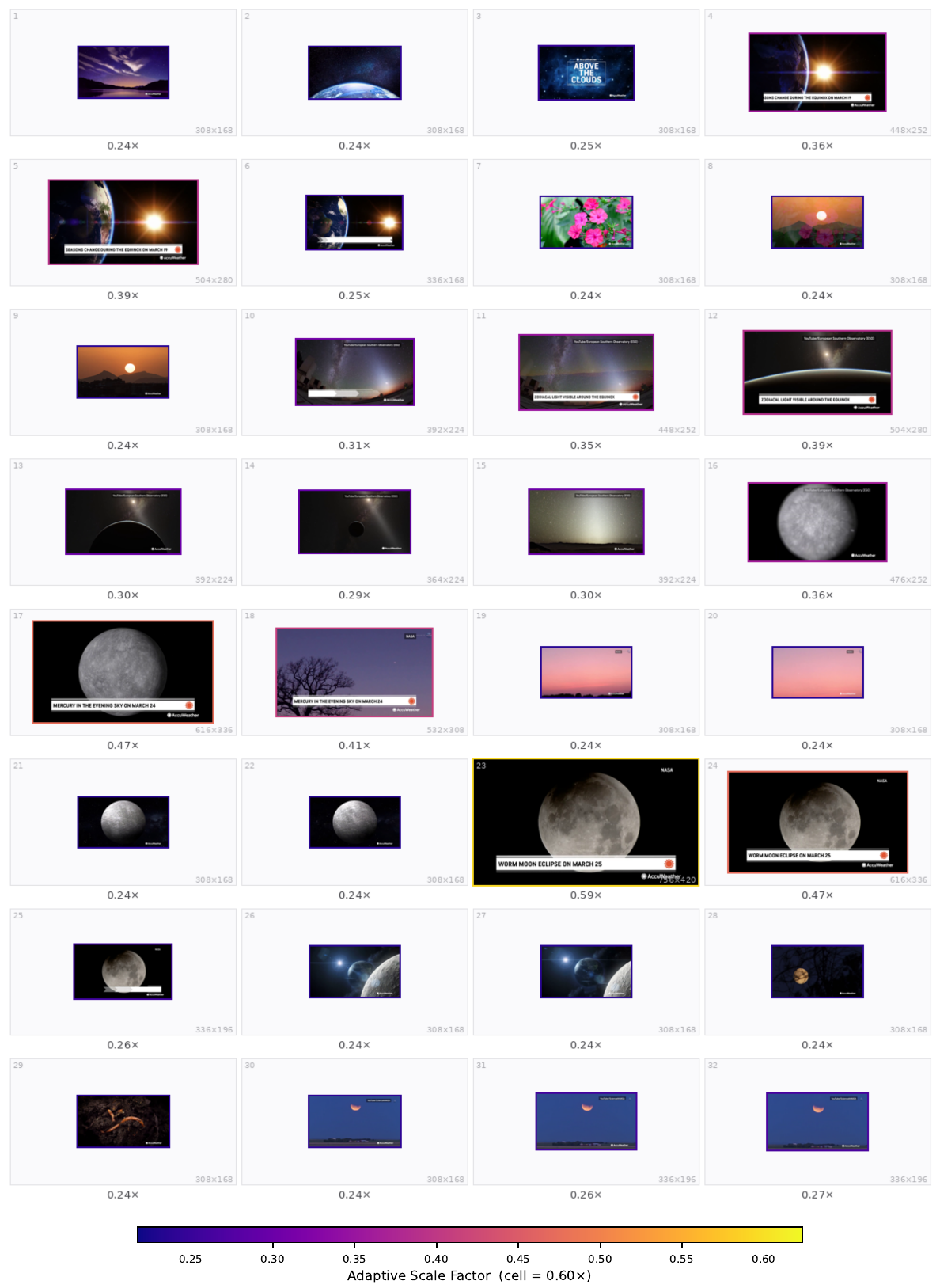}
    \caption{\textbf{Case~3: VideoMME~\citep{fu2025video} (Vanilla \textcolor{red}{$\times$} $\to$ \our{} \textcolor{green!60!black}{$\checkmark$}).} Frames containing the decisive date overlays are enlarged, while the largely homogeneous sky footage is compressed. The policy spends budget on answer-bearing evidence rather than on the surrounding context.}
    \label{fig:case_zodiac}
\end{figure*}

\begin{figure*}[t]
    \centering
    \begin{tcolorbox}[
        enhanced, colback=gray!3, colframe=gray!50, arc=1.5mm, boxrule=0.4pt,
        left=3mm, right=3mm, top=1.5mm, bottom=1.5mm, fontupper=\small
    ]
    \textbf{Q:} Which item does the man throw into the trash at the beginning of the video? (A)~A fork, (B)~A pair of chopsticks, (C)~A box of noodles, (D)~A spoon.
    \end{tcolorbox}
    \vspace{-2mm}
    \includegraphics[width=0.76\linewidth]{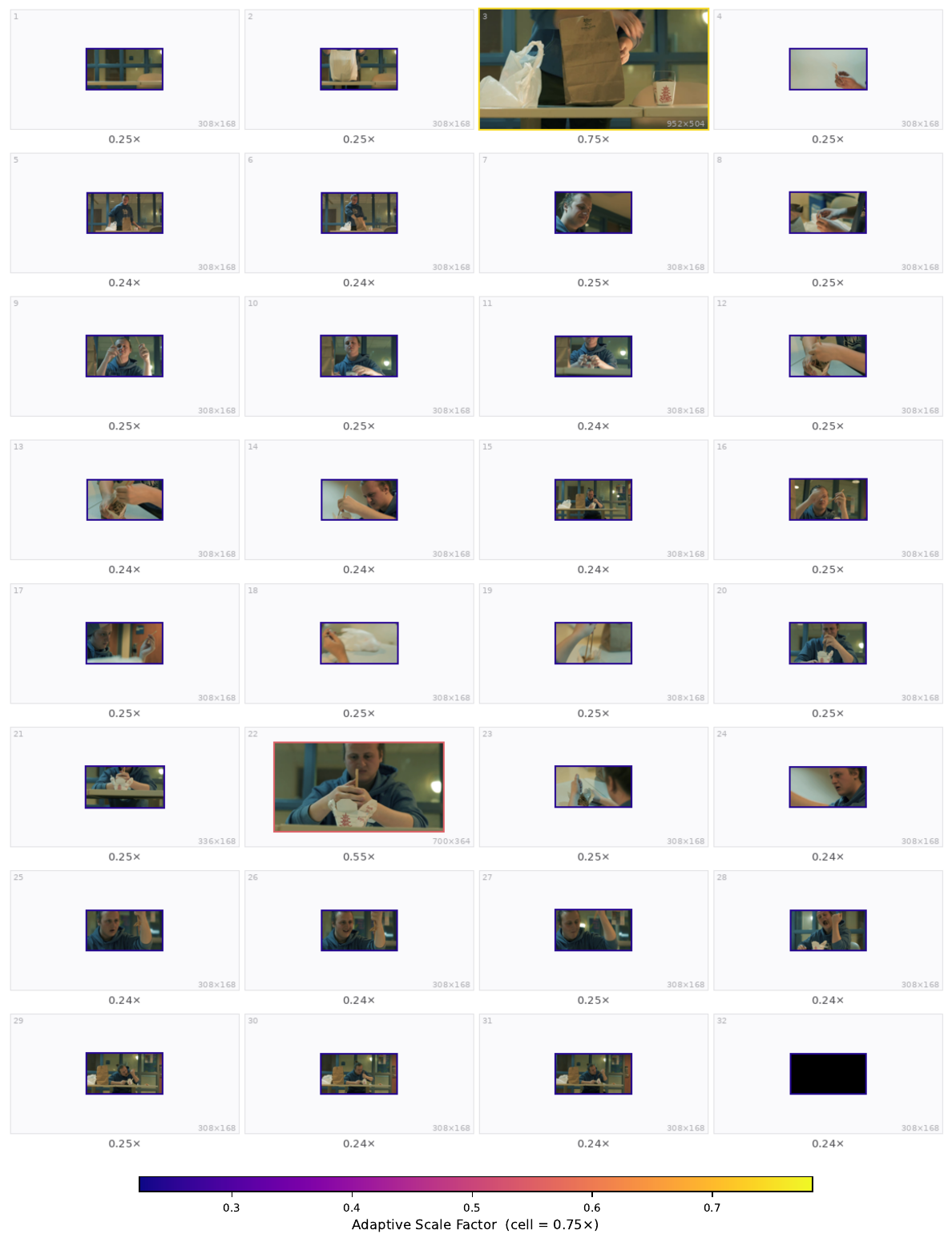}
    \caption{\textbf{Case~4: VideoMME~\citep{fu2025video} (Vanilla \textcolor{green!60!black}{$\checkmark$} $\to$ \our{} \textcolor{red}{$\times$}; failure case).} A nearby frame is enlarged, but the actual fork-bearing frame is compressed. The decisive fine detail is therefore lost at exactly the wrong moment.}
    \label{fig:case_fork}
\end{figure*}

We provide four qualitative analyses mapping allocation behavior to reasoning outcomes (Figures~\ref{fig:case_comp}--\ref{fig:case_fork}). We render 32 uniformly sampled frames at their predicted scales; warmer borders denote aggressive upscaling.

\textbf{Task-Dependent Operating Regimes.}
Figures~\ref{fig:case_comp} and~\ref{fig:case_adapt} contrast two Video-MMMU tasks drawn from identical educational domains that trigger markedly different allocation strategies. In the comprehension task, evidence localizes strictly within diagram-heavy slides. The policy executes a sparse regime, aggressively compressing lecturer frames and suppressing an irrelevant quiz slide. Conversely, the adaptation task requires parsing a dense numeric table to compute a $\chi^2$ statistic. The policy instantly shifts to a high-budget regime, broadly preserving fidelity and strongly upscaling the table frames. The policy reacts dynamically to downstream reasoning requirements, not merely superficial visual clutter.

\textbf{Evidence Localization and Failure.}
Figure~\ref{fig:case_zodiac} (VideoMME) demonstrates precision localization. The policy isolates and magnifies brief frames containing critical date overlays, aggressively downscaling repetitive sky footage. Figure~\ref{fig:case_fork} exposes the prevailing failure mode. A decisive visual cue (a fork) appears briefly against a simple background. The policy mistakenly upscales an adjacent frame while compressing the critical frame, destroying the fine-grained evidence at the exact moment of relevance. This aligns with our robustness analysis: \our{} excels at broad concentration but remains brittle against highly transient, low-contrast cues.

\subsection{Boundary-Case Transfer Beyond Video}
\label{app:image_transfer}

While \our{} targets video QA and temporal grounding, we probe image transfer to identify operational boundaries. Table~\ref{table:benchmark_image} shows that the video-trained policy occasionally identifies images requiring high fidelity (e.g., ChartQA), but does not deliver consistent efficiency-preserving gains on dense static-image benchmarks. The boundary is clear: input-side allocation generalizes across video tasks and operators, but a strictly video-trained policy requires explicit joint training to handle static image distributions reliably.

\begin{table*}[t]
    \centering
    \footnotesize
    \setlength{\tabcolsep}{0.55em}
    \renewcommand{\arraystretch}{1.08}
    \caption{\textbf{Exploratory zero-shot transfer to image benchmarks.} Parenthetical values denote per-task retention ratio $R$, and \our{}-RL additionally fine-tunes the MLLM via RL.}
    \resizebox{\linewidth}{!}{
    \begin{tabular}{lcccccc}
    \toprule
    \textbf{Model} & \makecell{\textbf{MathVista}\\testmini} & \makecell{\textbf{MMMU}\\val} & \makecell{\textbf{OCRBench}} & \makecell{\textbf{ChartQA}} & \makecell{\textbf{AI2D}} & \makecell{\textbf{TextVQA}\\val} \\
    \midrule
    Qwen2.5-VL-7B & 49.1\tiny{(100\%)} & 50.9\tiny{(100\%)} & 84.2\tiny{(100\%)} & 83.9\tiny{(100\%)} & 82.5\tiny{(100\%)} & 82.9\tiny{(100\%)} \\
    Random Drop & 44.8\tiny{(50\%)} & 49.0\tiny{(50\%)} & 74.8\tiny{(50\%)} & 71.6\tiny{(50\%)} & 80.3\tiny{(50\%)} & 78.1\tiny{(50\%)} \\
    ToMe~\citep{bolya2022token} & 46.2\tiny{(50\%)} & 49.6\tiny{(50\%)} & 79.3\tiny{(50\%)} & 78.1\tiny{(50\%)} & 81.9\tiny{(50\%)} & 81.2\tiny{(50\%)} \\
    VisionZip~\citep{visionzip} & 47.2\tiny{(50\%)} & 48.6\tiny{(50\%)} & 79.6\tiny{(50\%)} & 77.9\tiny{(50\%)} & 81.9\tiny{(50\%)} & 81.3\tiny{(50\%)} \\
    \cellcolor{metabg}\textbf{\our{}}\tiny{(Qwen2.5-VL-7B)} & \cellcolor{metabg}45.5\tiny{(42\%)} & \cellcolor{metabg}51.0\tiny{(29\%)} & \cellcolor{metabg}80.0\tiny{(64\%)} & \cellcolor{metabg}85.9\tiny{(105\%)} & \cellcolor{metabg}81.4\tiny{(41\%)} & \cellcolor{metabg}69.6\tiny{(30\%)} \\
    \cellcolor{metabg}\textbf{\our{}-RL}\tiny{(Qwen2.5-VL-7B)} & \cellcolor{metabg}46.7\tiny{(42\%)} & \cellcolor{metabg}50.9\tiny{(29\%)} & \cellcolor{metabg}80.8\tiny{(64\%)} & \cellcolor{metabg}86.6\tiny{(105\%)} & \cellcolor{metabg}81.1\tiny{(41\%)} & \cellcolor{metabg}70.1\tiny{(30\%)} \\
    \cdashline{1-7}
    Qwen3-VL-8B & 56.1\tiny{(100\%)} & 53.4\tiny{(100\%)} & 85.0\tiny{(100\%)} & 84.0\tiny{(100\%)} & 83.5\tiny{(100\%)} & 82.1\tiny{(100\%)} \\
    Random Drop & 47.3\tiny{(50\%)} & 48.7\tiny{(50\%)} & 62.9\tiny{(50\%)} & 70.2\tiny{(50\%)} & 79.7\tiny{(50\%)} & 76.6\tiny{(50\%)} \\
    VisionZip~\citep{visionzip} & 47.8\tiny{(50\%)} & 50.3\tiny{(50\%)} & 70.5\tiny{(50\%)} & 75.0\tiny{(50\%)} & 80.5\tiny{(50\%)} & 79.3\tiny{(50\%)} \\
    ToMe~\citep{bolya2022token} & 49.6\tiny{(50\%)} & 50.6\tiny{(50\%)} & 70.3\tiny{(50\%)} & 75.2\tiny{(50\%)} & 80.5\tiny{(50\%)} & 79.4\tiny{(50\%)} \\
    \cellcolor{metabg}\textbf{\our{}}\tiny{(Qwen3-VL-8B)} & \cellcolor{metabg}52.5\tiny{(42\%)} & \cellcolor{metabg}50.9\tiny{(29\%)} & \cellcolor{metabg}82.7\tiny{(64\%)} & \cellcolor{metabg}83.2\tiny{(105\%)} & \cellcolor{metabg}81.2\tiny{(41\%)} & \cellcolor{metabg}67.8\tiny{(30\%)} \\
    \bottomrule
    \end{tabular}
    }
    \label{table:benchmark_image}
\end{table*}

\end{document}